\documentclass[a4paper,fleqn]{cas-dc}

\usepackage[authoryear]{natbib}

\usepackage{multirow}
\usepackage{kotex}
\usepackage{blindtext}

\usepackage{diagbox}
\usepackage{subcaption}
\usepackage{multirow}
\usepackage{multicol}
\usepackage{siunitx}
\usepackage{epsfig}
\usepackage{graphicx}
\usepackage{amsmath}
\usepackage{amssymb}
\usepackage{pifont}

\usepackage{microtype}
\usepackage{booktabs} 
\usepackage{amsfonts} 
\usepackage{mathtools}
\usepackage{xcolor}

\usepackage{xspace}

\DeclarePairedDelimiterX{\infdivx}[2]{(}{)}{%
  #1\;\delimsize|\delimsize|\;#2%
}
\newcommand{\kld}[2]{\ensuremath{KL\infdivx{#1}{#2}}\xspace}

\DeclareMathOperator*{\argmax}{arg\,max}

\usepackage{tikz}
\usepackage{smartdiagram} 

\usetikzlibrary{mindmap,shadows,backgrounds}
\newcommand*{\info}[4][16.3]{%
  \node [ annotation, #3, scale=0.75, text width = #1em,
          inner sep = 2mm ] at (#2) {%
  \list{$\bullet$}{\topsep=0pt\itemsep=0pt\parsep=0pt
    \parskip=0pt\labelwidth=8pt\leftmargin=8pt
    \itemindent=0pt\labelsep=2pt}%
    #4
  \endlist
  };
}

\usepackage{amsthm}
\theoremstyle{definition}
\newtheorem{definition}{Definition}

\def\tsc#1{\csdef{#1}{\textsc{\lowercase{#1}}\xspace}}
\tsc{WGM}
\tsc{QE}
\tsc{EP}
\tsc{PMS}
\tsc{BEC}
\tsc{DE}

\begin{document}
\let\WriteBookmarks\relax
\def\floatpagepagefraction{1}
\def\textpagefraction{.001}
\shorttitle{A Wholistic View of Continual Learning with Deep Neural Networks}
\shortauthors{Martin Mundt et~al.}

\title [mode = title]{A Wholistic View of Continual Learning with Deep Neural Networks: Forgotten Lessons and the Bridge to Active and Open World Learning}                      

\author[1,3]{Martin Mundt}
\ead{martin.mundt@tu-darmstadt.de}
\cormark[1]
\author[2]{Yongwon Hong}
\ead{yhong@yonsei.ac.kr}
\author[1]{Iuliia Pliushch}
\ead{pliushch@em.uni-frankfurt.de}
\author[1]{Visvanathan Ramesh}
\ead{vramesh@em.uni-frankfurt.de}

\cortext[cor1]{Corresponding author. Main work conducted while affiliated with Goethe University. Present affiliation is with TU Darmstadt.}

\address[1]{Department of Computer Science, Goethe University, Theodor-W.-Adorno-Platz 1, 60323 Frankfurt, Germany}
\address[2]{Department of Computer Science, Yonsei University, 50 Yonsei-ro, Seodaemun-gu, 03722 Seoul, Republic of Korea}
\address[3]{Department of Computer Science, TU Darmstadt, Karolinenplatz 5, 64289 Darmstadt, Germany}

\begin{abstract}
Current deep learning methods are regarded as favorable if they empirically perform well on dedicated test sets. This mentality is seamlessly reflected in the resurfacing area of continual learning, where consecutively arriving data is investigated. The core challenge is framed as protecting previously acquired representations from being catastrophically forgotten. However, comparison of individual methods is nevertheless performed in isolation from the real world by monitoring accumulated benchmark test set performance. The closed world assumption remains predominant, i.e. models are evaluated on data that is guaranteed to originate from the same distribution as used for training. This poses a massive challenge as neural networks are well known to provide overconfident false predictions on unknown and corrupted instances. In this work we critically survey the literature and argue that notable lessons from open set recognition, identifying unknown examples outside of the observed set, and the adjacent field of active learning, querying data to maximize the expected performance gain, are frequently overlooked in the deep learning era. Hence, we propose a consolidated view to bridge continual learning, active learning and open set recognition in deep neural networks.  Finally, the established synergies are supported empirically, showing joint improvement in alleviating catastrophic forgetting, querying data, selecting task orders, while exhibiting robust open world application.
\end{abstract}


\maketitle

\section{Introduction}
With the ongoing maturing of practical machine learning systems, the community has found a resurfacing interest in continual learning \citep{Thrun1996,Thrun1996a}. In contrast to the broadly practiced \textit{learning in isolation}, where the algorithmic training phase of a system is constrained to a single stage based on a previously collected i.i.d. dataset, \textit{continual learning} entails a learning process that leverages data as it arrives over time. In spite of this paradigm having found various application in many machine learning systems, for a review see the recent book on lifelong machine learning by \citet{Chen2017}, the advent of deep learning seems to have steered the focus of current research efforts towards a phenomenon known as \textit{catastrophic interference} or alternatively \textit{catastrophic forgetting} \citep{McCloskey1989,Ratcliff1990}, as suggested by recent reviews \citep{Farquhar2018a,Parisi2019,DeLange2019,Lesort2020} and empirical surveys of deep continual learning \citep{DeLange2019,Lesort2019,Pfulb2019}. 
The latter is an effect particular to machine learning models that update their parameters greedily according to the presented data population, such as a neural network iteratively updating its weights with stochastic gradient estimates. When continuously arriving data is included that leads to any shift in the data distribution, the set of learned representations is guided unidirectionally towards approximating any task's solution on the data instances the system is presently being exposed to. The natural consequence is overwriting former learned representations, resulting in an abrupt forgetting of previously acquired information. 

Whereas current works predominantly concentrate on alleviating such forgetting in continual deep learning through the design of specialized mechanisms, we argue that there is a growing risk in the continual learning field becoming overly narrow.  
There clearly have been commendable efforts towards preserving neural network representations in continuous training. However, such a high focus is given on the practical requirements and trade-offs of metrics that surround catastrophic forgetting \citep{Kemker2018}, e.g. inclusion of memory footprint, computational cost, cost of data storage, task sequence length and amount of training iterations, $\ldots$ \citep{Diaz-Rodriguez2018,Farquhar2018a}, that it could almost be seen as misleading when most current systems break immediately if unseen unknown data or minor corruptions are encountered during deployment \citep{Matan1990,Boult2019,Hendrycks2019}. 
The assumption of a \textit{closed world} seems omnipresent. In other words, there is a common belief that the model will always exclusively encounter data that stems from the same data distribution as encountered during training. This is highly unrealistic in the real \textit{open world}, where data can vary to extents that are impractical to capture into training sets or users have the ability to give almost arbitrary input to systems for prediction. It has been a well known fact for decades that neural networks are wrongly \textit{overconfident} in such real world settings \citep{Matan1990}.
In spite of the inevitable danger of neural networks generating entirely meaningless predictions when encountering unseen unknown data instances, current efforts towards benchmarking continual learning conveniently circumvent this challenge. Select exceptions attempt to solve the tasks of recognizing unseen and unknown examples, rejecting nonsensical predictions or setting them aside for later use, typically summarized under the umbrella of \textit{open set recognition}. Nevertheless, the majority of existing deep continual learning systems remain black boxes that unfortunately do not exhibit desirable robustness to respective miss-predictions on unknown data, dataset outliers or common corruptions \citep{Hendrycks2019}. 

Apart from current benchmarking practices still being constrained to the closed world, another unfortunate trend is a lack of understanding for the nature of created continual learning datasets. Both continual generative modeling, \cite{Shin2017,Achille2018,Farquhar2018,Nguyen2018,Wu2018,Zhai2019}, as well as the bulk of class incremental continuous learning works \cite{Li2016,Kirkpatrick2017,Rebuffi2017,Lopez-Paz2017,Kemker2018,Kemker2018a,Xiang2019} generally investigate sequentialized versions of time-tested visual classification benchmarks. For instance, in popular class incremental MNIST \citep{LeCun1998}, CIFAR \citep{Krizhevsky2009} or ImageNet \citep{Russakovsky2015}, individual classes are simply split into disjoint sets and are shown in sequence. In favor of retaining comparability on a benchmark, questions about the effect of task ordering or the impact of overlap between tasks are routinely overlooked. Notably, lessons learned from the adjacent field of \textit{active machine learning}, a particular form of semi-supervised learning, do not seem to be integrated into modern continual learning practice. In active learning the objective is to learn to incrementally find the best approximation to a task's solution under the challenge of letting the system itself query what data to include next. As such, it can be seen as an antagonist to alleviating catastrophic forgetting. Whereas current continual learning is occupied with maintaining the information acquired in each step without endlessly accumulating all data, active learning has focused on the complementary question of identifying suitable data for the inclusion into an incrementally training system. Although early seminal works in active learning have rapidly identified the challenges of robust application and pitfalls faced through the use of heuristics \citep{Roy2001,Settles2008,Li2013}, the latter are nonetheless once again dominant in the era of deep learning \citep{Beluch2018,Geifman2019,Gal2015,Srivastava2014} and the challenges seem to be faced anew. \\
 
With the above challenges in mind, we can rapidly build our intuition for why they are connected if we briefly take a look at autonomous driving, as one practical example. If we aim to learn how to drive in new environments, it is not only sufficient to make sure that we are capable of learning from new data while preserving existing knowledge. It is similarly important to acknowledge that newly arriving data may be skewed in uninteresting or potentially harmful ways. New falsely predicted objects could appear, particularly rare events could pose a threat, and sensors may deteriorate or fail. Identifying what is already known and distinguishing it with unseen novel instances is essential. Deciding which of this new data is meaningful and which should be discarded for future learning provides closure to the learning cycle. \\
 
In this work we thus make a first effort towards a principled and consolidated view of deep continual learning, active learning and learning in the open world. We start with a historical perspective and by providing a review of each topic in isolation. We then proceed to identify notable previous lessons that appear to receive less attention in modern deep learning. Speaking hyperbolically, they appear to have been ``forgotten'' in many recent continual learning works. As we will see throughout the survey, individual fields may have been studied extensively by themselves in isolation, but their impact seems to be largely overlooked when considered together. We will continue to argue that these seemingly separate topics do not only benefit from the viewpoint of the other, but \textit{should} be regarded in conjunction. In this sense, we propose to extend current continual learning practices towards a broader view of continual learning as an umbrella term.  Our survey thus complements existing continual learning reviews \citep{Farquhar2018a,Parisi2019,DeLange2019,Lesort2020},  but instead of surveying the mathematical foundation of every individual algorithm to alleviate catastrophic forgetting in detail,  we provide a more critical overview. As a crucial difference, we connect thought patterns towards continual learning that naturally encompasses and builds upon prior insights from active learning and open set recognition. 
To highlight the correspondingly developed synergies and showcase their practical potential, we complement our consolidated survey with empirical evidence supporting various important aspects: extraction of exemplars or core sets, active data queries, robustness to open world corruptions, and choosing a task order curriculum.  For this purpose, we adapt and extend a recently proposed approach based on variational Bayesian inference in neural networks \citep{Mundt2022,Mundt2019} to illustrate one potential choice towards a comprehensive framework.  Importantly,  we emphasize that we do not propose this approach as a universal or unique solution,  but use it to highlight the importance of the viewpoints developed in this paper.

\section{Preamble: continual machine learning}
It is likely that the idea of continual machine learning dates back to a similar period of time to the surfacing of machine learning itself. There have been many attempts at defining concepts such as continuous, lifelong or continual machine learning. Often these terms feature negligible nuances and can generally be taken as synonyms. However it seems difficult, and perhaps is not constructive, to attempt to pin-point the exact onset of when something should be referred to as continual or lifelong learning.
Instead, in this preamble, we will present definitions and related paradigms that have come to enjoy great popularity in the machine learning community.  Note that many of these definitions are not necessarily formal or mathematical, but are nevertheless illustrated here for a historical perspective.  Some paradigms are already, or if not yet, should be considered subsets of continual learning (CL). As a standalone paradigm they vary primarily in their current evaluation protocols. We will briefly introduce each of these paradigms and then proceed to summarize and identify characteristic differences with respect to the broader term of modern continual learning.  \\

The first widely circulated definition of \textit{lifelong machine learning} (LML) originated in the work proposed by \citet{Thrun1996, Thrun1996a}. This definition is as follows:

\begin{definition}{\emph{\citet{Thrun1996, Thrun1996a} - Lifelong Machine Learning:}}\label{def:ThrunLLML}
The system has performed \textit{N} tasks. When faced with the (\textit{N}+1)th task, it uses the knowledge gained from the \textit{N} tasks to help the (\textit{N}+1)th task.
\end{definition}

Here, the unmentioned essence is that the data of the first $N$ tasks is generally assumed to be no longer available at the time of learning about the $(N+1)$th task. That is, the observed data is not just endlessly accumulated and stored explicitly. Whereas this definition captures the basic idea behind continued learning, it is also ambiguous with respect to the definition of task and knowledge. There have been many attempts to find a more concise definition across the literature over the years. One of the more succinct, yet still decently generic definitions followed in the work of \citet{Chen2017}: 

\begin{definition}{\emph{\citet{Chen2017} - Lifelong Machine Learning}:}\label{def:ChenLiu}
Lifelong Machine Learning is a continuous learning process. At any time point, the learner performed a sequence of \textit{N} learning tasks, $\mathcal{T}_1, \mathcal{T}_2, \ldots, \mathcal{T}_{\mathit{N}}$. These tasks can be of the same type or different types and from the same domain or different domains. When faced with the (\textit{N}+1)th task $\mathcal{T}_{\mathit{N}+1}$ (which is called the new or current task) with its data $\mathcal{D}_{\mathit{N}+1}$, the learner can leverage past knowledge in the knowledge base (KB) to help learn $\mathcal{T}_{\mathit{N}+1}$. The objective of LML is usually to optimize the performance on the new task $\mathcal{T}_{\mathit{N}+1}$, but it can optimize any task by treating the rest of the tasks as previous tasks. KB maintains the knowledge learned and accumulated from learning the previous task. After the completion of learning $\mathcal{T}_{\mathit{N}+1}$, KB is updated with the knowledge (e.g. intermediate as well as the final results) gained from learning $\mathcal{T}_{\mathit{N}+1}$. The updating can involve inconsistency checking, reasoning, and meta-mining of additional higher-level knowledge. 
\end{definition}

\begin{figure}[t]
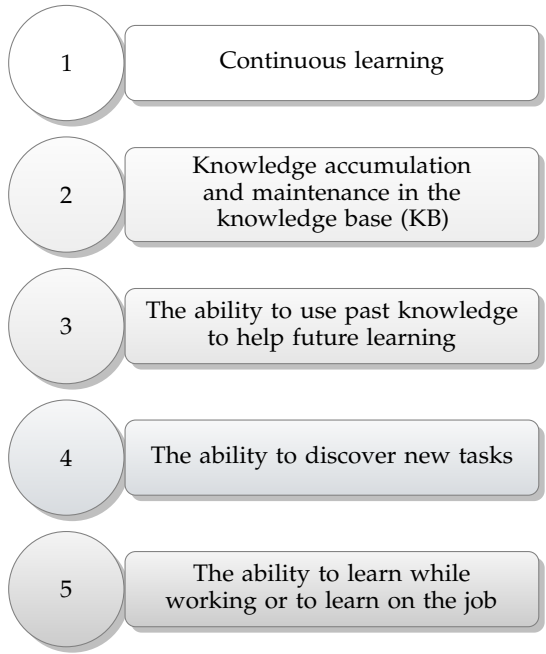

\include{LML_pillars}
\vspace*{-0.5cm}
\caption{\label{fig:LML_pillars} The five main pillars of lifelong machine learning according to \citet{Chen2017}. Note that the first three pillars were originally proposed and the last two added recently in a second edition redefinition to emphasize new frontiers.}
\end{figure}

The authors of this latter definition argue that it can be summarized into three key characteristics: continuous learning; knowledge accumulation and maintenance in the knowledge base (KB); the ability to use past knowledge to help future learning. In contrast to the previous definition by \citet{Thrun1996, Thrun1996a}, mainly the notion of a maintained knowledge base is introduced. Here LML is now defined such that at any given point in time performance can be optimized for any given task by treating all other tasks as previously presented, irrespective of their original order. Whereas the original definition optimized towards benefiting $\mathcal{T}_{\mathit{N}+1}$ in only one direction, thus allowing for performance of previous tasks to degrade over time, \citet{Chen2017} explicitly formulate the preservation of all accumulated information as a fundamental goal of LML. 
In a recent second iteration of this definition, the authors have added two additional desiderata: the ability to discover new tasks and the ability to learn while working. We have visualized these five essential pillars of LML in Figure \ref{fig:LML_pillars}.

Although acknowledged by the authors themselves, this extended definition still lacks with respect to certain aspects:
\begin{itemize}
\item a coherent description of domain. This is currently not used unanimously in the literature and often applied interchangeably with task.
\item a formalization of knowledge or respective representation thereof in the KB. Typically this is practically constrained to specific applications.
\item the essential question of evaluation practice, i.e. choosing, ordering and evaluating the sequence of tasks. This generally requires a human in the loop and considered evaluation scenarios can vary immensely between individual works.
\end{itemize}

There are many more encountered open questions with LML in practice, especially with respect to modern machine learning algorithms based on deep learning. As the latter is primarily based on the use of neural networks (NN), they will constitute the main focus of this paper. While the presented arguments will often be of generic nature, this has the advantage that the concept of a knowledge base and its maintenance collapses to the question of managing the model's learned representations and optional data memory buffers containing past experiences. This is an important distinction, and perhaps a simplification, in comparison to the way \citet{Chen2017} (and figure \ref{fig:LML_pillars}) originally use the term knowledge. Here, the latter is adopted in the spirit of the older ``never-ending learning'' systems for language and images (NELL and NEIL)\citep{Carlson2010, Chen2013, Mitchell2015}, which take a more ``traditional'' AI approach. In addition to learned parameters and original data instances, the respective knowledge bases leverage various neuro-symbolic techniques, such as accounting for explicit context, extracting relations, and involving rules. Concepts that are typically not accounted for in deep neural networks.

At the same time, the presently more collapsed notion of knowledge in a deep neural network can make the question of how to leverage prior information quite involved. Representations in NNs are densely entangled within layers as well as distributed hierarchically across layers. Although we constrain ourselves to NNs, we importantly emphasize that the terms knowledge and knowledge base will be used beyond a narrow interpretation of data instances or parameters in the remainder of the manuscript, retaining a broader interpretation for future work. Before delving into a review of contemporary works, their merits and current limitations, we will present various popular paradigms that are related to the former definitions of LML. This will then be followed by a brief summary on evaluation practices to highlight the nuances.
 
 \subsection{Related paradigms: subsets of continual learning}
Over the course of machine learning development, various different paradigms and evaluation practices have evolved. Throughout this paper, we will come to the already apparent conclusion that CL should ideally be defined as a superset. We will make an attempt towards a definition that is more encompassing of the potential elements at the end of this manuscript. For now, we start by introducing commonly considered machine learning paradigms. As a word of caution, the following definitions should be regarded as non-exhaustive. Even though we have made a considerable effort to provide a comprehensive amount of references, the practical use of certain terminology in particular may still vary largely from community to community. The following shall thus reflect the common use in modern deep learning. 
		
We begin with transfer learning as it can intuitively be regarded as the most related concept. Originally, transfer learning has been proposed as converting a weak learner, one that performs marginally better than random guessing, to one that produces stronger hypotheses \citep{Schapire1990}. The corresponding formulation that is more specific to neural networks is how the representations obtained by learning through backpropagation can be ``recycled'' for new tasks \citep{Pratt1991,Pratt1993}. This challenge initially wasn't unanimously referred to as transfer learning, but often was referred to as boosting \citep{Freund1997}. A pre-deep learning survey \citep{Pan2010} has summarized efforts and formalized transfer learning in the way used today: 
\begin{definition}{\emph{Transfer Learning} \cite{Pan2010}}:\label{def:transfer_learning}
Given a source domain and learning task, a target domain and learning task, \emph{transfer learning} aims to help improve the learning of the target predictive function in the target domain using the knowledge in the source domain and task, where the source and target domain, or the source and target task are unequal.\footnote{Note that mathematical symbols (such as $D_{S}$ or $D_{T}$ to denote source and target domain) have been omitted from the original definition for ease of readability. We will continue to omit these symbols in the follow-up definitions as they do not serve a higher purpose in the current overview.}
\end{definition}

Here, \citet{Pan2010} formalize the use of the terms \emph{domain} and \emph{task} in the context of supervised transfer with datasets consisting of a finite amount of data instances. They are defined by the following: Given a specific domain, defined as the pair of marginal data distribution and a corresponding feature space, a task consists of two components: a label space and an objective predictive function (which maps to the label space and is not observed, but can be learned from the training data, consisting of pairs of data instances and respective labels) \citep{Pan2010}. The concept of a domain is therefore defined as the pair of marginal data distribution and a corresponding feature space, where it is generally implied that source and target feature space, or source and target data sets are unequal. An effortless translation of transfer learning to unsupervised or reinforcement learning settings is possible.
Without further extensions, this definition of transfer learning is essentially a narrowed down version of the primitive lifelong learning definition \ref{def:ThrunLLML}, with the nuance that there typically only exist two tasks. It is similarly one directional in the sense that the source task is only used to improve learning the new target. 

Since then an enormous amount of works has sprouted, initiated by works that have started the investigation of transferability of deep neural network features beyond low-level patterns \citep{Oquab2014,Yosinski2014}, i.e. the higher abstractions and task-specific information believed to be encoded in deeper layers of the hierarchy. \citet{Weiss2016} have provided a survey on recent advances. In this context of feature transferability, a variant named \textit{multi-task learning} (MTL) has emerged. \citet{Caruana1997} summarizes the goal of MTL succinctly: \emph{``MTL improves generalization by leveraging the domain-specific information contained in the training signals of related tasks''}. Early works sometimes referred to this as including ``hints'' \citep{Suddarth1990,Abu-Mostafa1990} to improve learning. In contrast to transfer learning, generally multiple tasks are considered, with the requirement of the model performing well on all of them. However, in the MTL setting, tasks are all trained jointly and no sequence is assumed, corresponding to typical isolated learning practice. In modern day deep networks, MTL thus culminates in the question of how to exactly share the abundant amount of parameters in the architectural hierarchy, see e.g. the overview provided by \citet{Ruder2017} for variants of sharing architecture portions.

More recently, a very specific form of transfer or multi-task learning has evolved. \textit{Few-shot Learning} \citep{Fei-Fei2006} developed due to the inability of deep learning techniques to cope with small datasets and empirical risk optimization being unreliable in small sample regimes. \citet{Wang2020} summarized few-shot learning as a type of machine learning problem, where the dataset only contains a limited number of examples with supervised information for the target domain (and generally no constraints on the source domain). This implies that few-shot learning also tackles the issue of rare cases, apart from computational cost and the issue of data collection and labelling. When there is only one example with a label, it is commonly referred to as \textit{one-shot learning} \citep{Fink2005, Fei-Fei2006}. Respectively, if no supervised example is provided, the scenario is referred to as \textit{zero-shot learning} \citep{Lampert2009}. These scenarios are typically regarded under the hood of transfer learning with additional constraints on data availability. 

Apart from concerns about reasonably sized datasets, a different concern is as old as the search for stochastic approximations itself, namely when to conduct updates. Already in the work of \citet{Hebb1949}, \textit{online learning}, i.e. incorporating information immediately as data arrives as opposed to collecting batches before updating a model, was a natural requirement. This question has been elemental in later formalization of frameworks for empirical risk optimization \citep{Tsypkin1971, Vapnik1982}. Several works have elaborated on challenges in: online learning in NNs \citep{Heskes1993}, more generally online learning and stochastic approximations \citep{Bottou1999,Saad1999}, or specifically online gradient descent \citep{Zinkevich2003}, the workhorse of modern optimization. Given the instance based update nature, online learning in neural networks is inherently tied to the question of how to avoid catastrophic interference. It is thus not surprising that with the advent of DL immediate attempts have been made to consider online learning in DNNs \citep{Zhou2012}, see a recent survey by \citet{Sahoo2018}. Nevertheless, research towards online learning still revolves around the interaction between online desiderata and stochastic approximations, or the stochastic gradient descent with backpropagation procedure in particular. 

Ultimately, each paradigm arose for a reason and comes with its own value, namely that of providing better distinction to other works in concrete evaluation scenarios. However, it is important to remember that the emerging taxonomy is full of nuances that are at times indistinguishable in a more general framework. In consequence, evaluation protocols are central to any discussion. We therefore proceed with details of common evaluation methods in deep continual learning and then summarize the main differences to the paradigms introduced in this section for a compact overview.
	
\subsection{Continual learning evaluation}
In contrast to isolated machine learning, where the evaluation scenario can often be defined in a straightforward manner by employing performance or satisfying task metrics, continual learning does not directly allow for such an approach. Given that the interest lies in accumulation of information, there are many factors to consider in evaluation of corresponding algorithms. In general it is important to monitor the currently introduced task, yet also investigate semantic drift on previous tasks. One should consider the gain and the ability to leverage representations from task to task in progressive experimentation, yet take note of the task sequence that is crucial to the specific solution obtained. When introducing more tasks, the transfer behavior should be carefully examined, yet cautiously interpreted, as not all introduced tasks yield immediate benefits and thus a larger amount of tasks needs to be brought in to the system.
\begin{figure}[t]
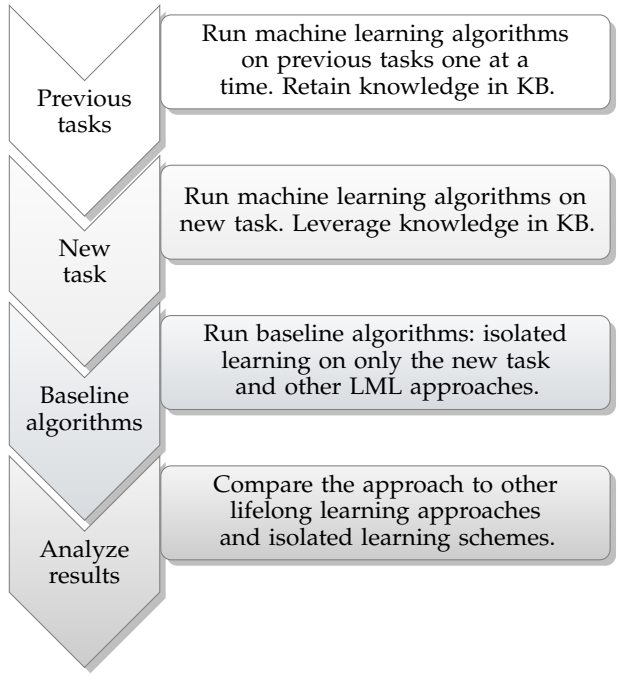

\include{LML_evaluation}
\vspace*{-0.5cm}
\caption{\label{fig:LML_evaluation} A widely used approach to evaluation of lifelong machine learning algorithms in the literature \citep{Chen2017}.}
\end{figure}

Before continuing with the discussion of evaluation difficulties and metrics, let us take a brief look at some currently employed evaluation methodology \citep{Chen2017}, summarized visually in Figure \ref{fig:LML_evaluation}. 
It seems that such an evaluation protocol is still largely inspired by the isolated machine learning practices. Whereas the notion of information transfer and the sequence of tasks is considered and benchmarked against isolated learning algorithms, such an approach to evaluating the value of continual learning algorithms disregards the relevance of the task sequence (or permutation thereof), choice of tasks or choice of data. 
Accordingly, recently developed experimental protocols in deep continual learning \citep{Farquhar2018a,Kemker2018,Parisi2019,DeLange2019,Lesort2019,Pfulb2019,Lesort2020} seem to mainly occupy themselves with evaluation procedures that are heavily inspired by decades of benchmarking learning algorithms in isolation. As a reminder to the reader, we refer to isolated learning as the practice of end-to-end training on a static dataset and evaluation on its predefined test set, without changes over time. As such, the majority of current empirical examination equates continual learning benchmarks with the monitoring of catastrophic forgetting in scenarios that are simple sequentialized versions of popular datasets, similarly to the steps shown in Figure \ref{fig:LML_evaluation}. With few exceptions, this means that existing datasets are simply split into $t=1, \ldots, T$ sets, where each of these sets is referred to as one task. These task- or time-stamped sets are then presented one by one to a deep learning system. Typically, each step is assumed to consist of a disjoint set of classes or entire datasets, usually independently of whether the probed task is of supervised, unsupervised or semi-supervised nature, see Figure \ref{fig:CL_dataset} for an illustration. Respectively analyzed metrics \citep{Kemker2018} are based on this dataset sequentialization and routinely monitor e.g. the degradation of a first task's classification accuracy, the ability to encode new task increments, the overall development of a chosen metric as tasks accumulate or various similar measures to gain an intuition for generative models. It is obvious how this is inspired by isolated learning as these metrics can simply be extracted from a conventional confusion matrix. 
For this reason, multiple efforts have been made to emphasize the need for more diverse evaluation \citep{Diaz-Rodriguez2018, Farquhar2018a}. Alas, the persisting focus on catastrophic forgetting remains visible from the formulated criteria and questions that are deemed necessary to compare methods \citep{Diaz-Rodriguez2018, Farquhar2018a}:
\begin{figure}
	\centering
	\includegraphics[width=0.875 \columnwidth]{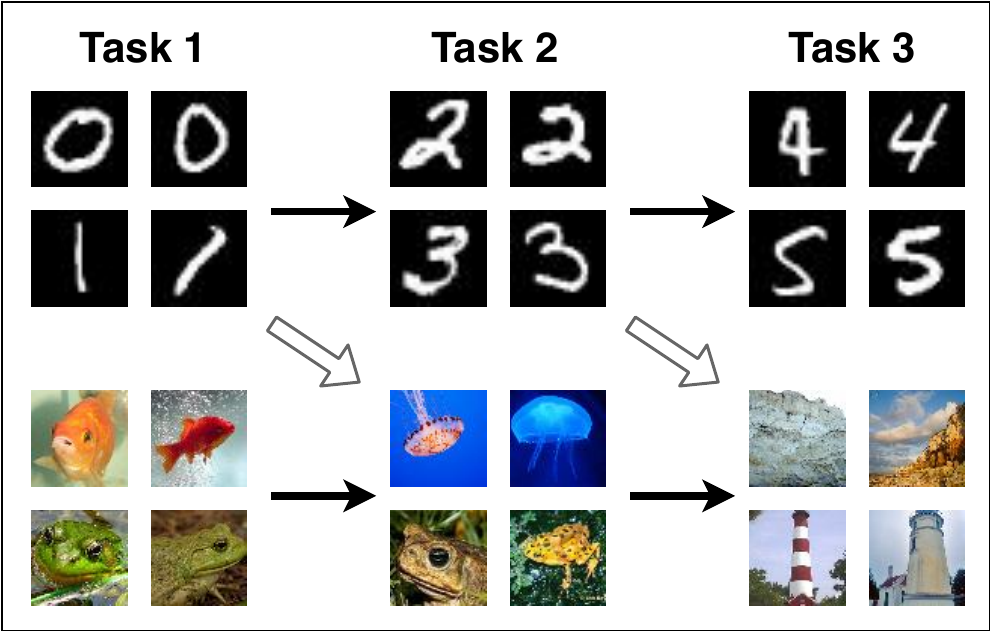}
	\caption{\label{fig:CL_dataset} A typical continual learning scenario dividing common benchmark datasets into a sequence of sub-tasks. Here, the digits one through six from the MNIST dataset \citep{LeCun1998} and the Wordnet ids ``n01443537'': goldfish, ``n01641577'': bullfrog, ``n01644900'': tailed frog, ``n01910747'': jellyfish, ``n09246464'': cliff, ``n02814860'': beacon from the ImageNet dataset \citep{Russakovsky2015}. Common evaluation either follows the filled dark arrows to incrementally learn one dataset or alternatively also switches dataset, as denoted by the hollow light arrows.}
\end{figure}
	
\begin{itemize}
	\item{{\color{orange} \textbf{Memory consumption:}}  \small{amount of required memory.}}
	\item{{\color{orange} \textbf{Amount of stored data:}}  \small{how much past data does the method need to retain explicitly?}}
	\item{{\color{orange} \textbf{Task boundaries:}} \small{does the method require clear task divisions?} }  
	\item{{\color{orange} \textbf{Prediction oracle:}}  \small{does the method require knowing the task label for prediction?}}
	\item{{\color{orange} \textbf{Amount of forgetting:}} \small{how much information is retained as measured through proxy metrics.} }       
	\item{{\color{orange} \textbf{Forward transfer:}}  \small{do older tasks accelerate learning of new concepts?}}   
	\item{{\color{orange} \textbf{Backward transfer:}} \small{do new tasks benefit old tasks?} }     
\end{itemize}

At this stage the reader might already notice that some of these listed items are rather particular to specific practices. For example, the idea that a prediction oracle would be required in the first place in order to give task labels is an artifact of several works that consider so called \textit{multi-head} scenarios. Such a multi-head setting makes use of separate disjoint classifiers per task to circumvent explicitly dealing with task prediction interdependence. In other words, each task is provided a separate label, which is commonly assumed to be additionally provided during inference to decide which of the classifiers should be selected. There exist recent reviews \citep{DeLange2019} that base their entire evaluation on such a scenario. Empirical surveys in the context of robotics \citep{Lesort2019}, generative models \citep{Lesort2020} follow similar trends and conduct a ``comprehensive application-oriented study of catastrophic forgetting'' \citep{Pfulb2019}. With catastrophic forgetting being the sole focus, these works at best cover the first three of the five earlier formulated continual learning pillars \ref{fig:LML_pillars}, if and only if they also conduct an analysis on how specific tasks benefit each other. The recent critiques that formulated above questions \citep{Diaz-Rodriguez2018, Farquhar2018a} therefore present valid attempts to rid current evaluation from such practices that can be seen as inherently violating real continual learning scenarios. Nevertheless, we argue that there are even larger factors at play that transcend these arguments. Although transfer and the sequential nature is considered and benchmarked against isolated learning, crucial aspects such as the \emph{relevance of the task order} (or permutation thereof), \emph{choice of tasks}, \emph{choice of data} and particularly any form of \emph{robustness in an open world} are frequently overlooked or may even be disregarded altogether. Open research areas such as curriculum learning \citep{Bengio2009}, i.e. benefiting from a data ordering of increasing complexity, open world learning \citep{Bendale2016}, i.e. equipping the model with awareness of unseen unknown data, and active learning, i.e. self-selecting data to query for the next step, try to address these crucial elements. We argue that it is imperative to take these perspectives into account in the evaluation of continual learning algorithms. 
Before proceeding to categorize individual works and consequently making an attempt at connecting the paradigms, we give a brief summary of the present evaluation differences. Here, we capture the essence of each paradigm, point out the main difference to continual learning \textit{if the paradigm is viewed in isolation}, and emphasize what role is contributed when considered in context of continual learning.

\begin{itemize}
	\item{{\color{orange} \textbf{Transfer Learning:}} Leverage a source task's representations to accelerate learning or improve a current target task. \\ \emph{Difference to CL when viewed in isolation:} unidirectional knowledge transfer between two tasks. \\ \emph{Role in CL:} enables forward transfer to benefit future tasks through feature re-use.}
	\item{{\color{orange} \textbf{Multi-task Learning:}} Exploit tasks relatedness by forming a joint hypothesis space. \\ \emph{Difference to CL when viewed in isolation:} isolated learning with multiple tasks. \\ \emph{Role in CL:} training of multiple tasks simultaneously before advancing to multiple new tasks continually.}
	\item{{\color{orange} \textbf{Online Learning:}} Retaining and improving a task where data arrives sequentially and real-time constraints require online adaptation.\\ \emph{Difference to CL when viewed in isolation:} typically continuous learning of one task over time, however generally applicable to any of the other paradigms. \\ \emph{Role in CL:} rapid learning without task boundaries, limited revisits of memory buffers, and encoding of new knowledge as data instances arrive in a stream.}
	\item{{\color{orange} \textbf{Few-shot Learning:}} Transfer or multi-task learning in a small data regime. \\ \emph{Difference to CL when viewed in isolation:} unidirectional transfer or isolation similar to transfer learning. \\ \emph{Role in CL:} fast adaptation in continual learning with very few data instances per task.}
	\item{{\color{orange} \textbf{Curriculum Learning:}} Finding a suitable curriculum that accelerates or improves training by means of introducing schedules of increasing data instance difficulty or data instance task specificity. \\ \emph{Difference to CL when viewed in isolation:} isolated learning that prioritizes certain data instances. \\ \emph{Role in CL:} scoring the difficulty of data instances and adapting the pacing of the learning to accelerate or improve training.}
	\item{{\color{orange} \textbf{Open World Learning:}} At any particular point in time the model needs to be able to identify and reject unseen data belonging to unknown tasks. These could be set aside and learned at a later stage.\\ \emph{Difference to CL when viewed in isolation:} Current CL is typically evaluated in a closed world scenario. \\ \emph{Role in CL:} robust learning and inference that discards task irrelevant data and identifies novel tasks.}
	\item{{\color{orange} \textbf{Active Learning:}} An iterative form of supervised learning, where the learner can query a user to provide labels for a subset of unlabelled examples that are deemed to yield the largest knowledge gain. \\ \emph{Difference to CL when viewed in isolation:} data and sampling efficiency is rarely taken into account in CL on predefined benchmarks. \\ \emph{Role in CL:} filtering data instances that are expected to yield large benefit to limit computational resource consumption or labelling costs in continual learning.}
\end{itemize}

\section{Critically surveying and bridging three perspectives}
We provide a critical review of the plethora of practices and historically grown methods in the context of deep continual learning, active learning and open set recognition. For this purpose, we start by surveying the three perspectives individually and categorize their respective trends. To give a visual guideline to the reader, we show an overall taxonomy in Figure \ref{fig:literature_diagram}, where each of the three main nodes will now be discussed in detail in sections \ref{sec:continual_review}, \ref{sec:active_review}, and \ref{sec:osr_review}. We then follow up on these individual perspectives by delving into details of potential pitfalls and shortcoming, in order to subsequently highlight synergies and the necessity for a consolidated view in section \ref{sec:bridge}. That is, we will highlight the illustrated interconnections between the three perspectives of the taxonomy diagram. This consolidated view, based on the primary conjecture that open set recognition provides the natural interface between active and continual learning, is finally presented towards the end of this section. What may at first seem like a tour de force review for the reader, is thus intended to initially gain an overview of the vast landscape and the deluge of methodology options, in order to ground our understanding of the interconnections between the presented elements. As the latter is the primary focus of this work, we re-emphasize that we limit our survey part to concise critical summaries and will forgo lengthy elaborations on algorithmic nuances and mathematical details that are not essential to a generic understanding. For this reason we strongly encourage the reader to go through the ensuing three sections (\ref{sec:continual_review}, \ref{sec:active_review}, and \ref{sec:osr_review}) in favor of a comprehensive and critical picture. However, we acknowledge that a \emph{very well-versed reader in all three paradigms} may want to directly continue with section \ref{sec:bridge} and consecutive content on bridging perspective.
 {\protect\NoHyper
\begin{figure*}[t]
	 \definecolor{historybg}{HTML}{D7DBDF}
\definecolor{scibg}{HTML}{EEEEEE}

\begin{center}
\resizebox{0.95 \textwidth}{!}{%
\tikzstyle{level 2 concept}+=[sibling angle=90]
\begin{tikzpicture}[every annotation/.style = {draw,
                     fill = white, font = \normalsize}]
  \path[mindmap, every node/.style={concept,execute at begin node=\hskip0pt},
root concept/.append style={
concept color=black, fill=white, line width=1ex, text=black,minimum size=4cm,text width=3cm,font=\Large\scshape},
level 1 concept/.append style={font=\normalsize},
level 2 concept/.append style={font=\small},
concept color=historybg,grow cyclic]

	node[root concept](CL){Deep Continual Learning}[clockwise from=-120]
      child[concept color=scibg] {
        node[concept] {Regularization} [clockwise from=-90]
          child {node[concept] (ParameterRegularization){Structural}}
          child {node[concept] (DataBasedRegularization){Functional}}}
      child[] {
      	node[concept] {Replay} [clockwise from=-135]
      	  child {node[concept] (Generative){Generative}}
          child {node[concept] (Exemplar){Exemplar Rehearsal}}}
      child[concept color=black!20] {    
      	node[concept] {Architectural} [clockwise from=-180]
          child {node[concept] (FixedCapacity){Fixed Capacity}}
          child {node[concept] (DynamicGrowth){Dynamic Growth}}}
      child[concept color=black!30] {    
      	node[concept] (CombinedCL){Combined Approaches} [clockwise from=180] }
                    
    node[root concept](OSR) at (0:8) {Open Set Recognition}[clockwise from=60]
      child[concept color=scibg] {    
      	node[concept] (MetaRecognition){Meta-recognition} [clockwise from=0]}
      child[] {
        node[concept] (PriorKnowledge){Prior Knowledge} [clockwise from=45]}
      child[concept color=black!20] {
      	node[concept] {Predictive Anomalies} [clockwise from=30]
           child {node[concept] (Thresholding){Thresholding}}
           child {node[concept] (Uncertainty){Uncertainty}}}

     node[root concept](AL) at (-60:8) {Active Learning}[clockwise from=-30]
      child[concept color=scibg] {    
      	node[concept] (ALuncertainty){Uncertainty Heuristics} [clockwise from=30]}
      child[] {
        node[concept] (Representation){Representations} [clockwise from=45]}
      child[concept color=black!20] {
      	node[concept] (Versionspace){Version Space} [clockwise from=-90]};
         
	\info[15]{ParameterRegularization.west}{above,anchor=north west,xshift=-12.5em,yshift=5em}{%
      \item \emph{EWC} \citep{Kirkpatrick2017}
      \item \emph{IMM} \citep{Lee2017}
      \item \emph{SI} \citep{Zenke2017}
      \item \emph{RWalk} \citep{Chaudhry2018}
      \item \emph{MAS} \citep{Aljundi2018}
      \item \emph{ALASSO} \citep{Park2019}
      \item \emph{UCL} \citep{Ahn2019}
      \item \emph{UCB} \citep{Ebrahimi2019}
    }      
    
    \info[14]{DataBasedRegularization.west}{above,anchor=north west,xshift=-12em,yshift=1.75em}{%
      \item \emph{LWF} \citep{Li2016}
      \item \emph{EBLL} \citep{Rannen2017}
    }      
    
    \info[21]{Generative.west}{above,anchor=north west,xshift=-17em,yshift=4.5em}{%
      \item \emph{Pseudorehearsal} \citep{Robins1995}
      \item \emph{Pseudo-recurrent nets} \citep{French1997}
      \item \emph{Reverberating NNs} \citep{Ans1997}
      \item \emph{DGR} \citep{Shin2017}
      \item \emph{RfF} \citep{VandeVen2018}
      \item \emph{OpenVAE} \citep{Mundt2022}
      \item \emph{ILCAN} \citep{Xiang2019}
    }      
    
    \info[20]{Exemplar.west}{above,anchor=north west,xshift=-16.5em,yshift=3em}{%
      \item \emph{GeppNet} \citep{Gepperth2016}
      \item \emph{GEM} \citep{Lopez-Paz2017}
      \item \emph{SER} \citep{Isele2018}
      \item \emph{CLEAR} \citep{Rolnick2018}
      \item \emph{A-GEM} \citep{Chaudhry2019}
      \item \emph{BiC} \citep{Wu2019}
    }    
    
    \info[18]{FixedCapacity.west}{above,anchor=north west,xshift=-15em,yshift=3.5em}{%
      \item \emph{Activation sharpening} \citep{French1992} 
      \item \emph{PathNet} \citep{Fernando2017}
      \item \emph{HAT} \citep{Serra2018}
      \item \emph{Piggyback} \citep{Mallya2018}
      \item \emph{UCB-P} \citep{Ebrahimi2019}
    }      
    
    \info[16]{DynamicGrowth.west}{above,anchor=north west,xshift=-13.5em,yshift=4.25em}{%
      \item \emph{DNC} \citep{Ash1989}
      \item \emph{PNN} \citep{Rusu2016}
      \item \emph{ExpertGate} \citep{Aljundi2017}
      \item \emph{NDL} \citep{Draelos2017}
      \item \emph{DEN} \citep{Yoon2018}
      \item \emph{RCL} \citep{Xu2018}
      \item \emph{Learn-to-Grow} \citep{Li2019}
    }    
      
    \info[17]{CombinedCL.east}{above,anchor=north east,xshift=8em,yshift=13em}{%
      \item \emph{iCarl} \citep{Rebuffi2017}
      \item \emph{MRGAN} \citep{Wu2018}
      \item \emph{VCL} \citep{Nguyen2018}
      \item \emph{VGR} \citep{Farquhar2018}
      \item \emph{VASE} \citep{Achille2018}
      \item \emph{FearNet} \citep{Kemker2018a}
      \item \emph{LLRNN} \citep{Sodhani2019}
      \item \emph{LLGAN} \citep{Zhai2019}
    }        
    
    \info[19]{MetaRecognition.east}{above,anchor=north east,xshift=15.75em,yshift=3.75em}{%
      \item \emph{CAP} \citep{Scheirer2014}
      \item \emph{OpenMax} \citep{Bendale2016}
      \item \emph{Mahalanobis} \citep{Lee2018}
      \item \emph{OWR-Survey} \citep{Boult2019}
      \item \emph{CROSR} \citep{Yoshihashi2019}
      \item \emph{Latent based EVT} \citep{Mundt2019} 
    }        
    
    \info[20]{PriorKnowledge.east}{above,anchor=north east,xshift=16.5em,yshift=3.25em}{%
      \item \emph{Universum Inference} \citep{Weston2006} 
      \item \emph{Confidence Calibration} \citep{Lee2018a}
      \item \emph{Objectosphere Loss} \citep{Dhamija2018}
      \item \emph{SCM} \citep{Feng2019}
      \item \emph{Discrepancy Loss} \citep{Yu2019}
    }        
    
    \info[20]{Thresholding.east}{above,anchor=north east,xshift=16.5em,yshift=2.75em}{%
      \item \emph{Softmax-Confidence} \citep{Matan1990}
      \item \emph{TCM-kNN} \citep{Li2005}
      \item \emph{Hinge Loss} \citep{Bartlett2008}
      \item \emph{Confidence} \citep{Hendrycks2017}
      \item \emph{ODIN} \citep{Liang2018}
      \item \emph{OCGAN} \citep{Perera2019}
    }        
    
    \info[23]{Uncertainty.east}{above,anchor=north east,xshift=18.75em,yshift=2.75em}{%
      \item \emph{BayesSegNet} \citep{Kendall2017}
      \item \emph{Image Resynthesis} \citep{Lis2019}
      \item \emph{Deep Generative Models} \citep{Nalisnick2019}
      \item \emph{Uncertainty under Dataset Shift \\} \citep{Ovadia2019}
    }        
    
     \info[19]{ALuncertainty.east}{above,anchor=north east,xshift=15.75em,yshift=3.25em}{%
      \item \emph{Entropy, maximum discrimination \\ between two models} \citep{MacKay1992}
      \item \emph{Confidence} \citep{Lewis1994}
      \item \emph{Query by committe} \citep{Seung1992}
      \item \emph{Ensembles} \citep{McCallum1998}
      \item \emph{BALD} \citep{Gal2017}
      \item \emph{Deep Ensembles} \citep{Beluch2018}
      \item \emph{iNAS} \citep{Geifman2019}
      \item \emph{BGAL} \citep{Tran2019}
    }     
    
    \info[26]{Representation.west}{above,anchor=north east,xshift=0.25em,yshift=1.75em}{%
      \item \emph{K-medoids pre-clustering} \citep{Nguyen2004}
      \item \emph{MEB-SVM} \citep{Tsang2005}
      \item \emph{Gaussian process information density} \citep{Li2013}
      \item \emph{GAAL} \citep{Zhu2017}
      \item \emph{Deep coreset AL} \citep{Sener2018}
      \item \emph{Efficient cGAN AL} \citep{Mahapatra2018}
      \item \emph{VAAL} \citep{Sinha2019}
      \item \emph{WAAL} \citep{Shui2020}
      \item \emph{ASAL} \citep{Mayer2020}
    }     
    
    \info[26]{Versionspace.west}{above,anchor=north west,xshift=-21em,yshift=1.5em}{%
      \item \emph{GMM} \citep{Cohn1996}
      \item \emph{Naive Bayes} \citep{Roy2001}
      \item \emph{SVM margin} \citep{Tong2001}
      \item \emph{Multi-class SVM margin} \citep{Joshi2009} 
      \item \emph{Meta-learning active learning} \citep{Konyushkova2017}
    }    
    
    \begin{pgfonlayer}{background}
    \draw [circle connection bar]
      (CL) edge[fill=white] node {\Large\scshape } (OSR) edge[fill=white] node {\Large\scshape } (AL)
      (AL) edge[fill=white] node {\Large\scshape } (OSR);
  \end{pgfonlayer}
    
\end{tikzpicture}%
}
\end{center}
	 \vspace*{-0.5cm}
\caption{\label{fig:literature_diagram} Visual taxonomy of neural network based methods for continual learning, active learning and open set recognition.}
\end{figure*}
\protect\endNoHyper}

\subsection{Continual learning}\label{sec:continual_review}
As indicated in the introductory section, continual learning should ideally encompass a variety of research questions. Our later sections will continue to argue that currently considered scenarios are too reductive, resulting in potential difficulty to chose among existing algorithmic options. For now, we will start with a typical categorization of existing deep continual works into the three categories of \textit{regularization, rehearsal} and \textit{architectural} approaches, in consistency with recent reviews \citep{Parisi2019,DeLange2019,Lesort2020}. We note that a strict organization into these groups is not always possible and hence also provide a forth category for works that combine multiple methods. In later sections we will argue that this is not only advantageous, but conceivably a necessity. 

\subsubsection{Regularization}
Continual learning approaches based on regularization aim to strike a balance between protecting already learned representations, while granting sufficient flexibility for new information to be encoded. Intuitively, a meaningful balance should be attainable for tasks with sufficient overlap in their high dimensional embeddings, i.e. if a considerable amount of the learned representations are shareable. Existing approaches can be further subdivided into two subgroups of regularization. One of these explicitly protects parameters by constraining changes on every level of a model architecture, which we refer to as \textit{structural}. The other preserves a model's output for seen tasks while ensuring full adaptability with respect to each individual model stage that leads to the prediction, which we refer to as \textit{functional}. 

\paragraph*{Structural:} Structural regularization approaches draw inspiration from the neuroscientific stability-plasticity dilemma \citep{Hebb1949}. That is, successful use of regularization of deep learning models for continual learning requires carefully balancing the trade-off between overwriting acquired representations in favor of sensitivity to new information and preservation of already existing formed patterns. Elastic Weight Consolidation (EWC) \citep{Kirkpatrick2017} aims to achieve this balance by estimating each parameter's importance through the use of Fisher information and respectively discouraging updates for parameters with greatest task specificity. Synaptic Intelligence (SI) \citep{Zenke2017} and Memory Aware Synapses (MAS) \citep{Aljundi2018}, where the biologically inspired term synapse is used synonymously with parameter, follow a similar approach by explicitly equipping each parameter with additional importance measures that keep track of past improvements to the objective. Asymmetric Loss Approximation with Single-Side Overestimation (ALASSO) \citep{Park2019} can be seen as a direct extension to SI and aims to mitigate its limitations by introducing an asymmetric loss approximation that is motivated from empirical observations. Riemannian Walk (RWalk) \citep{Chaudhry2018} has generalized EWC and SI by taking into account both the Fisher information based importance. The latter is based on a perspective of computing distances in the induced Riemann manifold, and the optimization trajectory based importance score. Incremental Moment Matching (IMM) \citep{Lee2017} approaches structural regularization from a perspective of Bayesian approximations and matching the moments of tasks' posterior distributions. Uncertainty based Continual Learning (UCL) \citep{Ahn2019} makes use of Bayesian uncertainty estimates to adaptively regularize weights online. Similarly, Uncertainty-guided Continual Bayesian Neural Networks (UCB) \citep{Ebrahimi2019} adapts the learning rate in dependence on the uncertainty defined in the probability distribution of the weights.

\paragraph*{Functional:}
Functional regularization approaches are generally inspired by knowledge distillation \citep{Hinton2014}, an approach originally proposed for model compression. A distillation loss is introduced by storing the prediction of a data sample for future use as a so called soft target. In learning without forgetting (LWF) \citep{Li2016} for class incremental continual learning, the soft targets for existing classes are calculated using newly arriving data. The hope lies in regularizing towards preserving the output for old tasks, even if these predictions might be nonsensical as the freshly added classes do not get correctly predicted yet. Encoder based lifelong learning (EBLL) \citep{Rannen2017} applies this concept to the unsupervised learning scenario by applying distillation to autoencoder reconstructions. Knowledge distillation rarely seems to be employed in isolation, but as will be apparent from the list of upcoming combined approaches is a popular technique in conjunction with other mechanisms.

\subsubsection{Rehearsal} 
As the name implies, rehearsal techniques for continual learning aim to preserve encoded information by replaying data from already seen tasks. Trivially, continual learning could be solved by simply storing and replaying all seen data, albeit at usually intolerable memory expense and growing computation time. Accordingly, a core aspect of rehearsal methods is to find a suitable subset of data that best approximates the entire observed data distribution. This is commonly referred to as selection of exemplars or construction of a core set. Alternatively, a generative modeling approach can be used to generate instances from a learned latent representation, as an encoding of the observed data distribution. Most replay techniques indicate their inspiration to be drawn from the complex biological interplay between hippocampus and neocortex (often referred to as complementary learning systems), wake + sleep cycles and dreaming in the brain \citep{McClelland1995,Kumaran2016}.  

\paragraph*{Exemplar Rehearsal:}
GeppNet \citep{Gepperth2016} explores the use of a dual-memory system that implements various short and long-term memory storages. These serve the purpose of storing newly arriving information or providing dedicated replay cycles of previously stored data. 
Selective Experience Replay (SER) \citep{Isele2018} concentrates on exemplar selection techniques and investigates trade-offs between preferring surprising experiences over rewarding ones, or maximizing distribution coverage. Gradient Episodic Memory (GEM) \citep{Lopez-Paz2017} extends the use of a memory that gets replayed episodically with constraints on the gradients to be non-conflicting with updates for previous tasks. A respective extension called Averaged Gradient Episodic Memory (A-GEM) \citep{Chaudhry2019} has introduced significant improvements on computational and memory cost for optimization under these constraints.
CLEAR \citep{Rolnick2018} uses experience replay together with off-policy learning to preserve old information and on-policy learning to learn new experiences in deep reinforcement learning. Bias Correction (BiC) \citep{Wu2019} rehearses exemplars and additionally corrects for biases in the classification layer.

\paragraph*{Generative:}
Generative replay is a specific version of rehearsal where the data to be rehearsed consists entirely of instances sampled from a generative model. Rather than making use of an episodic memory of previously seen data, generated samples of former tasks are typically interleaved with the current task's real data during training. The most elementary version of this procedure was coined pseudo-rehearsal \citep{Robins1995}, where the generative model is of simple nature. Here, binary patterns are sampled at random, their target value or label computed given the current state of the classifier, and the classifier then needs to maintain the discrimination on these patterns and learn new classes. Such pseudo-rehearsal has then successfully been leveraged in brain-inspired dual-memory architectures that use two distinct networks for acquisition and storage of information with generative rehearsal to consolidate the memory. Two early examples include pseudo recurrent networks \citep{French1997} and coupling two reverberating neural networks \citep{Ans1997}. 
Deep Generative Replay (DGR) \citep{Shin2017} have introduced a deep learning variant of this practice, where the generative model is taken to be a separate generative adversarial network \citep{Goodfellow2014} that gets trained in alternation with a classification model. Replay through Feedback (RfF) \citep{VandeVen2018} proposed generative replay using a single model that handles both classification and generation through the aid of feedback connections. Incremental learning using conditional adversarial networks (ILCAN) \citep{Xiang2019} follows a similar approach of using a single model, but additionally changes the generative replay component to rehearse feature embeddings instead of aiming at reconstructing original input data. 
Open Variational Auto-Encoder (OpenVAE) \citep{Mundt2022} further introduces the first approach to naturally integrate open set recognition with deep generative replay in a single architecture. This work will be used as an example in the empirical portion of our paper. We will demonstrate how suggested ideas can be extended to form one potential basis as means to broaden current continual learning practices.

\subsubsection{Architectural}
Architectural approaches attempt to alleviate catastrophic forgetting through modification of the underlying architecture. It might at this point be baffling to the reader why such modifications are listed distinctly from the works presented in previous subsections. They are almost by definition complementary to any method presented so far, and in fact most methods presented in this paper. For historical reasons, we will however stay consistent with former categorization of deep continual learning algorithms \citep{Parisi2019}. We further sub-categorize architectural approaches into implicit and explicit architecture modification, i.e. methods that use a fixed amount of representational capacity and methods which dynamically increase capacity in the process of continued training. 

\paragraph*{Fixed maximum representational capacity:}
Approaches that use a static architecture rely on task specific information routing through the architecture. An early example is a technique coined activation sharpening towards semi-distributed representations \citep{French1992}. Here, the essence is to tune and limit the amount of high neural network activations to a maximum of k nodes, such that there is less activation overlap for different representations. Consequently, there is less potential for interference of new examples. While fixed architecture methods differ in the specifically employed technique to disambiguate the learned dense representations, the common denominator is the assumption of an over-parametrized architecture. The latter is needed in order to warrant enough initial redundancy to permit overriding parameters without incurring catastrophic interference. PathNet \citep{Fernando2017} adopted this notion to deep neural networks and used a genetic algorithm to determine and freeze pathways that are deemed particularly useful for a specific task. Instead of using a separate algorithmic layer to determine task specific  subsets, Piggyback \citep{Mallya2018} and hard attention to the task (HAT) \citep{Serra2018} directly learn binary masks and use them to gate information propagation through the network. The UCB-P variant of the earlier introduced regularization approach Uncertainty-guided Continual Bayesian Neural Networks (UCB) \citep{Ebrahimi2019} confronts this challenge from a Bayesian perspective. They use uncertainty to prune the model and identify binary masks per task to index into the weights' Gaussian mixture distributions.

\paragraph*{Dynamic growth:}
Dynamic growth approaches administer representational capacity much more explicitly. The trivial solution would be to simply have one model per task and devise a mechanism to select the appropriate path for an input. Alas, such an arrangement doesn't fully leverage information from one task to positively transfer to another or newly arriving information to aid already acquired tasks respectively. First works in deep learning however nearly follow this naive, but also intuitive, approach to simply train on a task and consequently freeze all learned representations, such as demonstrated in Progressive Neural Networks (PNN) \citep{Rusu2016}. The amount of weights is then increased for a new task, with the twist that formerly learned representations laterally transmit their output to the new tasks' representations but not vice versa. Expert Gate \citep{Aljundi2017} is comparable and differs mainly in the introduction of a gating mechanism that automates the choice of a suitable expert in an ensemble. Recent, perhaps more practical, approaches can be viewed as once again drawing their inspiration from decades of biological findings and discussion on neurogenesis. The latter refers to the process of creation and incorporation of new neurons into the existing system, see the reviews by \citet{Aimone2014,Vadodaria2014}. For the last two decades it has now been acknowledged that this process persist beyond early stage human development and continues its function in adults \citep{Gross2000}. The seminal work of dynamic node creation in neural networks \citep{Ash1989}, where additional units are added whenever the loss plateaus, has thus found a renaissance in modern deep learning. Neurogenesis deep learning to accommodate new classes (NDL) \citep{Draelos2017} and lifelong learning with Dynamically Expandable Networks (DEN) \citep{Yoon2018} have adapted this heuristic approach for use in continual deep learning. The former by adding units whenever the reconstruction error of an autoencoder surpasses a predetermined threshold in the spirit of \cite{Zhou2012}, the latter based on an empirically found value of the classification loss in supervised learning. Reinforced Continual Learning (RCL) \citep{Xu2018} or Learn-to-Grow \citep{Li2019} further attempt to overcome the challenge of finding suitable loss cut-offs and cast dynamic unit addition into a meta-learning framework in order to separate the learning of the network structure and estimation of its parameters. 

\subsubsection{Combined Approaches}
A number of works have primarily advanced the state of the art on a set of benchmark datasets by blending techniques from the previous categories. We list some popular works in this category that have attempted such a blend for the first time, but also note that the amount of newly emerging combinations grows very rapidly. One of the most popularly cited works is iCarl \citep{Rebuffi2017}, which couples a knowledge distillation based regularization approach with rehearsal of exemplars, assembled through a greedy herding procedure \citep{Welling2009}. Variational Continual Learning (VCL) \citep{Nguyen2018} similarly fuses use of an episodic memory of exemplars with parameter regularization, but from a perspective of approximate Bayesian inference. FearNet \citep{Kemker2018a} has later criticized iCarl as a viable technique due to its heavy dependency on quantity of data. They have therefore additionally incorporated generative rehearsal to compensate the need to store large subsets of the original dataset. Variational Generative Replay (VGR) \citep{Farquhar2018} can be seen as concurrent to VCL, where instead of exemplar rehearsal generative replay is made use of. Memory replay GAN (MRGAN) and Lifelong GAN (LLGAN) \citep{Zhai2019} are more recent complements to these works and deviate in that they are based on GANs instead of variational inference in autoencoders. Whereas MRGAN uses a functional regularization approach to align the generator's output, LLGAN further applies such distillation loss based regularization across multiple places in the architecture to regularize encoders and discriminators. 
On the architectural front, Variational Autoencoder with Shared Embeddings (VASE) \citep{Achille2018} adopts dynamic architecture growth in conjunction with generative replay. Their proposal is to allocate additional representational capacity for new concepts, determined through larger reconstruction loss in a variational autoencoder, however, is limited to expanding the latent space and leaving the rest of the architecture static. 
Lifelong Learning for Recurrent Neural Networks (LLRNN) \citep{Sodhani2019} combines training of long short-term memory (LSTM) \citep{Hochreiter1997} with gradient episodic memory based exemplar rehearsal and a capacity expansion approach named Net2Net \citep{Chen2016}. The approach provides the means to transfer learned representations from an architecture to a larger untrained one before continuing to train the latter. 
While some of these works clearly exploit natural synergies, a generally desirable practice, we note that this can sometimes come at the expense of detailed analysis and comprehensive understanding of individual key ingredients. Therefore, we agree that all approaches in this subsection pursue commendable directions, but also wish to point out that considerable future analysis is still required. 

\subsection{Active learning}\label{sec:active_review}
\begin{figure}[t]
	\include{AL_diagram}
	\vspace*{-0.5cm}
	\caption{Active learning cycle that repeatedly expands a labelled dataset by querying and then annotating data instances from a larger unlabelled pool. The dashed arrow from the latter to the training process indicates the common closed world active learning scenario, where the presence of all data at all times is assumed. Respective works typically include the entire unlabelled dataset into the training procedure by employing methods from semi-supervised learning. Shaded parts of the diagram correspond to processes, whereas light components represent objects.
	\label{fig:AL_diagram}}
\end{figure}

Rather than focusing on the question of how to preserve representations in incremental continual learning, the topic of active learning asks the reverse question of how to pick data increments for future inclusion. Generally, this is cast into the framework of semi-supervised learning. Here, it is assumed that the model is trained on labelled data $\boldsymbol{X}_{L} = \{ \boldsymbol{x}_{L}^{1}, \ldots, \boldsymbol{x}_{L}^{n}\}$, and a larger pool of unlabelled data $\boldsymbol{X}_{U}$ exists. This is motivated from data acquisition being relatively cheap in the modern world, as opposed to human intensive data labelling that often requires highly skilled experts. The task of an active learner is thus to extract a set of $M$ data instances $\{ \boldsymbol{x}_{U}^{1}, \ldots, \boldsymbol{x}_{U}^{m}\}$ from the pool of unlabelled data, such that a maximum gain in performance on the inspected task is expected if a human in the loop provides the additional labels $\{y^{1}, \ldots, y^{m}\}$ for further training. The underlying mechanism on which the query is based is referred to as the acquisition function and forms the main pillar of active learning research. We have visualized this active learning cycle in Figure \ref{fig:AL_diagram}. 

There are multiple conceivable evaluation variants to gauge the usefulness of active learning acquisition function choices. They either explicitly assume the entirety of the unlabelled data to be accessible and usable upfront, or contrarily the query being informed solely by the available labelled data. Independently of the latter, the practical assessment of active learning strategies is generally conducted in a closed world scenario. That is, the entire pool of unlabelled data is expected to stem from the same data distribution as the initially labelled set. The oracle is respectively assumed to be infallible. In a crucial distinction to continual learning, evaluation of active learning however accumulates data and grows the labelled set, focusing primarily on the cost reduction of labour intensive annotation. In consequence, an active learner is deemed successful if each data query provides significant benefit over simply picking and labelling data at random. 

``A probability analysis of the value of unlabelled data for classification problems'' \citep{Zhang2000} provides an early analysis of the requirements for benefiting from semi-supervised or active learning approaches. The authors consider two types of models: parametric $p(\boldsymbol{x}, y | \boldsymbol{W}) = p(\boldsymbol{x} | \boldsymbol{W})p(y | \boldsymbol{x}, \boldsymbol{W})$ and semi-parametric: $p(\boldsymbol{x}, y | \boldsymbol{W}) = $ \\ $p(\boldsymbol{x}) p(y | \boldsymbol{x}, \boldsymbol{W})$. In the latter, the data probability $p(\boldsymbol{x})$ is decoupled and can have an unknown (or non-parametric) form independent of the weights $\boldsymbol{W}$, as is common in most discriminative models such as logistic regression or most neural networks. They argue that these models are particularly suited for active learning, as opposed to parametric models such as Gaussian mixtures being particularly suitable for semi-supervised learning. This is because they do not need to rely on potentially inaccurate estimates of the entire data distribution when only a fraction of the data is observable. However, we will see in the subsequent review that both of these model types have been used to form different perspectives to address active learning and come with their respective advantages. 
	
As with the majority of techniques, early active learning methods have rapidly cross-pollinated into applications with deep neural networks. However, due to the black-box nature of deep non-linear neural networks, many of these approaches are based on simple heuristics or approximations to uncertainty quantities that no longer have tractable closed-form solutions. We will start with these heuristic approaches, as they are often trivial to transfer to deep learning. We then continue to summarize more principled approaches, which can turn out to be genuinely challenging in the context of deep learning with neural networks.

\subsubsection{Uncertainty Heuristics}
One theoretically sound approach to querying useful data is based on entropy \citep{Shannon1948} sampling and other information theoretic acquisition functions \citep{MacKay1992}. An early approach based on training two neural networks to estimate query areas in binary classification problems \citep{Atlas1990} remarks that this is difficult for neural networks as they are often overly confident in their outputs. This overconfidence is going to be one of the main subjects of our next major section on learning in an open world. Interestingly, while carefully studied in early literature in isolation, this aspect seems to often be overlooked in the era of deep learning, particularly when placed in the context of continual and active learning. Here, simply using neural network prediction confidence, predictive entropy or other derived heuristics \citep{Lewis1994} are still practically employed in comparisons today \citep{Geifman2019}. 
This is because many approaches have been shown to empirically work well in specific contexts, although there is no guarantee for them to succeed. Early works have shown uncertainty sampling based active learning for logistic regression \citep{Lewis1994} and neural networks \citep{Seung1992,McCallum1998} based on ``query by committee'', an approach to estimate uncertainty by using an ensemble of neural networks. This idea has later found a one-to-one translation to deep ensembles for active learning \citep{Beluch2018}. Naturally, most black-box deep neural networks are not equipped with mechanisms to gauge uncertainty properly outside of using multiple parallel models. Bayesian active learning by disagreement (BALD) therefore provides an attempt at avoiding the necessity of ensembles and instead uses Monte Carlo Dropout \citep{Gal2015,Srivastava2014} to calculate points of high variance in the output \citep{Gal2017}. This has empirically been demonstrated to be effective and has been extended in Bayesian Generative Active Learning (BGAL). Here, BALD is used to query samples and then the labelled set is further augmented with generated examples \citep{Tran2019}. Deep incremental learning with Neural Architecture Search (iNAS) \citep{Geifman2019} does not propose a new query mechanism and instead provides an evaluation of above acquisition functions in the context of architecture selection. They include the option of progressive architecture growth after each query, to illustrate that small models generally fare better in a small data regime, whereas large models are required when a certain degree of task complexity is reached. 

\subsubsection{Version Space and Expected Error Reduction:}
A theoretically more substantiated approach to basing the acquisition function on heuristics is to query data that provably reduces the expected error. Clearly, such proof is beyond the current understanding of deep neural networks, but has been shown to be feasible in the context of parametric models such as Gaussian mixture models \citep{Cohn1996} or naive Bayes \citep{Roy2001}. These works use the formal concept of a version space \citep{Mitchell1982}. At the example of classification, its respective definition is the set of all hypotheses that are consistent with the observed data in achieving a possible separation in the induced feature space. An appropriate active learning strategy is to sequentially and monotonically reduce the size of this version space, i.e. shrink the amount of conceivable hypotheses. In models such as SVMs for binary classification this can intuitively be explained based on the margins \citep{Tong2001}. Here, new points are chosen according to hyperplanes that maximize the restriction with respect to the set of possible hyperplanes for correct classification. The latter was later extended to a multi-class SVM based approach \citep{Joshi2009}, however still based on multiple binary classifiers. Above efforts allowed for theoretical guarantees on sample complexity and necessary amount of queries to be analyzed with respect to these binary classification problems with linear decision boundary in the context of greedy active learning strategies \citep{Dasgupta2005}. Whereas ``learning active learning from data'' \citep{Konyushkova2017} provides a recent effort to train a meta-learning based regressor to predict expected error reduction for binary classification using random forests, the idea is yet to be broadly adapted to deep neural networks. First efforts at scale based on approximations to expected model output changes are presented in \cite{Kaeding2016b,Kaeding2016a}.  

\subsubsection{Representation based approaches:} 
Although version space reduction can come with provable guarantees, respective application to deep neural networks is inconceivable before a mature theory of how their hypotheses are formed has evolved. At the same time, Roy et al. \citep{Roy2001} have pointed out that the earlier summarized uncertainty sampling, or estimates thereof through ensembles, are generally insufficient. They argue that they are prone to querying outliers, as a result of sampled instances being viewed in isolation and without regarding the underlying density of the full data distribution. Similar conclusions were empirically observed in the large scale empirical evaluation of active learning for text applications \citep{Settles2008}. As a solution, the authors suggest a representation based information density measure. Although heavy to compute, it implicitly takes into account the underlying data distribution. This can be seen as an approach that is orthogonal to minimizing the version space. Typically the distribution coverage on the entire dataset according to the model representations is now maximized instead of reducing the number of possible hypotheses. The often necessary core assumption is thus the presence of the entire unlabelled pool of data and its auxiliary use in optimization of the labelled set. We have attributed our third category to approaches that follow this objective.

Active learning using pre-clustering \citep{Nguyen2004} uses a k-medoids algorithm in conjunction with a SVM or logistic regression to select data from the pre-clustered embedding of the unlabelled pool. Similarly, SVM based core vector machines \citep{Tsang2005} use a set of minimum enclosing balls to create a core set that best approximates the entire distribution. Li et al. estimate information density by using the unlabelled data in a Gaussian process \citep{Li2013}. The idea in these works have since been abstracted to deep neural networks. \citet{Sener2018} base their active learning procedure on construction of core sets based on a k-medians algorithm. \citet{Shui2020} achieve distribution coverage by matching distributions through minimization of the Wasserstein distance in Autoencoders (WAAL). Variational adversarial active learning (VAAL) \citep{Sinha2019} approximates the data distribution by learning the latent space in a variational autoencoder \citep{Kingma2015} and simultaneously trains a latent based adversarial network to discriminate between unlabelled and labelled data. 

In complement to these works, various query-synthesizing methods have been proposed \citep{Zhu2017,Mahapatra2018,Mayer2020}. Here, the challenge of active learning is tackled by using a deep generative model to generate informative queries. Instead of querying from an unlabelled pool directly, generative adversarial active learning (GAAL) \citep{Zhu2017} and ``efficient active learning using conditional generative adversarial network'' (Efficient cGAN AL) \citep{Mahapatra2018} both train GANs to synthesize and label queries. The core assumption is the ability to adequately capture the data distribution to generate meaningful instances. The usefulness of the generated samples with respect to a classifier can then either be assessed through uncertainty heuristics or by matching the synthesized data with samples from the pool and retrieving the most similar instance, as demonstrated in Adversarial Sampling for Active Learning (ASAL) \citep{Mayer2020}. 

In our upcoming discussion, we will argue that the assumption of upfront presence of all data should, and in fact can be lifted when a natural bridge to the other paradigms is constructed. Before that, we first proceed to conclude our last leg of the review by delving into what will later constitute the ``glue'' in our wholistic perspective: learning in an open world and open set recognition.

\subsection{Open set recognition}\label{sec:osr_review}
The term open set recognition was formally coined only recently \citep{Scheirer2013,Bendale2015}. However, its foundation and associated challenge in neural networks dates back to at least several decades before, when discriminative neural networks were found to yield overconfident mispredictions on unseen unknown data \citep{Matan1990}. To get an intuitive understanding, let us briefly consider the types of data we can expect our model to encounter. As soon as we move beyond the closed world benchmark scenario, we can no longer expect our trained models to be tested exclusively on some held-out data from the same distribution as observed during training. In the earlier introduced transfer learning parlance, for prediction, data can thus generally not be presumed to originate from the same domain. We can now distinguish three types of possible inputs to our model \citep{Scheirer2013}:
\begin{enumerate}
\item \textbf{Knowns}: examples belonging to the distribution from which the training set was drawn. The model's prediction is accurate and confident.
\item \textbf{Known unknowns}: unknown instances that a model cannot predict confidently. Examples can optionally be labelled as not being affiliated with the set of known concepts for explicit training of negatives. Prediction uncertainty can indicate a model's awareness of its own limitation.
\item \textbf{Unknown unknowns}: unseen instances belonging to unexplored, unknown distributions or classes for which the prediction is generally overconfident and false.
\end{enumerate}
The broader inspiration for this categorization is commonly attributed to a notorious, machine learning unrelated, quote by Rumsfeld \citep{Naylor2010,Scheirer2013}: \textit{``We know that there are known knowns; these are things we think we know. We also know there are known unknowns; that is to say we know there are some things that we do not know. But there are also unknown unknowns; these are the ones we don't know, we don't know!''}. 
In the context of neural networks, known unknowns can be identified through gauging model uncertainty or relying on derived related heuristics, in correspondence to many of the methods employed in the active learning setting. However, as detailed in a recent survey \citep{Boult2019}, separating the known data from the essentially indistinguishable high-confidence mispredictions for unknown unknowns is far from trivial.

As any machine learning model is trained on a finite dataset, and the imaginable set of unknown unknowns is infinite, we refer to the challenge of recognizing the latter as open set recognition in analogy to prior works \citep{Scheirer2013,Scheirer2014,Bendale2015,Bendale2016,Boult2019}. Formally, these works define the closed space as a union of balls $S_{K}$ that enclose the entire training set $\boldsymbol{X}_{K}$, whereas the open space $\mathcal{O}$ constitutes the remainder of the input or feature space: $\mathcal{O} \subset = \mathcal{X} - \mathcal{S}_{K}$. Correspondingly, works that provide attempts at addressing open set recognition aim to find the respective boundaries between known and unknown spaces \citep{Scheirer2013,Scheirer2014,Bendale2015,Lee2018,Mundt2022,Mundt2019,Yoshihashi2019}. 
We will review these works last in favor of historically preceding approaches based on explicit inclusion of negative classes and rejection through anomalies in prediction patterns, even though the latter have been argued to be insufficient for open set recognition \citep{Matan1990, Scheirer2013, Boult2019}.

The above widespread categorization can technically be extended to encompass a fourth category, by splitting the knowns into \textit{known knowns} and the set of \textit{unknown knowns} \citep{Munro2020}. We do not consider this further distinction as the existence of unknown knowns can be condensed to one of two options: A willfully ignorant false prediction, because we in fact know the concept but choose to nevertheless treat it as unknown. The more charitable alternative in which our chosen machine learning model has an inherent inability to represent the investigated concept and its structure altogether. We also note that there are other related concepts, such as novelty detection \citep{Bishop1994} or equipping classifiers with rejection options. These are different in such that they are typically still evaluated in the closed world and data is generally still expected to reside in a similar domain. The aim is to recognize outliers of the distribution that are uninformative or represent a particularly interesting rare event. Although these works can have considerable merit in their respective closed world application context, we do not review them in favor of the more generic open set recognition, where considered inputs are allowed to be of almost arbitrary nature.
We further note that we naturally cannot provide every example that has ever attempted open set recognition through simple heuristics like using the output values to distinguish examples.

\subsubsection{Prior Knowledge}
A conceivably simple effort to address unknown unknowns is by assuming that the human modeler has enough awareness about what forms of unknown inputs to expect during deployment to directly incorporate this prior knowledge into the model. As inclusion of prior knowledge into neural networks and other types of deep models turns out to be remarkably complex, the natural analogue is to steer efforts towards dataset design. ``Inference with the universum'' \citep{Weston2006} has accordingly proposed to embrace prior knowledge by representing it through a collection of ``non-examples''. Hence, the optimization algorithm decides how to include the presented information into the model. Unfortunately, this does not provide a general solution for open set recognition as upfront knowledge can only ever truly cover the family of known unknowns. At best, a mere work-around for major failure cases is therefore supplied, although without any associated guarantees for remaining unknown unknowns. This lack of guarantees is further enforced by the necessity to rely on machine learning algorithms extracting the information and composing abstractions from the supplied ``non-example'' data population. 

Since then, the idea to include a ``background'' concept has been adopted so widely across applications, that singling out and thus giving preference to select works is difficult. Take as an example large-scale datasets surrounding the task of material classification and semantic segmentation. Because there is an abundance of material types, it has become the de-facto standard to collapse any available imagery that is connected to less important materials or where meager amounts of data are available into a single ``other'' material \citep{Cimpoi2015,Bell2015}. Not only is it impractical to gather data for every material variation, but also unknown unknowns can feature other significant statistical deviations. These could be due to e.g. previously unencountered illumination, acquisition and sensor differences, superposition of dirt and surface markings, or any type of perturbation and previously unencountered noise. Imaginably, in real applications beyond a closed world, inclusion of an endless universe is by definition infeasible. Nevertheless, multiple recent works follow this route. They propose mechanism to calibrate output confidences in deep models \citep{Lee2018a}, formulate a discrepancy loss between knowns and known unknowns \citep{Yu2019} or modify the embedding to explicitly separate them. Examples include semantic categorical and contrastive mapping (SCM) \citep{Feng2019} or the Objectosphere loss \citep{Dhamija2018}. Although these approaches are not tantamount to a comprehensive solution, we note that they can still in principle be sufficient for tasks in partially constrained environments that naturally limit the world's openness. 

\subsubsection{Predictive Anomalies}
From an unsuspecting angle, a model will consistently yield accurate predictions only for observed data and produce highly uncertain output otherwise. Yet, it still generalizes correctly to data that is from the same domain but has not been included in training. In this view, determining a prediction threshold and obtaining an uncertainty estimate is sufficient to recognize any form of unknowns. This can work surprisingly well in models with thorough understanding of the decision boundary and its neighborhood, such as the Transduction Confidence Machine-k Nearest Neighbors (TCM-kNN) \citep{Li2005}. Even though it is well known that the entangled dense representations of neural networks result in overconfident predictions on any data \citep{Matan1990, Boult2019}, a variety of practical approaches nevertheless proposed to simply rely on a hinge loss to reject during classification \citep{Bartlett2008} or even to take the straightforward route and directly trust the softmax confidence \citep{Hendrycks2017}. As the quantitative outcome leaves room for improvement, multiple works have argued that uncertainty estimation is required to corroborate the decision to gain awareness of the unknown. 
\begin{figure}[t]
	\centering
	\includegraphics[width=0.775 \columnwidth]{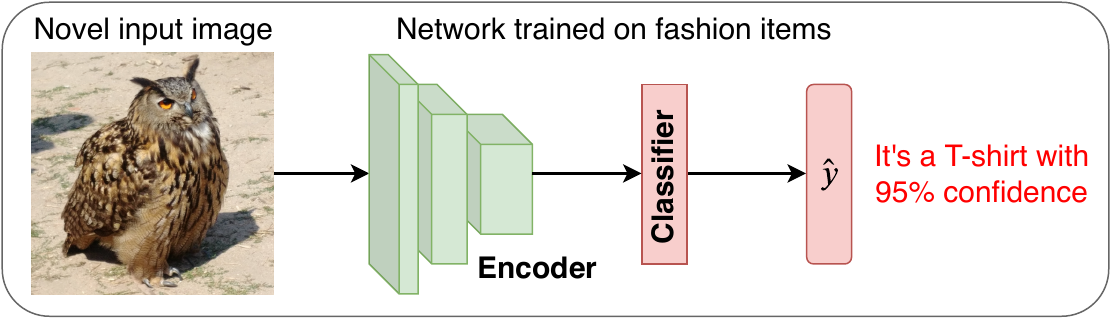} \medskip \\
	\includegraphics[width=0.45 \columnwidth]{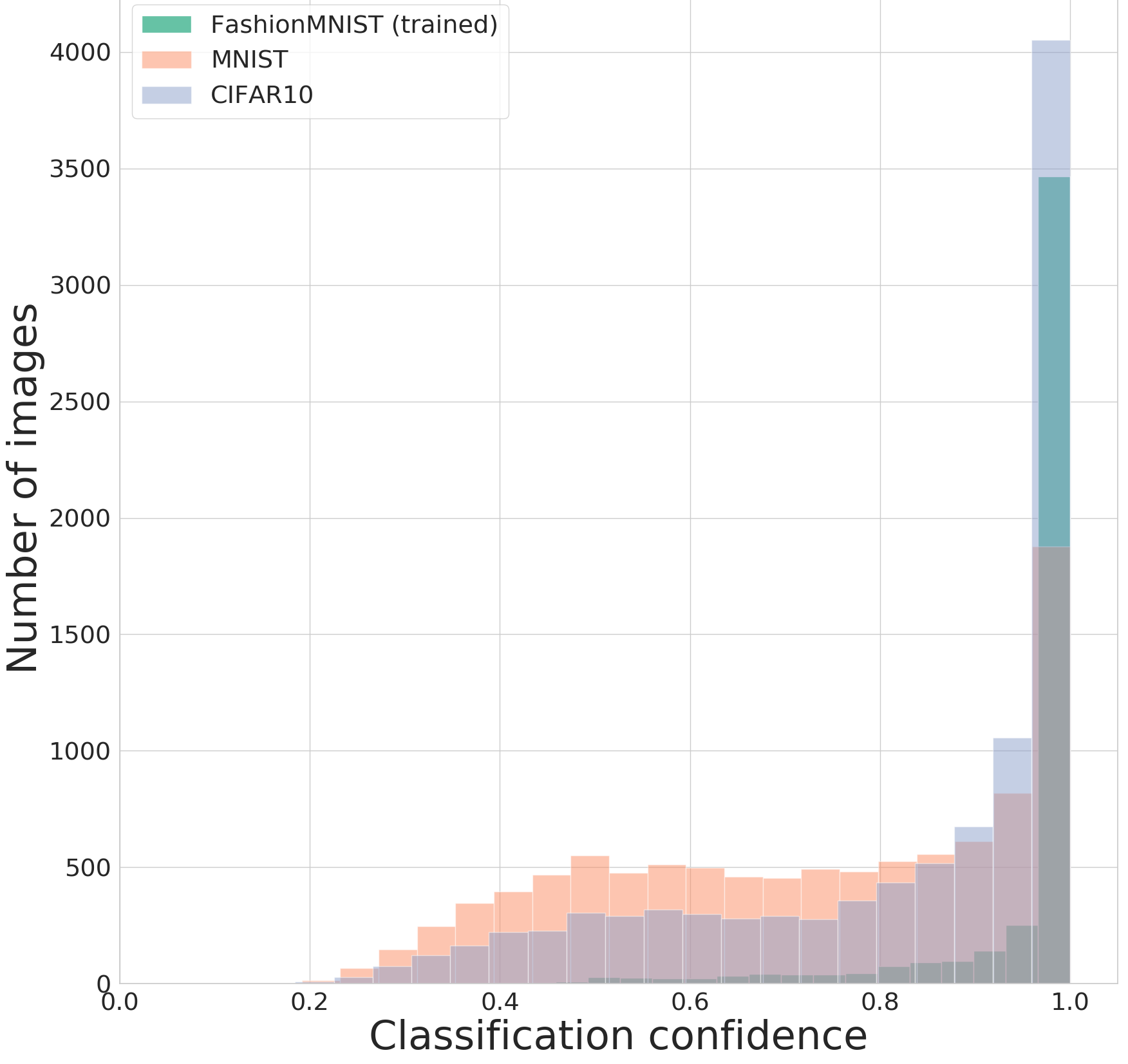} \quad
	\includegraphics[width=0.45 \columnwidth]{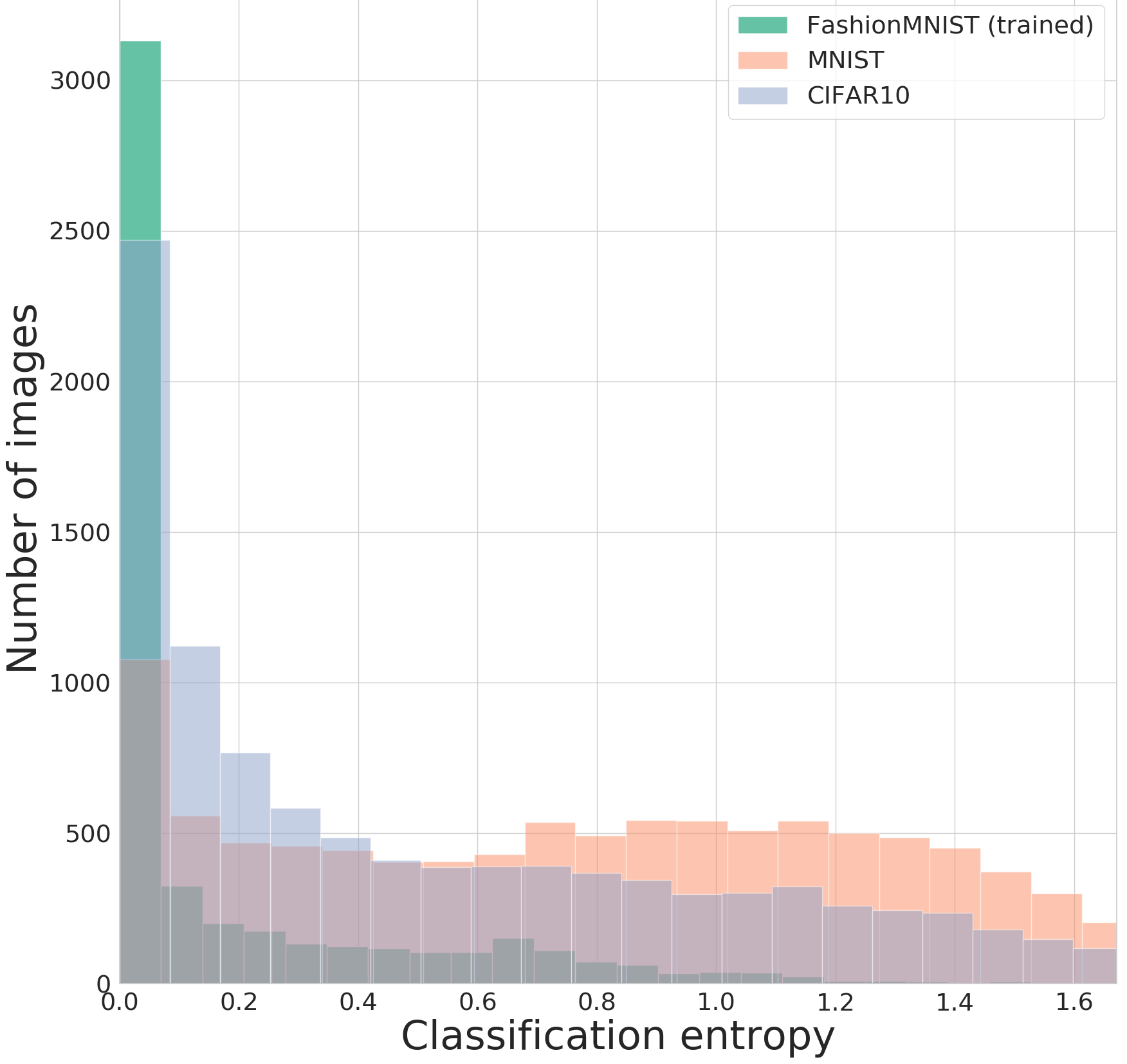} 
	\caption{\label{fig:openset} Top panel: Qualitative illustration of the challenge of open set recognition. A neural network that has been trained to discriminate fashion items misclassifies the unknown concept of an owl and assigns it to the t-shirt class with very high confidence. Bottom panel: A quantitative example of a deep wide residual neural network trained on the FashionMNIST dataset, asked to classify unrelated unencountered digits and objects from the MNIST and CIFAR10 datasets. Even though uncertainty is estimated using 50 Monte Carlo Dropout passes, misclassified unseen data still overlaps significantly with the known dataset in prediction confidence or entropy. Knowns and unknowns are largely indistinguishable. The shown quantitative results are a reproduced subset of our previous work investigating the limits of deep neural network uncertainty for open set recognition \citep{Mundt2019}. Best viewed in color.}
\end{figure}
In deep networks this could be achieved by assessing the variations of stochastic forward passes through a neural network with dropout \citep{Srivastava2014,Kendall2017,Miller2018}, as a variational Bayesian approximation to a distribution on the weights \citep{Gal2015}. Alternatively, one could empirically estimate the output's variability with respect to introduced perturbations, such as done in ODIN (outlier detection in neural networks) \citep{Liang2018}, and calibrate the prediction accordingly \citep{Lee2018a}. In similar spirit, an often employed argument is that generative modeling is required to obtain meaningful prediction values that allow to recognize out of distribution samples. For this purpose, \cite{Lis2019} use image resynthesis and equate detection of unknown concepts with identification of discrepancies in poorly reconstructed image regions. Likewise, one-class novelty GAN (OCGAN) \citep{Perera2019} generates examples from sparsely populated latent space regions in order to use them in explicit training of a binary out-of-distribution classifier. Although predictions and uncertainty from generative models have been shown to improve outlier and adversarial attack detection in contrast to purely discriminative models \citep{Mundt2022,Mundt2019,Li2019a}, there is strong empirical evidence that this is still insufficient to provide a generic solution \citep{Nalisnick2019,Ovadia2019,Mundt2022,Mundt2019}. It is clear that former reported cases of success can be attributed to the specific constrained empirical studies. We illustrate some simple failure cases of prediction confidence and entropy in Figure \ref{fig:openset}, even when uncertainty is assessed with Monte Carlo Dropout. This is to provide an intuitive picture of the challenge of open set recognition with neural networks and to summarize and repeat the findings of the much more detailed experiments presented in numerous prior works \citep{Mundt2022,Mundt2019,Nalisnick2019,Ovadia2019}. 

\subsubsection{Meta-recognition}
Rather than assuming that predictions are somehow calibrated for any data, a more rigorous approach is to prevent overconfident misclassification by confining the model to the known closed space and averting any prediction from little-known open areas in the first place. Whereas it is evident how to achieve this when explicitly modeling the distribution, such as done in probabilistic mixture models, a straightforward approach is not typically applicable in the often complex feature hierarchies of modern discriminative machine learning approaches. A common technique is thus to resort to meta-recognition. In this context, the term meta is to denote a recognition procedure on top of the empirically emerged features obtained through the original black-box optimization procedures. \citet{Scheirer2014} give an intuitive example based on support vector machines. Here, the menace of erratic predictions for unknown unknowns results from examples being projected close to the linear decision boundary, while at the same time being mapped arbitrarily far away from the training data along a different dimension. The authors therefore define a compact abating probability (CAP) model, where the key idea is to make use of insights from extreme value theory (EVT). The essential notion is to take into account inherently present extreme statistical differences in the long tail of an extreme value distribution, here the Weibull distribution. Subsequently, a data point's probability of belonging to the observed closed set is monotonically decreased with increasing distance from the observed data population. In other words, a prediction is discarded in sparsely populated areas, independently of a sample's proximity to the decision boundary. \citet{Bendale2016} have extended this approach to discriminative deep neural networks, where the above idea is transferred to the network's penultimate layer. They propose the OpenMax algorithm that lowers softmax prediction probabilities with increasing distance from the average penultimate layer's activation values. A strongly related approach has been proposed by \citet{Lee2018}, where the affinity of a data point to the known set is measured based on a Mahalanobis distance in the feature space of the penultimate layer. More recent works have come to the conclusion that although the latter approaches have a strong theoretical foundation for open set recognition, they are still limited by activation values in discriminative neural networks being optimized exclusively towards predicting a correct class \citep{Yoshihashi2019,Mundt2022,Mundt2019}. In particular, the penultimate layer activation values do not generally encode all the information about the data that might be required for open set recognition. ``Classification Reconstruction learning for Open-Set Recognition'' (CROSR) \citep{Yoshihashi2019} has thus suggested to additionally append a generative model's latent variable to the OpenMax classification procedure. Concurrently, open variational autoencoders (OpenVAE) \citep{Mundt2022,Mundt2019} translate the EVT based meta-recognition to a variational Bayesian setting. Here, the open set recognition is based directly on the approximate posterior in a deep generative model, which enables a natural interpretation based directly on the underlying generative factors of the data distribution, instead of activation value heuristics. 

\subsection{Bridging perspectives: notable past insights and their synergies}\label{sec:bridge}
Although our survey parts up to now have contained a critical perspective, we have largely kept up the tradition to treat continual machine learning, active learning and open set recognition as three distinct challenges. While well studied in isolation, distinctly categorized approaches are rarely coupled and synergies exploited only in select works, such as the combined continual learning approaches. More importantly, the intersection between the three machine learning paradigms remains largely unexplored, as has also been made evident from our visual summary in the taxonomy of Figure \ref{fig:literature_diagram}. Highlighting the necessity for unification of the latter into a single viewpoint is the primary purpose of this work. The remainder of the section, and the overall paper as whole, will now serve the purpose of revealing the natural interface. In fact, by identifying former lessons, stressing shortcomings of prevailing evaluation practices and bridging seemingly forgotten connections, we develop a wholistic view that simplifies the deluge of ongoing research questions into a single intuitive framework. To better understand why this is imperative for future progress, let us briefly recall the earlier mentioned predominant evaluation routines and link insights from prior works to their current limitations. 

If we look back at Figure \ref{fig:LML_evaluation} and the corresponding section's discussion, we recall that deep continual learning typically collapses its practical evaluation to measuring catastrophic forgetting between task increments. These task increments belong to simple sequentialized versions of existing benchmark datasets. A continual learning technique is deemed successful if the model that is trained over time approaches the expected performance when trained in isolation. In almost complete analogy, active learning evaluation revolves around accuracy gains between query steps. In the majority of the aforementioned related works, the focus is exclusively on whether a specific query mechanism surpasses another in terms of quickly approaching the overall error achieved on a complete dataset. For empirical benchmarking purposes, the model is simply trained in isolation on multiple selected subsets of known data, where the difference between these subsets corresponds to the inclusion of one active query. 

Before we continue with the limitations of such evaluation protocols, we emphasize that our intention at no point in this paper is to discredit and devalue the bulk of previously proposed methods. However, we would argue that made advances of individual methods can be significantly improved beyond their present constrained benchmark evaluation towards progress on a larger machine learning scale. We believe a major contributing factor in this next step is to revisit key insights from past, often neural network unrelated, literature, that have surprisingly gone unnoticed or have been written off in the era of deep learning. To attach a slightly provocative connotation, we have termed these overlooked insights forgotten lessons. Although the term ``forgotten'' certainly is an exaggeration with regard to the ML field as a whole, the absence of derived practical implications is strongly manifested in deep continual learning evaluation schemes. 

\subsubsection{Forgotten lessons from past literature} 
\smallskip
\emph{\textbf{Forgotten lesson 1:} Machine learning models are by definition trained in a closed world, but real-world deployment is not similarly confined. Discriminative neural networks yield overconfident predictions on any sample.} \\

Independently of whether additional metrics such as training speed-ups through representation transfer, computational cost or memory consumption are taken into account, currently considered experimentation features closed world train and test sets. This is occasionally amplified by continual learning works assuming the presence of a task oracle for testing or respectively the assumption of an infallible oracle to yield flawless data when labelling active learning queries. As such, open issues concerning continual training of a model or active learning queries in an open world are generally neglected. However, real-world deployment almost always inhabits an open world. In the extreme case, the model has to handle data from completely unknown type in previously unfamiliar conditions, think outdoor environments or uncontrolled arbitrary user inputs in web-based applications. Instead of the common overconfident misprediction that falsely attributes this data to any known concept, any machine learning model should at least be equipped with the ability to identify unencountered scenarios and warn the practitioner. As a much milder, but heavily realistic form of an open world, even commonly occurring corruptions are frequently disregarded, think blur or camera noise in images. The menace of the latter has recently been demonstrated in deep learning by \citet{Hendrycks2019}. The authors empirically demonstrate that current deep neural networks not only exhibit severe instability with respect to various simple perturbations, but advances in neural network architectures are reflected in only diminutive changes in robustness.  
Whereas certainly this hazard is universal to all machine learning research that is deployed in practice, continual and active learning are particularly prone to the threat of corrupted and unknown data as their goal is to accumulate knowledge from previously unseen sources already in the training process. \\

\emph{\textbf{Forgotten lesson 2:} Uncertainty is not predictive of the open set. Active learning resides in an open world and common heuristics based query mechanism are susceptible to meaningless or uninformative outliers.} \\

Although early works have rapidly identified the fallacy that uncertainty sampling is a meaningful strategy to query \citep{Roy2001,Settles2008} in active learning or respectively detect unknown unknowns \citep{Matan1990, Atlas1990}, the belief that uncertainty provides a generic solution seems to have resurged with the advances of deep learning. This is apparent from the many approaches in our previous literature review basing querying strategies or detection of unseen examples on heuristics that rely on output variability or similar entropic quantities, see the branches labelled with uncertainty and predictive anomalies in our literature review diagram \ref{fig:literature_diagram}. Indeed, the challenge of accurate uncertainty quantification in deep learning is already genuinely difficult and does provide advantages in contrast to less principled empirical thresholding. However, paying homage to the detailed argumentation of the recent review by \citet{Boult2019}, any machine learning model is still trained in a closed world scenario, independently of whether e.g. a Bayesian formalism is employed to obtain uncertainties. Predictions are known to be overconfident, uncertainty is not calibrated for points outside of the training distribution and the posterior is often unusable, regardless of how well it is approximated. \\
In other words, given any parametrized model and its latent variables, we don't know if evaluating the posterior (approximation) will produce something meaningful for an unknown unseen data input. This issue is by no means exclusive to detecting unknown unknown examples, but comes with the same implications for realistic active learning scenarios. Take for example a more realistic set-up beyond a crafted benchmark where data is scarce and the investigated domain is demanding even for experts. The earlier reviewed VAAL has considered such a scenario with medical imaging, where correct oracle labelling and a noiseless image cannot always be expected. Sample selection based on uncertainty does not protect the query from such noise and there is a large chance that meaningless outliers are included into the system. \\

\emph{\textbf{Forgotten lesson 3:} Confidence or uncertainty calibration, as well as explicit optimization of negative examples can never be sufficient to recognize the limitless amount of unknown unknowns.} \\

At a first look, one might believe that impressive successes where demonstrated with approaches that extend the basic idea of ``inference with the universum'' \citep{Weston2006}. Explicitly using prior knowledge in terms of expectations on what form of inputs can be anticipated, or respective inclusion of negative data that is believed to play a role in deployment, are popularly exhibited by works that have identified and attempt to address the first two lessons. The common presumption across all these works is the upfront presence of a larger, possibly unlabelled, dataset that can explicitly be included into the optimization process. Just as supposed out-of-distribution examples are made use of to modify loss functions and calibrate the output for detection of unknown unknowns \citep{Bell2015,Lee2018,Yu2019,Dhamija2018,Feng2019}, active learning techniques often resort to conditioning their procedure on the entire data pool \citep{Nguyen2004,Sener2018,Li2013,Shui2020,Sinha2019}, e.g. through clustering \citep{Nguyen2004,Sener2018} or fitting a generative model to the unseen data \citep{Li2013,Shui2020}. 
Unfortunately, this impedes evaluation beyond a constrained closed set benchmark and more realistic continual and active learning scenarios where data becomes available at different times cannot be considered. In a sense the problem seems to be addressed from a reverse perspective. Instead of acquiring explicit knowledge about the nature of the trained data distribution, the challenge is sidestepped by reformulating it as an optimization problem that attempts to find the boundary between known and an existing set of unseen data, which by definition then does not consist of unknown unknowns. Thus, we receive no guarantees, as the pool of unlabelled data at any point in time is limited and can never truly approximate the unknown space. 

The obvious argument is now that it is impossible to include all forms of variations and exceptions upfront, else we could have just modeled and hand-crafted the entire system from the start instead of falling back on purely data driven approaches. As a second additional argument, previous works have also asserted that the particular form of representations of discriminative deep neural networks can further confound predictions. The early work of \citet{French1992} has already pointed out that a major complication of continually training neural networks is their distributed representations. It has subsequently investigated mechanism to obtain semi-distributed representations with sharp activations that are concept specific. We argue that with the onset of deep learning the challenge of distributed representations is further magnified due to distribution across the layer hierarchy. First, consider as an example a neural network that is trained to discriminate cars from aeroplanes. Such a scenario is often assumed when incrementally training the popular CIFAR10 dataset \citep{Krizhevsky2009}. As the neural network is not explicitly encouraged to encode information about the data distribution, the obstacle of predicting overconfidently on unseen data is further magnified by the ubiquitous option for any classifier to differentiate a concept based on a combination of noise patterns, the absence of a specific pattern, or background patterns altogether \citep{Xiao2020}. In the car versus aeroplane scenario, depending on how well and diverse the dataset is constructed, this could be as trivial as distinguishing the two classes by identifying the presence of some feature that describes the sky. As neural networks have been demonstrated to rely heavily on texture rather than object boundaries \citep{Geirhos2019}, this is not far fetched. In fact, a prominent recent work on ``Unmasking Clever Hans'' predictors \citep{Lapuschkin2019} has shown that the decision making of a discriminative deep neural network can be based on entirely trivial features, such as a certain object always occurring at a specific location in every image or almost imperceivable photography tags. ``Adversarial examples are not bugs they are features'' \citep{Ilyas2019} takes this one step further and empirically showcases how classes can be distinguishable solely based on noise patterns. In a trivial case of our above car versus aeroplane example, presenting the trained model with images of ships that feature the similarly blue background of the sea is then not surprisingly resulting in overconfident misclassification. Using ships as a background class could initially solve this problem of attributing blue to aeroplanes. However, if a significant portion of our learned features were indeed to be composed of noise, background and adversarial patterns, then we would argue that overconfident mispredictions are impossible to overcome, as the extent of data on which these features activate is inconceivable to any human modeler. This makes the approach to handle outlying and unknown unknown data through prior knowledge even less feasible. \\

\emph{\textbf{Forgotten lesson 4:} Data and task ordering are essential. Although this forms the quintessence of active learning it is yet untended to in continual learning.} \\

It is well known that each dataset instance does not contribute equally to the overall objective. This forms the foundation and rationale behind active learning. In general, when conducting active learning queries, there is a trade-off between exploring the unknown space and exploiting more of the already known to avoid misclassification \citep{Joshi2009}. Alas, the implications of the latter statement are more nuanced and go beyond the simple question of whether a certain subset spans the entire data distribution. As an example, \citet{Joshi2009} found certain active learning strategies to benefit primarily from creating a class imbalance, as more difficult classes might require a denser sampling than others. \citet{Bengio2009} have similarly found that sorting data in a curriculum that introduces classes into the training process according to their difficulty improves the obtained accuracy. Recently, \citet{Hacohen2020} have empirically observed that deep neural networks seem to build such a curriculum inherently during the training process. Consistently across multiple architectures, they always learn the same examples first when given access to the entire dataset, even though the mini-batch stochastic gradient descent shuffles the data differently every time. \citet{Pliushch2022} have subsequently observed that this phenomenon can be correlated to increase/decrease in various image measures and statistics. This notion of learning according to some measure of complexity seems intuitive, as describing some inputs necessitates less nuanced patterns than others. 

Even though there is significant empirical evidence that data selection and task order plays a vital role for any learned algorithm, modern deep continual learning seems to pay little attention to a careful experimental design. Out of the numerous works of the previous review, less than a handful of works consider the question of task order at all. The rest remains in the comfort of benchmark datasets, where the classes are split and introduced in sequence for continual learning according to a class id that often just reflects an alphabetic ordering. However, there is no rigorous investigation of the effect of task order. Two out of the four works that examine task order \citep{Serra2018,Isele2018} only randomize the order across multiple experimental repetitions to obtain an average performance estimate. The other two \citep{DeLange2019, Javed2018} follow this practice, but go even further and make the statement that task ordering has minimal influence towards continual learning methods. We will later demonstrate that this is obviously not the case, and can simply be attributed to the experimentation being a narrow trial of five randomly obtained orderings without any attached semantics. When selecting tasks from the overall pool of available data according to their similarity or dissimilarity with the already observed data distribution, we will observe a major divergence of obtained results.

Whether or not having access to all future tasks in order to select an ideal order is unrealistic in real-world continual learning scenarios, we believe task ordering to be an imperative factor that should be considered when designing our benchmarks to further our understanding. In particular, we note that a very common practice to reduce the computational cost of incrementally learning large scale datasets such as ImageNet \citep{Russakovsky2015} is to extract subsets \citep{Rebuffi2017,Wu2019,DeLange2019,Park2019}. The main problem here is that selecting e.g. 50 from a larger pool of 1000 classes heavily influences the achievable result and using random selection mechanisms can essentially render works unreproducible.  \\

\subsubsection{Open set recognition forms the natural interface between continual and active learning}
As indicated in the previous sections, contemporary continual and active learning are prone to an large amount of threats due to their development and evaluation inhabiting a closed world. In this section we argue that awareness of an open world is not only required to overcome the threat of designing a non-robust system, but provide the natural means to merge techniques into a common perspective. \\

\textbf{Different sides to the same question:} First recall that a majority of continual learning techniques alleviates the challenge of catastrophic inference by regularizing parameters for known tasks, rehearsing a subset of data from known tasks or respectively generating it with a generative model. Independent of the specific algorithm, a key concern is thus to identify exemplars, learn the generative factors of our known tasks or determine the parameters that are responsible for the majority of previously seen data. At the core, we need to thus \textit{find a good approximation of the known data distribution}.

 In active learning, also recall from our survey how the task is very much alike, although the underlying question seems to be of reversed nature. Instead of protecting or sampling from the known data distribution, \textit{a query is conducted with respect to yet unobserved distributions}. In a similar distinction to the continual learning mechanisms, query-acquiring active learning methods pick samples that are estimated to yield the best model improvement, whereas query-synthesizing methods attempt to tackle this challenge through generative modeling by generating these most informative examples.

Interestingly, in open set recognition, the task is to \textit{precisely gauge the boundary between the seen known data distribution and yet unseen unknown data}. The original motivation stems from a perspective of outlier detection and thus model robustness in practical application in the presence of unknown unknowns. However, knowing this boundary also gives us the means to restrict a continual learning technique to protect the already seen knowns or respectively query active learning examples that are sufficiently statistically different without the fear of selecting uninformative noise. We thus propose that open set recognition forms a natural general interface between active and continual learning. \\

\begin{figure*}[t]
	\includegraphics[width=\textwidth]{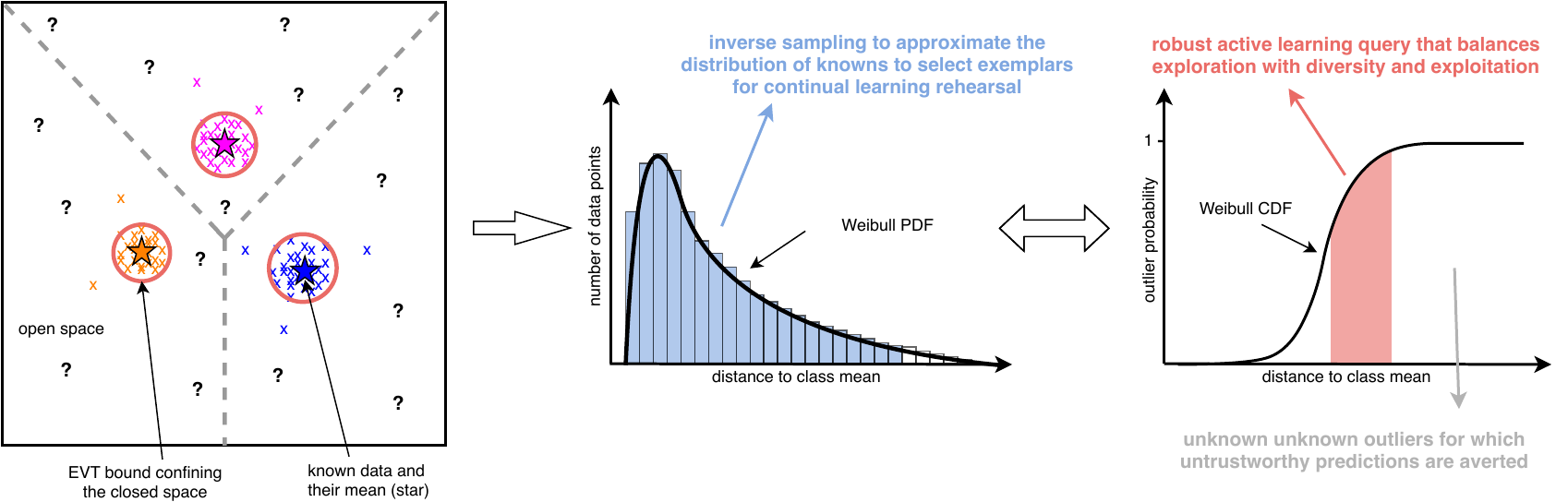}
	\caption{\label{fig:EVT_diagram} Conceptual diagram to illustrate how extreme value theory based meta-recognition in neural networks can serve as a common denominator to protect knowledge in continual learning, conduct principled queries in active data selection, while having the capability to reject or set aside unknown unknown data at any point in time. The leftmost figure of an embedding showcases the threat of the open space, where any examples that are very far away from known clusters always get falsely assigned to a known class and can be arbitrarily close to the decision boundary. The mid panel shows how a Weibull distribution, which models the extreme distance values to the mean of the correctly predicted trained data in a heavy tail, can enclose the known space (suggested by the red circles in the embedding). The corresponding cumulative distribution function in the right panel can be used to reject or set aside outliers and balance active learning queries to sample diverse, yet meaningful data (shaded red area). Alternatively either curves can be sampled inversely to select a subset of inlying data to approximate the entire known distribution in continual learning rehearsal (shaded blue area). Best viewed in color.}
\end{figure*}

\textbf{A unified framework based on meta-recognition \\ through EVT:} Out of the previously reviewed works, we posit that works that employ EVT based meta-recognition to identify unknown unknowns are one particularly suitable example to build a unified framework for our wholistic view. We schematically illustrate this proposed framework in Figure \ref{fig:EVT_diagram}. We will delve into conceivable mathematical details and a potential realization in deep neural networks in the next section, where we also further corroborate our view with empirical findings. For now, consider a generic embedding as a result of some deep neural network encoding. In the figure's leftmost panel, we have visualized an example embedding for three classes, with their mean indicated by a star and a potential decision boundary by dashed lines. In order to confine predictions to the known space, EVT based meta-recognition makes use of data instances with extreme distance values to the average embedding of a class. Typically, a Weibull distribution is used to model the distance distribution for the entire dataset and capture samples that feature stronger deviation in a heavy tail. In the original works that have proposed this model for open set recognition \citep{Scheirer2013,Scheirer2014, Bendale2015}, the cumulative distribution function is then used to estimate whether a new unseen example should be regarded as an unknown unknown, outlying data point. In our own previous work \citep{Mundt2022}, we have identified this technique to also be fundamental in judging whether a randomly sampled latent vector is proximate enough to the observed data such that it results in a clear output of a generated model. 

We now close the circle and tie this method to retention of a core set for continual learning, as well as a query mechanism for active learning, while retaining the method's innate ability to reject and set aside unknown unknowns. 
	\begin{enumerate}
		\item First, we postulate that the Weibull distribution for each data point's distance to the mean embedding equips us with a tool to approximate the known distribution with a subset. Specifically, we can employ inverse sampling from the Weibull probability density function to create a set of distance values with an arbitrary prior on how much of the distribution's tail should be disregarded, i.e. how many outliers are already assumed to be inherently present in the original dataset. Practically, we can then approximate the data distribution with a subset by selecting data instances whose embedded value lies closest to the drawn sample. Alternatively, as indicated in the diagram, we could discretize the distribution and sample a certain number of examples from each bin.
		\item Conversely, for active learning, we are less interested in sampling from the known distribution, but much more in the heavy tail. To our advantage, the long tail models data that is statistically deviating, but can still be attributed to the distribution of interest. We can thus balance exploitation with exploration. First and foremost, data instances for which the outlier probability is unity are avoided altogether in order to prevent sampling of uninformative noise or other corrupted data. Recall, that this is the primary pitfall of uncertainty sampling. At the same time, we want to avoid samples that have a minute probability of being an outlier, as these samples are too similar to previously observed data and are therefore also uninformative due to redundancy. As such, we can constrain our query to the center area of the cumulative distribution function (CDF), illustrated by the shaded area under the CDF in the diagram. The rationale for this approach can intuitively be understood by looking back at the theoretically grounded works on version spaces. We can implicitly reduce the space of possible hypotheses, even in complex models such as neural networks, as we incrementally expand the radius of the ball that encloses the closed space by sampling carefully along its boundary with each active learning query. This way, we avoid the vast open space and the redundant highly dense areas of known data, while making sure that previously unseen information is acquired.
		\item Lastly, as already hinted at in the previous second point, we obtain a robust model, where data instances for which the outlier probability is very high, i.e.~close to unity, can not only be discarded to favor a trade-off in exploration versus exploitation in active learning, but also simply to flag instances where predictions should not be trusted if the model is outside of a training phase. 
	\end{enumerate}

In the next section, we will further contextualize this wholistic view on the basis of a unified framework realization in neural networks at the hand of empirical examples. Before we proceed, we note that there are two works that have previously initiated a bridge between active learning and open set recognition, alas have not fully built it yet. The recently introduced open world learning \citep{Bendale2016} and the concurrently named cumulative learning \citep{Fei2016} advance the pure open set identification step by proposing to set aside the unknown unknowns and including them into a later active learning cycle. Whereas these works made first steps towards formulating learning in an open world, they however assume the presence of labels for the entire dataset and the addition of classes itself is in the form of a fixed sequence that is injected by the human. The system is limited as it does not self-select which classes or instances should be learned next, nor does it protect its knowledge for continual learning, where the assumption of availability of all data at all times is lifted. As a result, the empirical evaluation is simply an investigation of the performance on the entire test set at each state of the growing known training set. Finally, the suggested open world learning \citep{Bendale2016} is based on nearest mean classifiers based on simple SIFT features and is yet to be extended to the context of modern deep neural networks. 

\section{Highlighting natural synergies with empirical evidence}
In order to support our critical survey and make its derived proposition for a unified view practically tangible, we now highlight the emerged natural synergies with empirical evidence. For this purpose, we conduct four sets of quantitative experiments, which follow our previous narrative and relate to the formulated forgotten lessons. Each experiment will be summarized and discussed in a respective subsection in detail. Crucially, we emphasize that they all share the common denominator of making use of the same unified neural network framework. Although individual improvements over some related works will be shown, the key novelty thus lies in making use of the same mechanisms across all applications, which is untypical in respective related works and previous reviews. More specifically, we showcase:

\begin{enumerate}
	\item \textit{Continual exemplar selection:} we start with a quantitative comparison of exemplar selection mechanisms to prevent catastrophic forgetting in continual learning. Here, we will first show that the proposed common EVT based foundation surpasses several conventionally employed techniques.
	\item \textit{Active queries:} in similar spirit, we show the complementary perspective by investigating querying strategies in active learning. Again, we will show that the proposed common EVT based foundation surpasses typical approaches.
	\item \textit{Corruption robustness:} we then proceed to further highlight the method's superiority in the open world. In contrast to most methods that are developed with a unidirectional focus on improving a specific active learning or continual learning benchmark, our framework has the critical advantage of not breaking down in the presence of corruptions that commonly occur in practical application in the wild.
	\item \textit{Task ordering \& learning curricula:} to conclude the experiments, we investigate the role of task order and the effect of a curriculum for evaluation. We show that a task curriculum constructed through our framework consistently results in considerable improvements.
\end{enumerate}

The corresponding detailed set-up and quantitative evaluation is provided in section \ref{sec:empirical_evidence}. Before diving into this discussion in detail, section \ref{sec:gen_framework} first introduces one conceivable proposition for a unified framework on the basis of deep generative models and variational inference. Although this specific framework surely contains parts that are novel to a certain degree, we wish to emphasize that our aim is not to promote this particular neural network realization or advocate it as a unique solution to the developed view. In contrast, the mathematical tools and neural network variant shall serve as a \textit{teaching example}, valued primarily for its interpretability in illustrating the potential of embracing a wholistic view grounded in a common mechanism. For this purpose, some approximations will be made and pointed out, to draw attention to relevant factors and promote the reader's comprehension.  As such, illustrated performances are one important factor to take into consideration, but the primary goal of the experimental section is not to develop a novel state-of-the-art technique. Instead, the focus is on developing practical intuition and providing empirical evidence to encourage adoption of a wholistic view beyond prior constrained evaluation in future research. 

\begin{figure*}[t]
	\includegraphics[width=\textwidth]{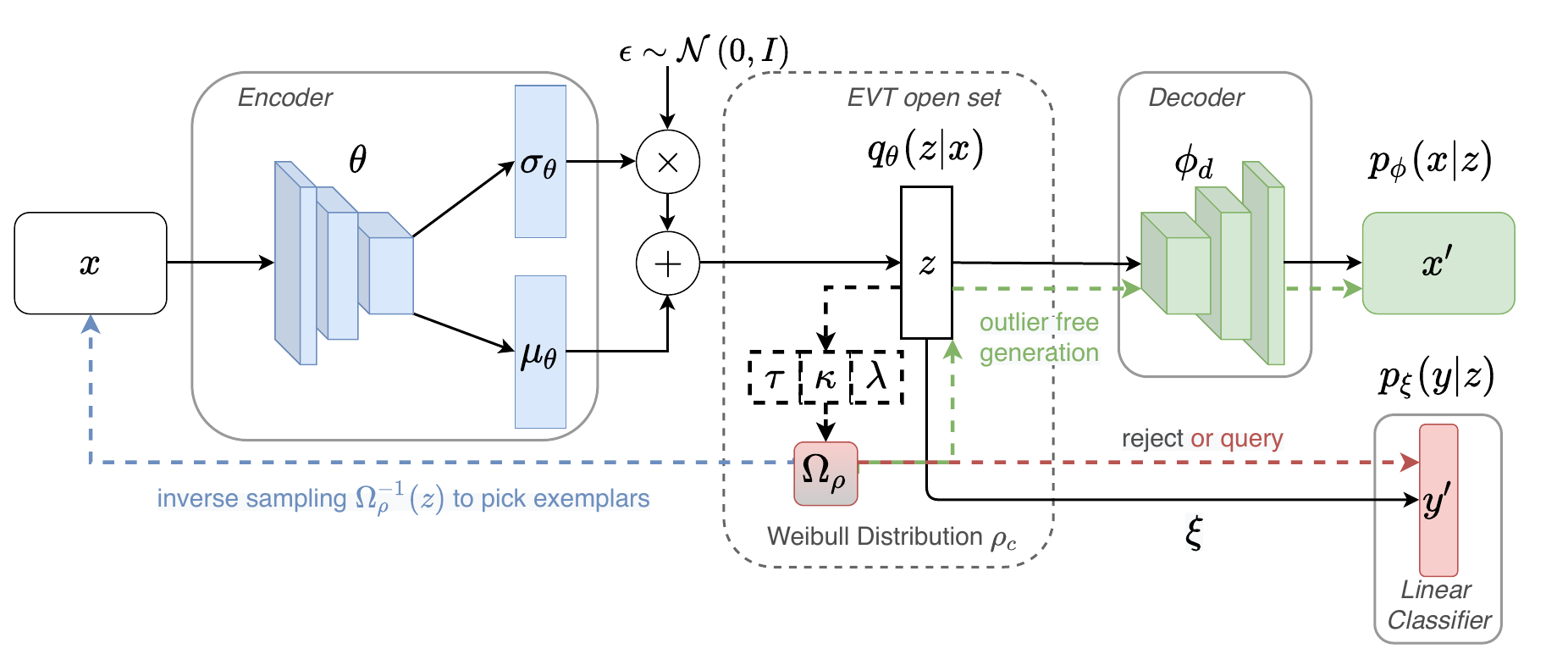}
	\caption{\label{fig:VAE_EVT_diagram} Architecture diagram for our particular VAE based practical neural network framework realization. Here, the solid outlined boxes represent a traditional VAE model, whereas the dashed boxes and lines correspond to the auxiliary EVT based components. The color coding of the diagram is picked to resemble our previous figure \ref{fig:EVT_diagram}, which focuses on the details behind the intuition of the dashed components. Similar to this earlier figure, we can again see how EVT serves as a common denominator to protect knowledge in continual learning, conduct principled queries in active data selection, while having the capability to reject or set aside unknown unknown data at any point in time. Respectively, the blue color denotes meaningful choice of exemplars, the green color represents outlier free generative replay (not shown in figure \ref{fig:EVT_diagram}), whereas the red and grey colors represent the choice to query or altogether reject data. Best viewed in color}
\end{figure*}

\subsection{One way to unite perspectives with deep generative neural networks}\label{sec:gen_framework}
How can we realize our proposed unified framework in a meaningful way in deep neural networks? As emphasized by prior work \citep{Yoshihashi2019,Mundt2019}, identification and correlation of unseen data with average activation patterns of known data is not necessarily sufficient in discriminative models. This holds even when extreme values are modeled to obtain closed space boundaries, see prior works \citep{Mundt2022,Mundt2019} for empirical verification. It is because a neural network based classifier is generally not encouraged to aggregate the whole information describing the data, merely the features that allow for class distinction. These features themselves, come with a variety of further pitfalls, as summarized in the forgotten lessons. In our own previous work \citep{Mundt2022,Mundt2019}, we have overcome this limitation by formulating the problem from a perspective of deep generative models trained with variational Bayesian inference, i.e. variational autoencoders (VAE) \citep{Kingma2013}. We will lean on this viewpoint, follow the notation of prior works and extend it towards one potential solution to consolidate continual and active learning through open set recognition. To provide visual intuition behind our practical framework realization, in addition to the earlier generic figure \ref{fig:EVT_diagram}, we provide an architecture sketch in figure \ref{fig:VAE_EVT_diagram}. 

The rationale to build upon VAEs is rather straightforward: the Bayesian formulation lets us learn about the distribution of seen data $p(\mathbf{x})$ by capturing it through latent variables $\mathbf{z}$. However, as $p(\mathbf{x}) = \int p(\mathbf{x, z}) dz$ is intractable, we do this by optimizing a lower-bound to the marginal distribution $p(\mathbf{x})$, since the densities of the marginal and joint distribution are related through Bayes rule $p(\mathbf{z} | \mathbf{x}) = \frac{p(\mathbf{x, z})}{p(\mathbf{x})}$. As we do not know our real posterior $p(\mathbf{z} | \mathbf{x})$, we typically resort to variational inference and introduce a variational approximation $q(\mathbf{z} | \mathbf{x})$ to the posterior. In a neural network, this approximation $q(\mathbf{z} | \mathbf{x})$ is learned through the parameters of a probabilistic (blackbox) encoder, whereas a probabilistic decoder is trained for the joint distribution $p(\mathbf{x}, \mathbf{z}) = p(\mathbf{x} | \mathbf{z}) p(\mathbf{z})$ and thus forms the generative component. This generative model can effortlessly be augmented to additionally discriminate classes by including their label into the latent variable, e.g. by enforcing a linear class separation on $\boldsymbol{z}$. The corresponding factorization and generative process is then $p(\boldsymbol{x}, \boldsymbol{y}, \boldsymbol{z}) = p(\boldsymbol{x}|\boldsymbol{z})p(\boldsymbol{y}|\boldsymbol{z})p(\boldsymbol{z})$ \citep{Mundt2022,Mundt2019}. Such formulation of a classifying variational autoencoder comes with the main advantage that using latent variables $\mathbf{z}$ allows us to base our decision regarding unknown unknowns on the underlying generative factors of variation. We can determine whether an example is close to the high density regions of our approximated data distribution. 

\subsubsection{The boundary between known and unknown}
The first step towards open world aware active and continual learning is to train the above mentioned classifying variational autoencoder, followed by determining the boundary between the open and closed spaces for the observed distribution with the help of EVT. For ease of readability, we repeat the training and fitting procedure described in our previous work \citep{Mundt2022, Mundt2019}.
The model's probabilistic encoder and decoder are trained jointly by minimizing the divergence between the variational approximation $q_{\boldsymbol{\theta}}(\boldsymbol{z}|\boldsymbol{x})$ and a chosen prior $p(\boldsymbol{z})$, typically $\mathcal{N} \sim (0, I)$, and the conjunction of reconstruction loss and the linear classification objective, parametrized through $\boldsymbol{\phi}$ and $\boldsymbol{\xi}$ respectively. For a dataset consisting of $n=1,\ldots,N$ elements, the following lower bound to the joint distribution $p(\boldsymbol{x,y})$ is thus optimized:

\begin{equation} \label{eq:general_loss}
\begin{aligned}
& \mathcal{L}\left(\boldsymbol{x}^{(n)}, \boldsymbol{y}^{(n)}; \boldsymbol{\theta}, \boldsymbol{\phi}, \boldsymbol{\xi} \right) = - \beta \kld{q_{\boldsymbol{\theta}}(\boldsymbol{z} | \boldsymbol{x}^{(n)})}{p(\boldsymbol{z})} \\
& \quad \, \, + \mathbb{E}_{q_{\boldsymbol{\theta}}(\boldsymbol{z} | \boldsymbol{x}^{(n)})} \left[ \log{p_{\boldsymbol{\phi}}(\boldsymbol{x}^{(n)} | \boldsymbol{z})} + \log{p_{\boldsymbol{\xi}}(y^{(n)} | \boldsymbol{z})} \right] 
\end{aligned}
\end{equation}

Here, $\mathbb{E}_{q_{\boldsymbol{\theta}}(\boldsymbol{z} | \boldsymbol{x})}$ is generally estimated via Monte-Carlo sampling. At any point in time of training this model, there is a natural discrepancy between the prior and the approximate posterior. The added $\beta$ factor in above equation serves the purpose of controlling this gap, with a $\beta > 0$ typically determined through cross-validation. Whereas one could believe this distributional mismatch to be an undesired property, we recall the arguments conjectured in multiple previous works \citep{Hoffman2016,Burgess2017,Mathieu2019}. In essence, they state that the overlap of the encoding needs to be reduced in order to avoid indistinguishability, but at the same time prevent latent variables to consist of individual uncorrelated data points that resemble a pure look-up table. In the intuitive picture of diagram \ref{fig:EVT_diagram}, think of the former as multiple classes collapsing and thus being inseparable. Think of the latter as the dense clusters being scattered to allow differentiation of each and every single data point without a strong encoding of correlations. Therefore, the actually captured encoding of the data distribution should not simply be assumed to correspond to the prior, but rather corresponds to an empirically determinable distribution referred to as the aggregate posterior:
\begin{equation}\label{eq:aggregate_posterior}
q_{\boldsymbol{\theta}}(\boldsymbol{z}) = \mathbb{E}_{p(\boldsymbol{x})} \left[ q_{\boldsymbol{\theta}}(\boldsymbol{z} | \boldsymbol{x})\right] \approx \frac{1}{N} \sum_{n=1}^{N} q_{\boldsymbol{\theta}}(\boldsymbol{z} | \boldsymbol{x}^{(n)})
\end{equation}
Using EVT to find the boundaries of this distribution now corresponds to identification of our model's closed space. For emphasis, we repeat that this is necessary because VAEs generally assign non-zero density to any point in the latent space, the analogue of overconfident classifier predictions \citep{Nalisnick2019,Ovadia2019}. The boundary is not analogous to the extent of the prior because low density areas exist inside the prior as well. Practically, an EVT based fit can be obtained by empirically accumulating the mean latent variable for each class $c$ for all correctly predicted known data points $m=1,\ldots,M$:
\begin{equation}\label{eq:mean_z}
\boldsymbol{\bar{z}}_{c} = \frac{1}{|M_{c}|}  \sum_{m \in M} \mathbb{E}_{q_{\theta}(\boldsymbol{z}|\boldsymbol{x}^{(m)})} \left[ \boldsymbol{z} \right]
\end{equation}
and defining a respective set of latent distances as: 
\begin{equation}\label{eq:latent_distance}
\Delta_{c} \equiv \left\{ f_{d} \left( \boldsymbol{\bar{z}}_{c}, \mathbb{E}_{q_{\theta}(\boldsymbol{z}|\boldsymbol{x}_{t}^{(m)})} \left[ \boldsymbol{z} \right] \right) \right\}_{m \in M_{c}}
\end{equation}
Here, $f_{d}$ represents a chosen distance function, which prior works have typically chosen to be either euclidean or cosine distance \citep{Scheirer2013,Scheirer2014,Bendale2015,Mundt2022}. As this set represents the distances to the class conditional aggregate posterior, we can fit a Weibull distribution with parameters $\rho_c = (\tau_c , \kappa_c , \lambda_c )$ on $\Delta_{c}$ to model the trustworthy regions of high density that represent the observed data distribution, where the heavy-tail indicates a decaying reliability:
\begin{equation}\label{eq:Weibull_PDF}
\omega_{\boldsymbol{\rho}} (\boldsymbol{z}) = \frac{\boldsymbol{\kappa}}{\boldsymbol{\lambda}} \left( \frac{| f_{d} \left( \bar{\boldsymbol{z}}, \boldsymbol{z} \right)- \boldsymbol{\tau}  |}{\boldsymbol{\lambda}} \right)^{\boldsymbol{\kappa} - 1} \exp \left(-  \frac{| f_{d} \left( \bar{\boldsymbol{z}}, \boldsymbol{z} \right)- \boldsymbol{\tau} |}{\boldsymbol{\lambda}} \right) ^{\boldsymbol{\kappa}}
\end{equation}
Here, $\tau$ defines the location, $\lambda$ the scale and $\kappa$ the shape of the distribution. We can now make use of this distribution to pinpoint the observed data distribution, as a surrogate to the otherwise highly complex aggregate posterior. We proceed to highlight its various use cases in the following sections, which will in turn be used in the respective four experiments. 

\subsubsection{Approximate posterior based open set recognition}
As described in previous works \citep{Mundt2022, Mundt2019}, the most direct use of the aggregate posterior based Weibull parameters $\boldsymbol{\rho}$ is the identification, rejection or storage of unknown data. Using the corresponding cumulative distribution function (CDF) to the probability density function of equation \ref{eq:Weibull_PDF}, we can now estimate any data instance's statistical outlier probability for every known class:
\begin{equation}\label{eq:outlier_probability}
\Omega_{\rho_c}(\boldsymbol{z}) =  1 - \exp \left(-  \frac{| f_{d} \left( \bar{\boldsymbol{z}}_c, \boldsymbol{z} \right)- \tau_c |}{\lambda_c} \right) ^{\kappa_c} 
\end{equation}
When we have observed multiple classes, we will typically take the minimum $\min{(\Omega_{\boldsymbol{\rho}})}$ of this equation across all known classes $c$ and the respective mode's parameters $\rho_c$. This expresses the basic condition that a data point should be considered as a statistical anomaly only if its outlier probability is large for each known class. A respective decision should thus be based on the class where the smallest deviation to known data is observed. The more dissimilar a sample is with respect to the observed data distribution as approximated by the aggregate posterior, the more the outlier probability will approach unity. Irrespective of whether a machine learning algorithm is developed for active learning, continual learning or in fact any other paradigm, this robustness towards unknown unknown data is essential for any practically deployed system that operates outside of extremely narrow conditions. 

\subsubsection{Outlier and redundancy aware active queries}
Equation \ref{eq:outlier_probability} gives us the direct means to estimate a sample's similarity with the already known data. For active learning this almost directly translates to the informativeness of a query. Small CDF values signify large similarity or overlap with already existing representations, larger values indicate previously unobserved data. Naively, one would follow the earlier strategies developed in uncertainty based active learning and simply query batches that consist of the most outlying data points. However, this would neither grant protection from exploring noisy, perturbed and uninformative data, nor balance it with exploitation to foster partially known concepts. Our proposition is thus to query a variety of data that is well distributed across the center part of the CDF. For instance, we could chose data that surpasses an outlier probability of e.g. $0.5$ and at the same time is limited on the upper end by e.g. a value of $0.95$ (Note that these values present an assumption that is not fixed and simply present a teaching example). As explained in the earlier introduction of the framework, this is tantamount to sampling on the outer edge of the sphere that encloses the currently known closed space. 

\subsubsection{Core set selection for continual learning rehearsal}
In contrast to active queries that need to select meaningful unknown data, in the currently formulated continual learning paradigm the main goal is to protect the known knowledge while learning a predetermined new task. We will investigate the role of the order prearrangement in the next subsection. Here, we focus on open world aware techniques to preserve previously acquired representations. Depending on available memory, the most successful approaches either store and rehearse a small subset of exemplars or alternatively generate data for former tasks with a generative model. In our previous work \citep{Mundt2022} we have shown how we can use equation \ref{eq:outlier_probability} to reject samples from the prior $\boldsymbol{z} \sim p(\boldsymbol{z})$ that do not fall into the obtained bounds of the aggregate posterior for generative rehearsal. The choice for this sampling with rejection originated from the decision to employ the cosine distance, which collapses the distance to a scalar. A different distance function, such as a euclidean distance per dimension would allow to directly inversely sample a highly multi-modal Weibull distribution, i.e. with one mode per dimension per class. We will stick to the easier cosine distance case, both in order to remain at a level of intuitive understanding and because it seems to suffice empirically. Independently of the selected distance metric, we can leverage inverse sampling for the construction of a small data subset. Specifically, drawing at uniform from the inverse of the CDF in equation \ref{eq:outlier_probability} yields samples that approximate the aggregate posterior:
\begin{equation}\label{eq:inverse_weibull}
f_{d} (\bar{\boldsymbol{z}}, \boldsymbol{z}) = \Omega^{-1}(p|\boldsymbol{\tau}, \boldsymbol{\lambda}, \boldsymbol{\kappa}) = \boldsymbol{ \lambda} \left( - \log{ \left(1 - p\right)}^{\frac{1}{\boldsymbol{\kappa}}} \right) - \boldsymbol{\tau}
\end{equation}
The core set can now simply be obtained by picking the data points that are closest to the obtained distance values, if the chosen distance metric collapses the distance to a scalar, or directly to the latent vector, if the chosen distance metric preserves the dimensionality. Note that we have chosen to inversely sample the CDF of equation \ref{eq:outlier_probability} in favor of a more compact equation. It should however be clear that equation \ref{eq:Weibull_PDF} can alternately be sampled equivalently. 
The advantage of such a core set selection procedure is that we always attempt to approximate the underlying distribution. The quality is defined by the desired amount of exemplars, while excluding statistical anomalies by limiting outlier probability values to e.g. $p <0.95$. As anticipated, the latter plays the additional crucial role of robust application when the system has finished learning and is deployed. 

\subsubsection{Class incremental curricula and task order}
Continual learning methods are mostly evaluated in the context of class incremental learning. The classes of a benchmark dataset are typically split into disjoint sets and introduced to the learner in alphabetical or class index sequence. Due to the large computational effort of training neural networks to convergence on long task sequences, several works choose to evaluate on subsets of classes \citep{Rebuffi2017,Wu2019,DeLange2019,Park2019}. An important remaining question is thus how such evaluation affects comparability and reproducibility, or more generally the role of task order and curricula. As mentioned earlier, selecting a meaningful ordering is in most cases non-trivial. Large-scale dataset such as ImageNet are often composed by scraping data from the internet, social media or through uncontrolled acquisition that prioritizes as large as possible datasets. We as humans thus lack the knowledge to build an intuitive learning curriculum when paired with our lack of understanding of deep neural network representations. Consequently, scarcely any works have attempted to address this challenge beyond a simple randomization of the class order. Fortunately, we can provide at least a partial remedy to the seemingly arbitrary class incremental evaluation setting. Although we do not have access to explicit data distributions for any task, equation \ref{eq:outlier_probability} allows us to assess the similarity of new tasks with the aggregate posterior for known tasks. In the spirit of our earlier formulated active learning query, we can start with any task $t$ and proceed to select future tasks $t \in T$ that feature the least overlap with already encountered tasks (or most overlap, depending on what is desired):
\begin{equation}\label{eq:task_order}
t_{\mathtt{next}}  = \argmax_{t \in T} \left\{ \mathbb{E}_{p_{t}(\boldsymbol{x})}\Omega_{\boldsymbol{\rho}}\left( \mathbb{E}_{q_{\boldsymbol{\theta}}(\boldsymbol{z} | \boldsymbol{x})} \left[ z \right] \right) \right\}
\end{equation}
To provide an example, if our objective was to incrementally expand a system to recognize individual animal species, one assumption could be to accelerate training by always including the species that is most similar to what has already been learned, as this could be hypothesized to require only small representational updates. An alternative objective could be to design a system that expands its knowledge in an attempt to cover and generalize to an as large as possible variety of concepts. In this scenario, one could choose to always include the next task with the smallest amount of overlap with existing tasks to maximize diversity in learning. 

Note that these scenarios present two extremes, that are again chosen because they will provide a good teaching foundation in the context of this paper's message. In principle, one could opt to create more complex task ordering measures beyond the max or min overlap around equation \ref{eq:task_order}. We could now also delve into a philosophical debate on when it is reasonable to assume access to future tasks in continual learning to undergo above selection, and when the task sequence is unavoidably dictated by other external factors. We deliberately keep such a discussion short and only emphasize the following aspects. There exist both scenarios in which curricula and task order can realistically be chosen in continual learning settings. An intuitive example of this would perhaps be school like learning environments, where we can flexibly decide what to learn next while retaining prior knowledge. An alternative example could be a robot exploring and learning about an unknown environment, think of a different planet. On the reverse side, some settings will impose the order in which data arrives onto us, such as typically encountered in many real world data streams. Independently of these differences in the ability to alter data curricula and task order, we wish to highlight the large effect on performance when task curricula are chosen by above mechanism in the following empirical investigation. At best, such analysis is relevant for two important elements of learning: 1. the option to decide to set a task aside and learn it at a later point in time, if the curriculum is permitted to be controlled. 2. The effect of task and curriculum choice when considering pre-trained models. At the very least, we hope that our efforts will invoke a more careful and consistent evaluation on existing benchmarks, instead of picking arbitrary data subsets, selecting different random class orders and nevertheless attempting to compare results across methods.

\subsection{Empirical evidence}\label{sec:empirical_evidence}
We base our experiments on the MNIST \citep{LeCun1998}, CIFAR10 and CIFAR100 datasets \citep{Krizhevsky2009}. Although these datasets could be regarded as fairly simple, they are advocated as the predominant benchmarks in all of the presented continual learning works and still present a significant challenge in this context. They are further sufficient to point out major differences between methods, particularly with respect to robustness, showcasing a disconnect with real application and realistic evaluation. For this very reason, we employ them as ``teaching examples'' (similar to our aforementioned framework), despite being the subject of our own initial critique earlier. We take this conscious decision in favor of the reader being able to directly relate to our experiments and understand the importance of the developed synergies in the different set-ups. A discussion on very recent developments concerning efforts towards dataset creation for continual learning can be found at the end of our paper.
We use a 14 layer wide residual network (WRN) \citep{Zagoruyko2016,He2016} encoder and decoder with a widening factor of 10, rectified linear unit activations, weight initialization according to \citet{He2015} and batch normalization \citep{Ioffe2015} with $\epsilon = 10^{-5}$ at every layer, to reflect popular state-of-the-art practice. To avoid finding elaborate learning rate schedules or resorting to other excessive hyper-parameter tuning, we use the Adam optimizer \citep{Kingma2015} with a learning rate of $0.001$ and a sufficiently high-dimensional latent space of size $60$ for all training. We use this common setting to corroborate our wholistic view and describe further details for specific experiments in the consecutive subsections on: exemplars, active queries, corruption robustness, and task ordering.  

\subsubsection{Exemplar selection and core set extraction}
Before we dive into a quantitative comparison of methods that aim to alleviate catastrophic forgetting through the selection and maintenance of a core set, we need to address a potential evaluation obstacle with respect to the nuances of how a core set is used for continual training. \\

\textbf{Interleaving a core set in training:} In continual learning works, the typical evaluation relies on monitoring the decay of a metric over time when training is conducted on new tasks and old tasks are retained by continued training on a few select exemplars. However, there seemingly is no common protocol of how these exemplars are interleaved. Apart from obvious factors such as the amount of chosen exemplars, works such as variational continual learning \citep{Nguyen2018} use the exemplars only at the end of each task's training cycle to fine-tune and recover old tasks. Most other works \citep{Rebuffi2017, Isele2018, Wu2019} simply concatenate exemplars with newly arriving data. Ultimately, the different works make use of different methods for exemplar selection and attempt to compare their effectiveness through the final metric, even though they are generally not trivially comparable due to their distinct choices of the training procedure. 

\begin{figure}[t]
	\centering
	\includegraphics[width=0.975 \columnwidth]{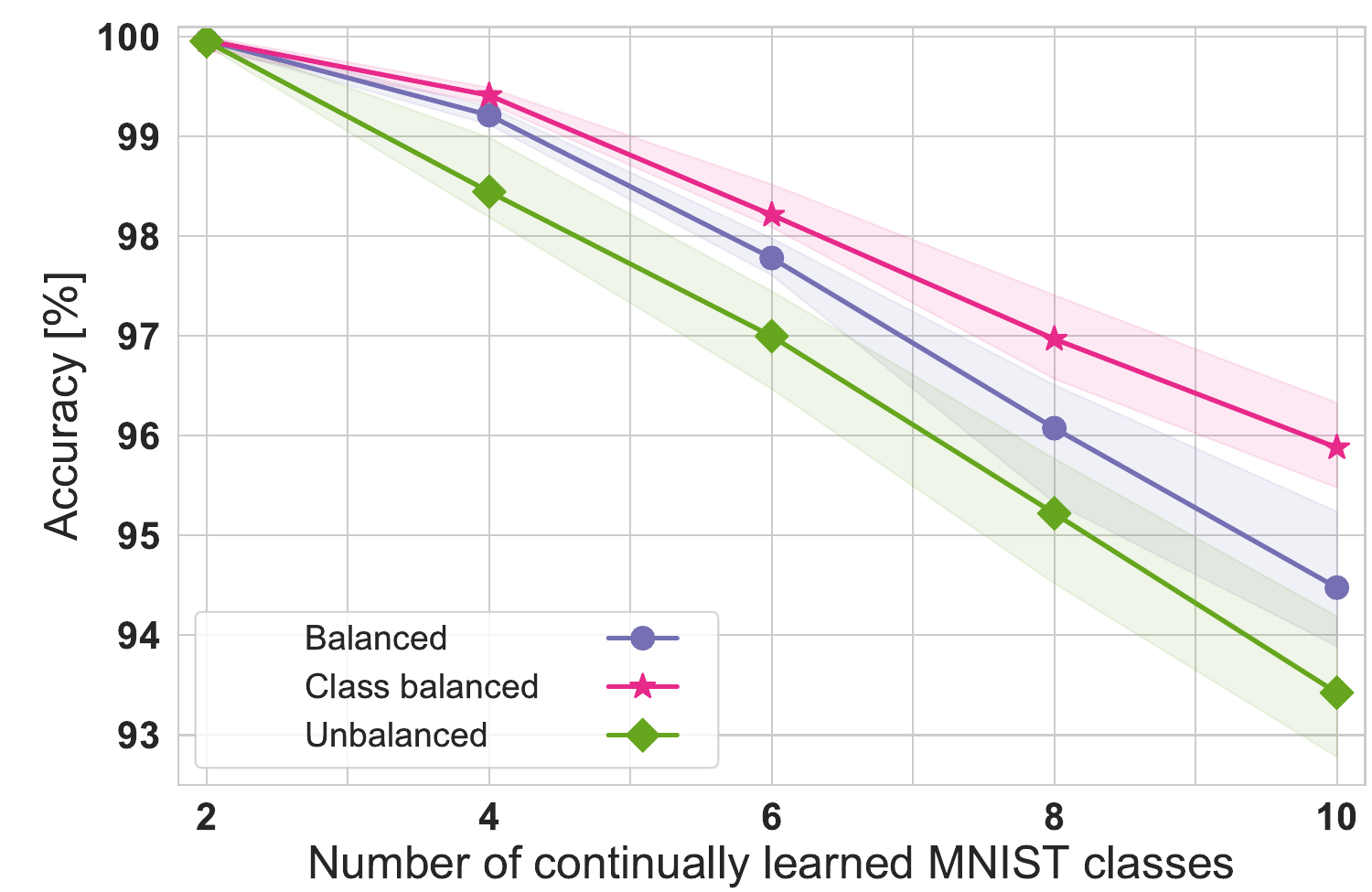} \quad
	\includegraphics[width=0.975 \columnwidth]{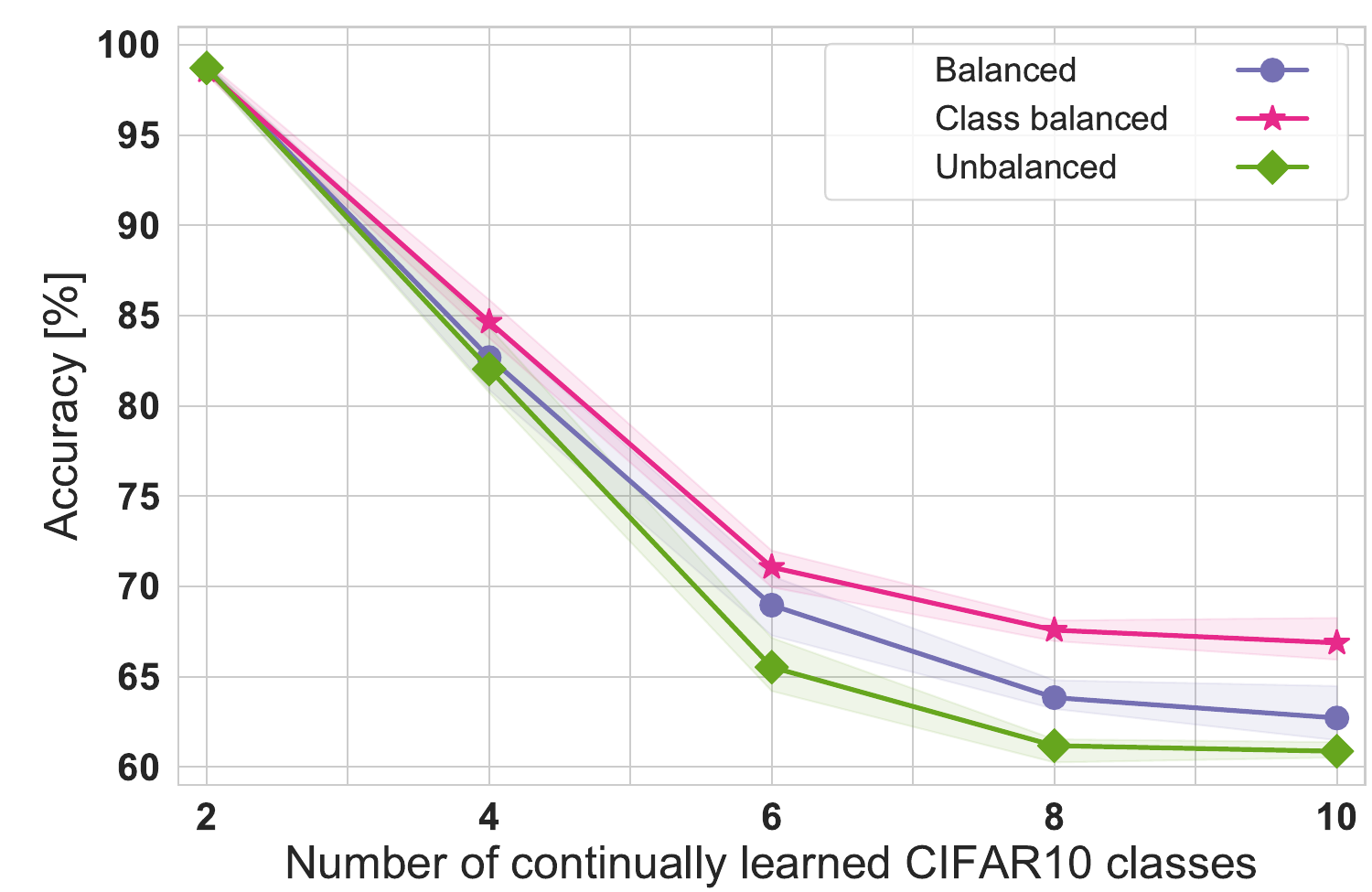}
	\caption{\label{fig:coreset_training_sampling} Influence of mini-batch sampling in continual learning with core sets on MNIST and CIFAR10. The green squared line represents unbalanced sampling, the naive practice of sampling mini-batches uniformly from the concatenated pool of the new task's data and the retained core set. The purple dotted line weights the sampling to oversample the much smaller core set to balance the mini-batch equally. The latter is further corrected with respect to classes in the pink starred line, where the sampling is adjusted to draw mini-batches that are comprised of the same amount of instances per class independently of their origin. We have repeated the experiments five times, illustrated by the shaded regions ranging from the minimum to the maximum obtained values. We can observe that such training details result in very significant performance differences beyond the statistical deviations of a specific core set selection strategy. This imposes an additional challenge in the evaluation of core sets for continual learning. Core sets have been selected with the proposed EVT based method and consist of 240 and 300 exemplars per class for MNIST and CIFAR10 respectively. }
\end{figure}

To highlight this argument we have trained the typical split MNIST and CIFAR10 scenarios, where classes are introduced sequentially in pairs of two and only the new task's data is available to an incrementally growing (single head) classifier. The old task is approximated through a core set of size 2400 and 3000 respectively. That is, we pick 240 and 300 exemplars per class that correspond to retention of 4\% and 6\% of the original data. We train the model for 150 epochs per task to assure convergence and interleave exemplars selected by our proposed EVT approach in three different manners: 
	\begin{itemize}
		\item We conduct the predominant naive concatenation of the core set with the new task's data and continue training with mini-batch gradient descent that samples data uniformly (unbalanced mini-batch sampling).
		\item We recognize that the former combination and sampling leads to a heavy imbalance as the core set size is generally much smaller than the new task's available data. We naively correct this through weighted sampling that samples a mini-batch such that it consists in equal portions of former tasks' exemplars and new task's data. This generally oversamples the exemplars (balanced mini-batch sampling).
		\item We identify that the latter weighted balanced sampling always results in an equal amount of exemplars and new data in a mini-batch. This is independent of the number of classes that the core set or the new task increment are comprised of. To correct for the number of classes, we further investigate class balanced sampling. Here, each mini-batch is sampled such that each class is equally represented. To give an example, if we have seen two tasks of two classes and proceed to learn the next task, the core set with its four classes will be oversampled to constitute two thirds of a mini-batch. The remaining third is made up of the two classes of the third task. 
	\end{itemize}

We show the obtained empirical continual learning accuracies in Figure \ref{fig:coreset_training_sampling}. With gaps of over 5\% it is evident that balancing mini-batches is essential. More so, it is clear that a comparison of different core set works, just because they have used a similar core set size, can result in an apples to oranges comparison if other aspects such as the detailed training procedure are not taken into account. \\

\begin{figure}[t]
	\centering
	\includegraphics[width=0.975 \columnwidth]{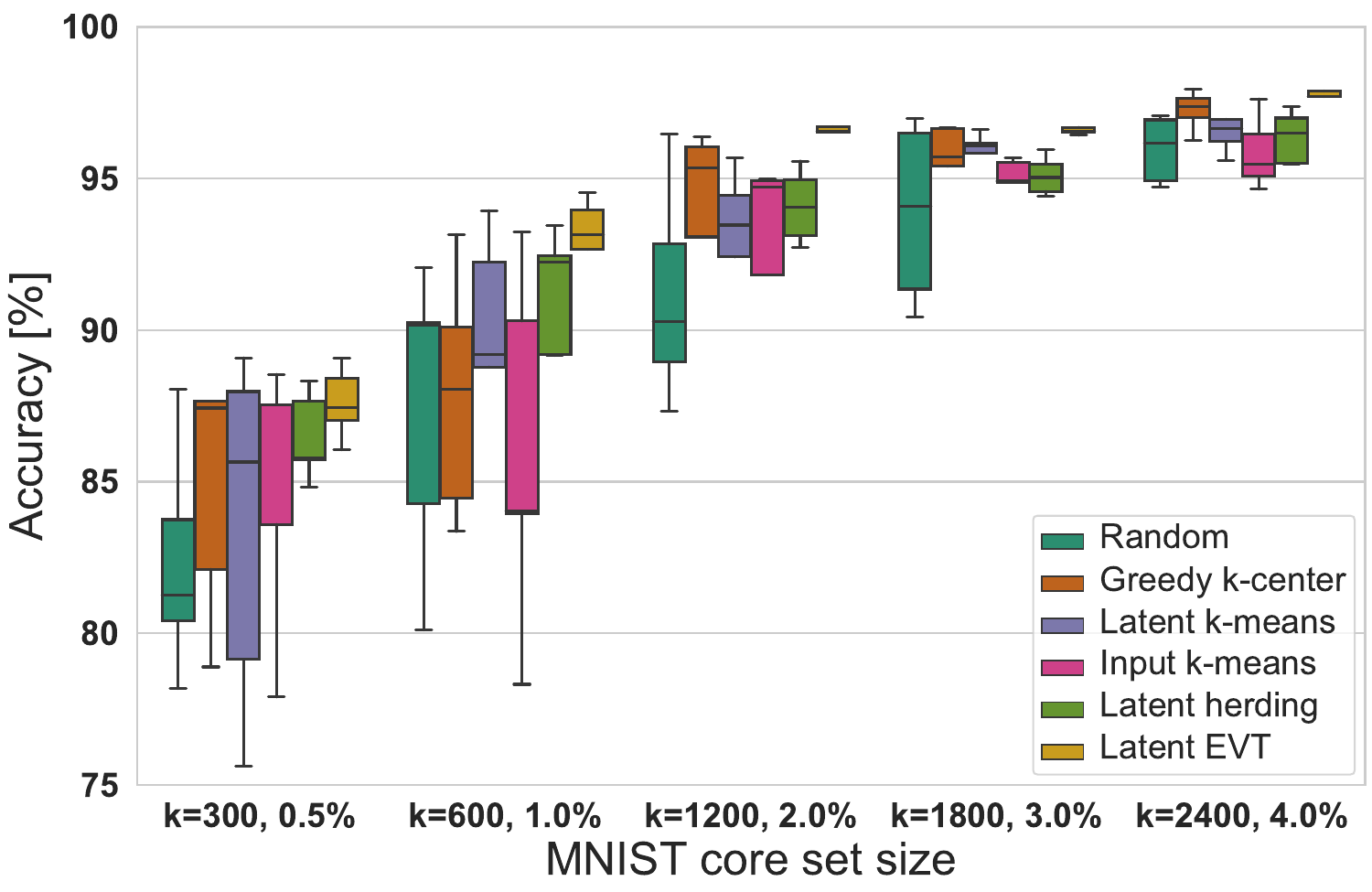}  \\
	\includegraphics[width=0.975 \columnwidth]{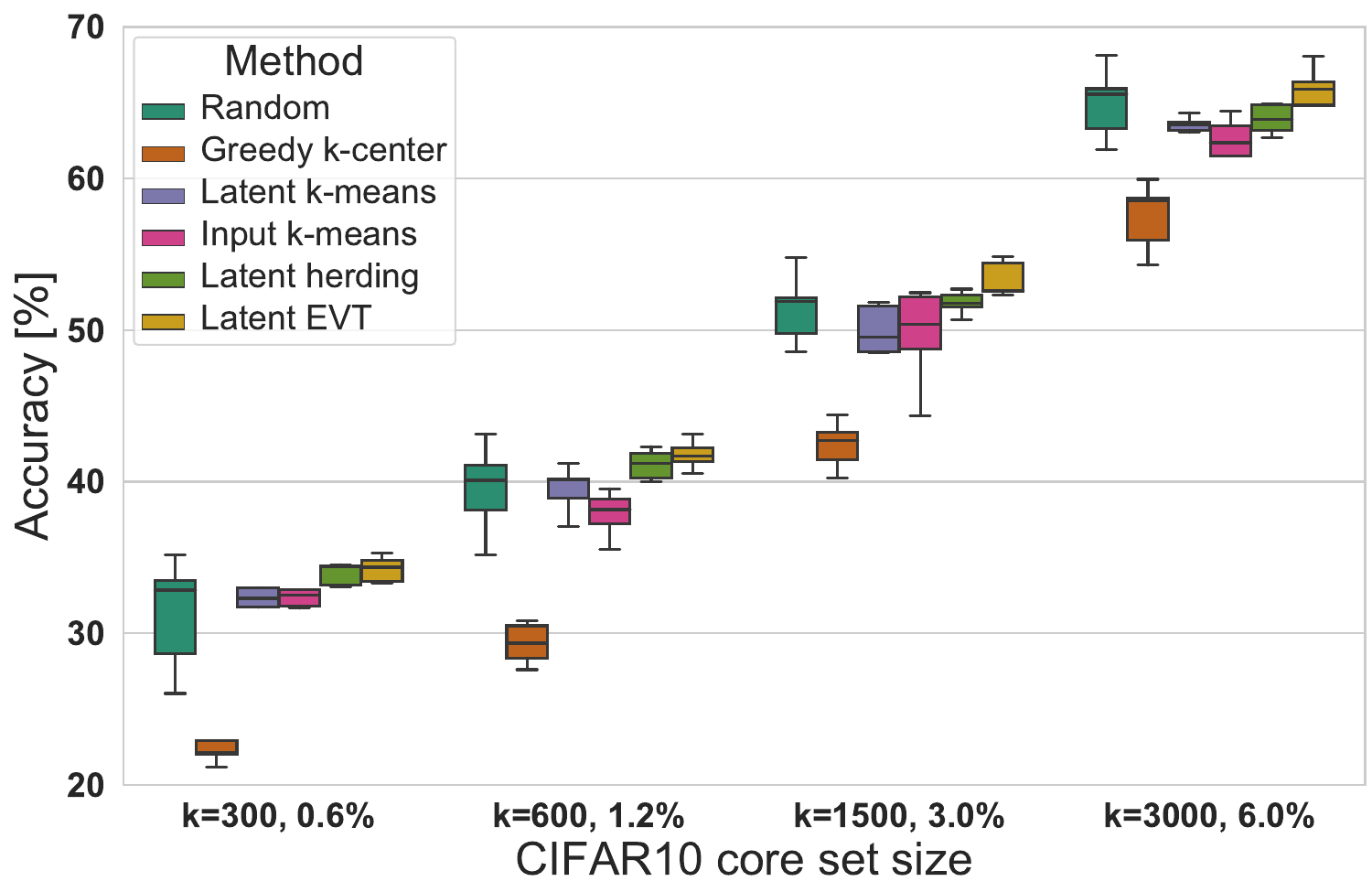}
	\caption{\label{fig:coreset_training} Training accuracy on core sets constructed by different popular strategies. Results for different core set sizes, characterized through their size k and the respective percentage of the dataset, are illustrated in a box plot to show the median, first and third quartile and minimum and maximum values obtained from five experimental repetitions. If viewed without color, methods are displayed from left to right in order of the legend from top to bottom. Best viewed in color.}
\end{figure}

\begin{figure*}[t]
	\centering
	\includegraphics[width=0.975 \columnwidth]{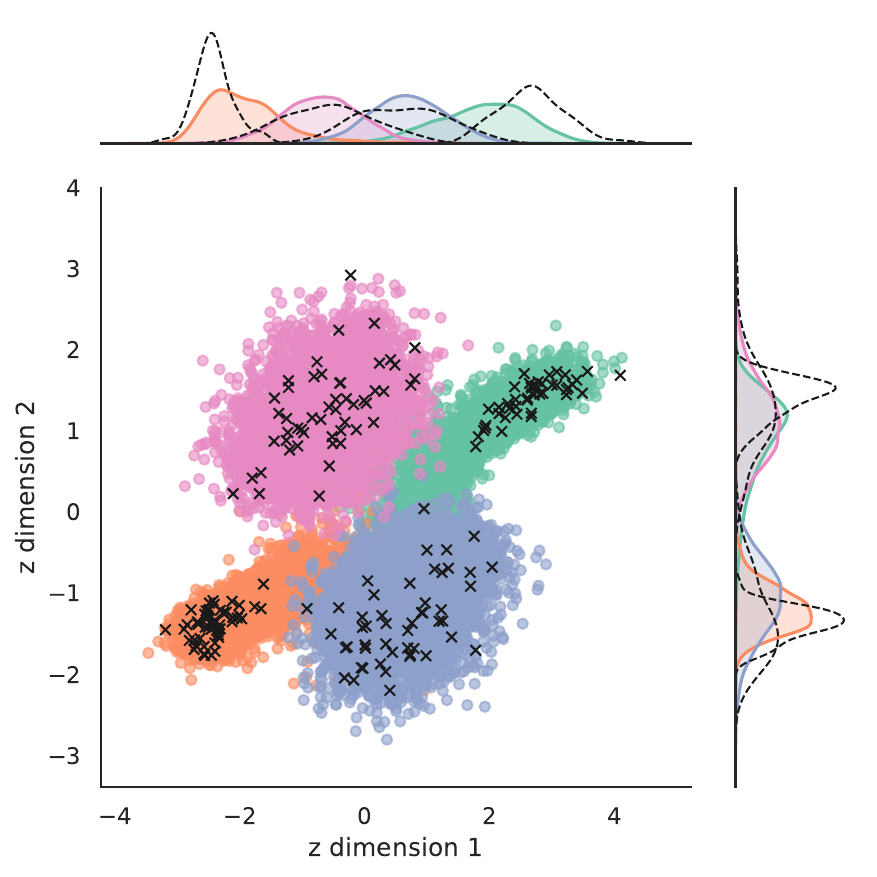}
	\quad
	\includegraphics[width=0.975 \columnwidth]{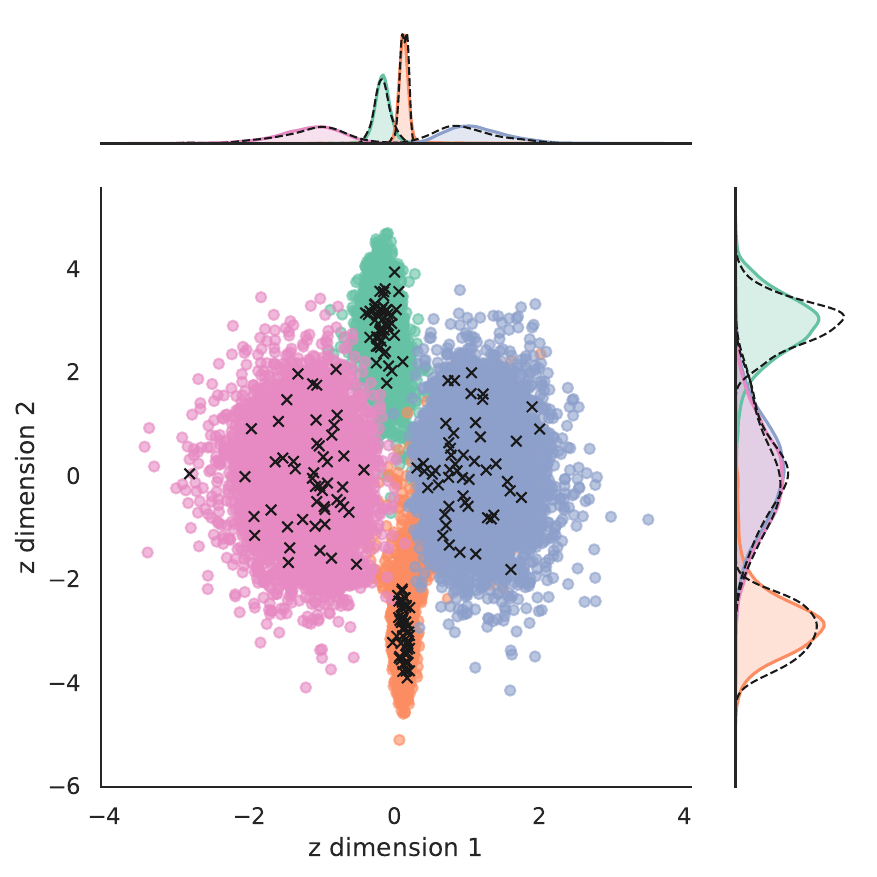}
	\caption{\label{fig:coreset_distribution} Visualization of the aggregate posterior for a model with two-dimensional latent space trained on the first four classes of the CIFAR10 dataset and $200$ selected core set exemplars. The left panel shows the greedy k-center approach, whereas the right panel shows our proposed EVT based core set construction. Classes are color coded points and the core set elements are illustrated through black crosses. A kernel density estimate of the per class aggregate posterior (in color) and the corresponding distributional approximation of the selected core set elements (dashed black) are added on each dimension. In contrast to the greedy k-center approach that features large discrepancies, insignificant differences are observable for our proposed method, painting an intuitive picture for our methods quantitative success of Figure \ref{fig:coreset_training}. Note that in this case the shown ``improved fit'' is not the main goal itself, but rather instrumental to improving long-term learning through core set selection. Best viewed in color.}
\end{figure*}

\textbf{Comparison of core set selection strategies:} Based on the insights of the last paragraph and our main focus on analysis of core set selection strategies, we proceed to compare different strategies in isolation from the precise continual learning training scenario. In analogy to \citet{Bachem2015} and the ``reverse accuracy'' evaluated in LLGAN \citep{Zhai2019}, we first train the model on the entire dataset, then select core sets of different sizes, and finally retrain the model exclusively on the core set to assess the approximation quality of our strategy. We repeat this entire procedure five times to gauge statistical consistency and estimate deviations. Without a doubt, methods that select a core set that yields a better approximation of the overall population and results in larger accuracies when trained in isolation, also provide better means to alleviate catastrophic forgetting in continual learning. We compare six different methods:

\begin{enumerate}
	\item \textbf{Random:} select exemplars uniformly at random.
	\item \textbf{Greedy k-center:} greedy k-center approximation \citep{Gonzalez1985} for core set selection as used in Variational Continual Learning \citep{Nguyen2018}. In essence, exemplars get picked one by one to obtain a cover of the distribution by maximizing their distance in latent space to all existing data points in the core set. 
	\item \textbf{Input k-means:} k-means clustering with k being equal to the number of exemplars. Raw data points get selected that are closest to each obtained mean. Suggested as an alternative to greedy k-center in variational continual learning \citep{Nguyen2018}. 
	\item \textbf{Latent k-means:} analogous to above input based k-means, but with the difference that the clustering is conducted on the lower dimensional latent embedding. 
	\item \textbf{Latent herding:} an adaptation of the herding procedure, used by \citet{Rebuffi2017,Wu2019}, to operate on the latent space instead of an arbitrary neural network feature space. Herding greedily selects exemplars one by one such that each exemplar addition best approximates the overall data's mean embedding. 
	\item \textbf{Latent EVT:} our proposed EVT based inverse Weibull sampling in our common wholistic view.
\end{enumerate}

We show the obtained accuracies by training on differently sized core sets selected by the above mechanisms in Figure \ref{fig:coreset_training}. As expected, random sampling features large variations, with the best attempts rivaling the other methods and in the worst case yielding substantially worse results. The k-means methods both perform similarly, with the latent space version operating on a lower-dimensional embedding showing minor improvements over the clustering obtained on the original image data. The smaller the core set size, the worse these methods seem to perform. This is not surprising and \citet{Bachem2015} have already argued that k-means with well separated clusters with sufficiently different amount of data points per cluster can be prone to inaccurately estimating multiple cluster centers in highly populated areas versus none in more sparsely populated clusters. This is further amplified by k-means generally necessitating a sub-sampled initialization to operate in high dimensions and at large scale. As such, we also observe larger variations for these methods. Latent herding is subject to much less overall variation and seems to initially do very well. However, in contrast to the proposed latent based EVT procedure, we notice an increasing gap in accuracy with larger core set sizes. Intuitively, we attribute this to herding picking increasingly redundant samples due to the objective relying exclusively on the best mean approximation, which does not simultaneously tend to diversity. Our latent based EVT approach that aims to approximate the underlying distribution features the least deviation and consistently outperforms all other methods. \\

\textbf{Intuition behind the strategies:} To provide a better intuition, we have re-trained the model with a two-dimensional latent space to visualize the aggregate posterior and compare it with the selected core sets. Figure \ref{fig:coreset_distribution} shows the a CIFAR10 latent embedding for the first four classes (to promote visual clarity). The colored points correspond to the embedding of the entire set of data points and the respective curves correspond to kernel density estimates of the aggregate posterior. The black crosses indicate the points selected for a small core set of size $200$, i.e. $50$ per class. The left panel illustrates the greedy k-center approach, whereas the right panel shows the EVT aggregate posterior based approximation. Evidently, the approximation of the distribution is adequate for our proposed approach, with the greedy alternative leaving much to be desired. We argue that this is due to the greedy k-center procedure optimizing for a cover based on maximal distances. Such a procedure does not explicitly replicate the density or take into account inherently present outliers and unrepresentative examples. While this quality of fit might not be much of an issue for the highly redundant clean MNIST dataset, the arbitrarily collected real world data of the CIFAR10 dataset entails complete failure for the greedy k-center approach. In fact, by introducing a few naturally occurring image corruptions, we will show that such lack of robustness can be observed for all but our proposed method in a later experiment. Although an improved distribution fit may thus not be the main goal itself, it seems to be instrumental in improving long-term learning.
 
\subsubsection{Active queries}
In addition to the previous experiments showing the advantages in construction of core sets, we empirically demonstrate the benefits when conducting EVT based queries for active learning. Recall that active learning is challenging because we generally desire to query batches of informative data at a time instead of querying, re-training and re-evaluating one by one. This is particularly imperative for computationally expensive deep learning and adds a further constraint of not only querying meaningful samples, but also making sure to query diversely without too much redundancy between the queried examples. We consider this typical deep active learning scenario for MNIST and CIFAR10, where we start with a random subset of 50 and 100 data points respectively, train for 100 epochs to assure convergence and then make a query to include 100 further data points. We then proceed to train the network with the additional instances before repeatedly querying and training again. In a crucial distinction to the majority of active learning works that only investigate the quality of the query by re-training the entire model from scratch, we do not reset our weights in continued incremental training. This implicitly introduces a stronger impact of ordering and further acknowledges that not only labelling, but also training itself is expensive. Each experiment is repeated five times, always with the same initial random subset to preserve comparability between individual repetitions and across methods. \\

\textbf{Comparison of active query strategies:} We investigate popular metrics and mechanisms on which current deep active learning is based. The majority of these are techniques that attempt to take optimal action without explicitly approximating the entire set of unknown data. To estimate and account for uncertainty we make use of Monte Carlo Dropout (MCD) \citep{Gal2015} where appropriate. Although we believe that there is an inherent limitation in earlier introduced approaches that explicitly use the entire unlabelled pool for optimization, we also investigate the proposed technique to query based on a k-means core set extracted from the unknown data \citep{Nguyen2004, Sener2018}. Whereas we certainly regard such methods as valuable in a closed world context, we note that these methods are infeasible without prior knowledge outside of a constrained pool or for sequentially arriving data subsets. As we will see in the next section, they feature little robustness to nonsensical data that might be present in the pool, as the entire unlabelled pool is included and assumed to be useful. The metrics and methods that we investigate are:

\begin{enumerate}
	\item \textbf{Random:} sampling uniformly at random from the unlabelled pool.
	\item \textbf{Reconstruction loss:} in our particular scenario, because our proposed framework includes a generative model, we can query examples based on largest reconstruction loss. This is typically unavailable in a purely discriminative neural network classifier. 
	\item \textbf{K-means core set:} use the entire unlabelled pool to base the query on an extracted core set that is equivalent in size to the query amount. Nguyen et al. had suggested such pre-clustering \citep{Nguyen2004} and it was later used in deep active learning with k-means as the core set algorithm \citep{Sener2018}.
	\item \textbf{MCD - classification confidence:} query based on lowest softmax confidence \citep{Lewis1994}. As neural network classifiers are known to be overconfident, we additionally gauge uncertainty with MCD as a suggested remedy by \citet{Gal2017}.
	\item \textbf{MCD - classification entropy:} query based on largest predictive entropy \citep{MacKay1992}. Similar to lowest confidence, we use uncertainty from MCD to obtain better entropy estimates \citep{Gal2017}. 
	\item \textbf{Latent EVT:} our proposed EVT based approach that balances exploration with exploitation by querying instances that distribute across outlier probabilities, but limited by an upper rejection prior to avoid uninformative outliers. 
\end{enumerate}

\begin{figure}[t]
	\centering
	\includegraphics[width= 0.975 \columnwidth]{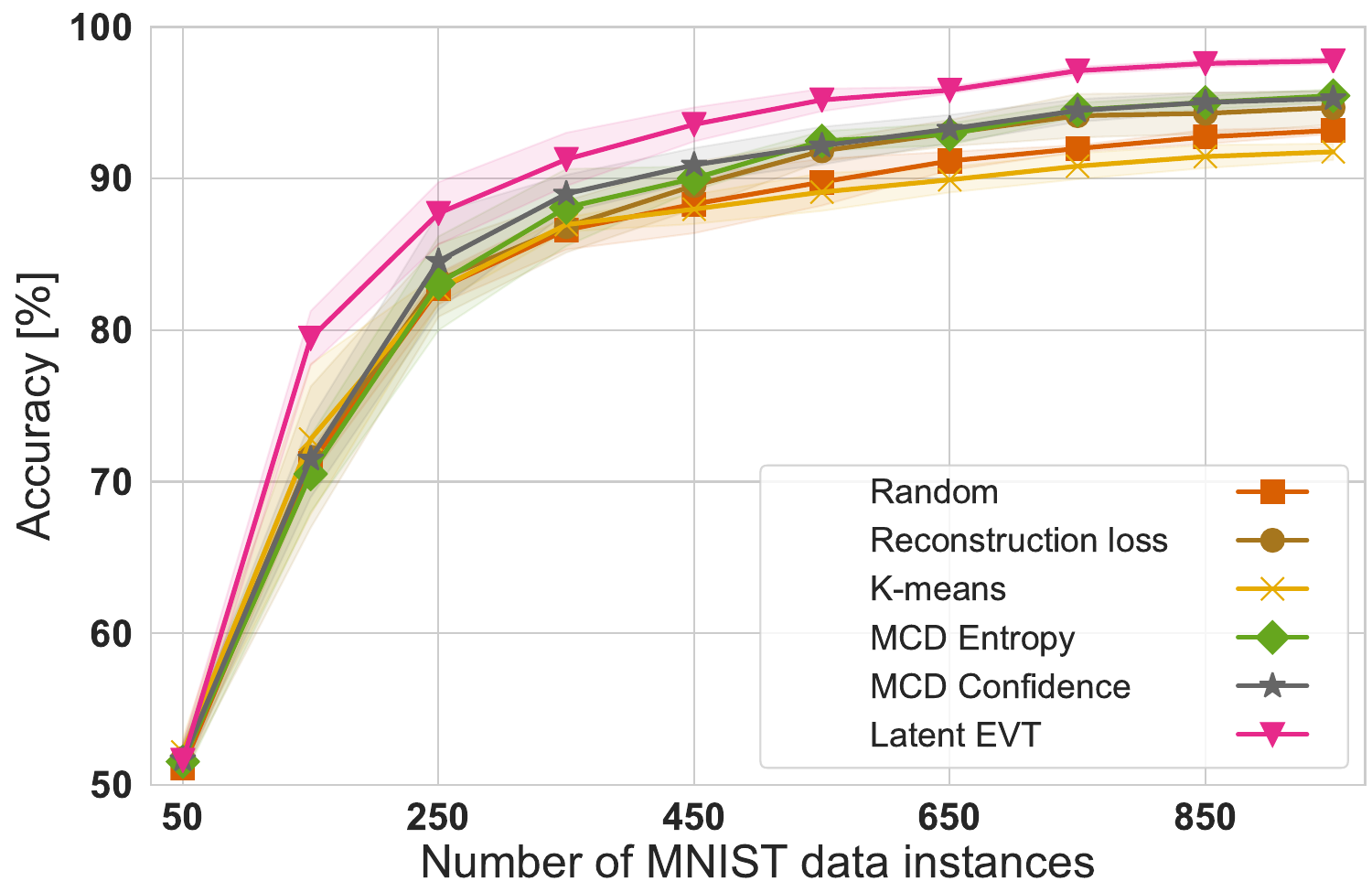} \\
	\includegraphics[width= 0.975 \columnwidth]{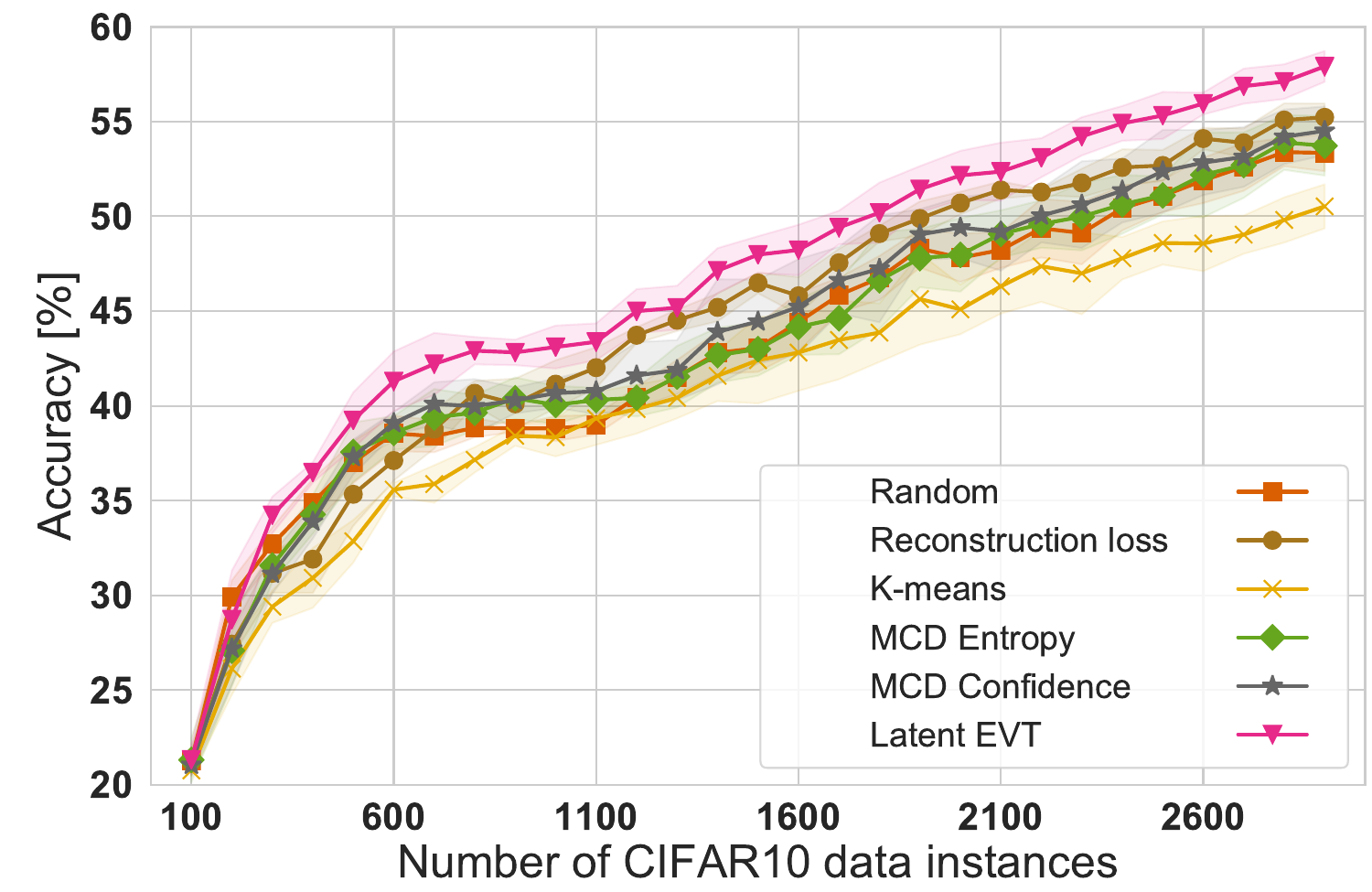}
	\caption{\label{fig:active_learning} Active learning accuracy for different methods on the MNIST and CIFAR10 datasets. All experiments start with the same randomly sampled $50$ and $100$ dataset examples. In each step, an additional $100$ data instances are queried from the remaining unlabelled pool and included for further continued training. Results show the average over five experiments, with the shaded areas ranging from the minimum to the maximum obtained values. }
\end{figure}

We first note that we have included classification confidence and entropy with MCD because omitting uncertainty estimates resulted in no improvement of the active learning query upon simple random selection. This has previously been argued and corresponds to the empirical observations made by \citet{Sinha2019}. For our proposed EVT approach we empirically distribute the query uniformly across examples that fall into the range of 0.5 to 0.95 outlier probability, as estimated by equation \ref{eq:outlier_probability}. Although it never occurred in practice, we note that it would likely be preferential to extend this range to the lower end if not enough samples in the pool were available in the mentioned range, rather than including complete outliers. We will provide empirical evidence for this in the next section. 

Figure \ref{fig:active_learning} shows the quantitative results of our active learning experiments. On both datasets, the k-means based core set is either similar or slightly worse than simply sampling at random. This reflects our previous observations in the core set continual learning section. On the contrary, the uncertainty based methods surpass random sampling. Using largest reconstruction loss similar results can be accomplished, although at the additional computational expense of calculating the decoding. However, all methods are significantly outperformed by our proposed latent EVT method at all times. The respective rationale behind this improvement is quite intuitive. In contrast to the considered baselines, which have a sole focus on novelty, our strategy balances completely novel examples with less novel examples that are still required to strengthen the existing learned features. More importantly, it rejects uninformative outliers that are inherently present in the pool, a threat that uncertainty based methods can be particularly prone to. This threat is magnified with even less knowledge about the acquired dataset and even more unconstrained data acquisition.

\subsubsection{Robustness to open world corruptions} 
The past two experimental subsections have focused on showing our method's advantage in the typical continual and active learning benchmark perspective in the closed world scenario, devoid of any analysis with respect to robustness. In our prior work \citep{Mundt2022,Mundt2019}, we have empirically demonstrated that the proposed EVT based approach can successfully distinguish between known and unknown sets of classes (recall the earlier Figure \ref{fig:openset}), which is otherwise difficult due to overconfident misprediction \citep{Scheirer2013,Scheirer2014,Bendale2015,Bendale2016,Boult2019}.We will now investigate a perhaps equivalently large threat: data that is statistically deviating for other reasons, such as corruption and perturbation. Again, we illustrate common limitations and the advantages of our wholistic view. However, we also note that robustness to corrupted data is merely a necessity, not sufficient, for robustness to truly any out-of-sample data. Again, we have picked the example of corrupted data as it provides a solid teaching foundation to highlight and support our previously made arguments. \\

\textbf{Choice of corruptions and perturbations:} In a recent effort to benchmark the performance against 15 types of various corruptions, \citet{Hendrycks2019} have shown that none of the developed neural network models feature any intrinsic robustness, even if they converge to more accurate solutions on the initial benchmark. This was concluded from experiments where neural networks are trained on the uncorrupted benchmark dataset and evaluated on the corrupted data. We extend this evaluation by investigating the presence of a minor portion of corrupted data in the training process, as can realistically be assumed for active or continual learning. We examine whether common query strategies in active learning and core set construction in continual learning are robust, or whether querying and including this unrepresentative corrupted data into core sets leads to performance degradation in comparison with the clean benchmark. We believe that this is critical for two reasons: 1.) The necessity to carefully curate every single example in the unknown data pool can outweigh the active learning human labelling effort and thus renders active learning ineffective in the first place. 2.) Data cleaning itself is extremely challenging and it is often not immediately clear whether the inclusion of a data instance is beneficial or is accompanied by side effects.

We make use of corruptions across four categories: noise, blur, weather and digital corruptions, as introduced by \cite{Hendrycks2019}. These can further be distinguished into 15 types: low-lighting Gaussian noise, electronic shot noise, bit error impulse noise, speckle noise, Gaussian blur, defocus blur, glass blur, zoom blur, motion blur, snow, fog, brightness, contrast, saturation and elastic deformations. Each corruption is algorithmically generated with five discretized levels of severity, of which the first two are at times barely discernible from a typical image by a human. We accordingly corrupt 7.5\% of the data across these 75 corruptions. We add the additional constraint that each image can only be corrupted once. Note that in principle some corruptions, such as noise resulting from low lighting conditions and out of focus blurring, could occur simultaneously. We have deliberately chosen this amount of corruption to, on the one hand be small enough to not affect overall performance if trained on the entire dataset, on the other hand be larger than the core set size or active learning query amounts used in previous sections. Hypothetically, in the absolute worst case this could result in only corrupted images being selected and the entire chosen set being much less representative of the complete dataset than a selection of clean examples would be. We repeat the previous CIFAR10 experiments under these conditions. \\

\begin{table}[t]
\resizebox{0.975 \columnwidth}{!}{%
\begin{tabular}{ccc|cc|cc}
 & \multicolumn{6}{c}{\textbf{Accuracy [\%]}: mean $\tiny{\substack{+\texttt{difference to maximum} \\ -\texttt{difference to minimum}}}$ }\\
\cmidrule{2-7}
\textbf{CIFAR10 queries, dataset size} & \multicolumn{2}{c}{8, 900} & \multicolumn{2}{c}{18, 1900} & \multicolumn{2}{c}{28, 2900} \\ 
\textbf{Dataset} & regular & corrupted & regular & corrupted & regular & corrupted \\ 
\toprule
\textbf{Random} & $38.80\tiny{\substack{+0.69 \\ -1.75}}$ & $38.97\tiny{\substack{+1.03 \\-1.87}}$ & $47.81\tiny{\substack{+2.02 \\ -3.93}}$ & $47.91 \tiny{\substack{+2.13 \\ -3.58}}$ & $53.36\tiny{\substack{+1.17 \\ -2.34}}$ & $53.53\tiny{\substack{+1.13\\ -2.42}}$ \\
\textbf{Reconstruction loss} & $41.14\tiny{\substack{+2.06 \\ -3.89}}$ & $38.26\tiny{\substack{+0.64 \\-1.89}}$ & $50.70\tiny{\substack{+0.69 \\ -1.50}}$ & $46.49 \tiny{\substack{+0.82 \\ -2.13}}$ & $55.22\tiny{\substack{+1.37 \\ -1.92}}$ & $50.85\tiny{\substack{+1.03\\ -1.57}}$ \\ 
\textbf{K-means} & $38.34\tiny{\substack{+1.46 \\ -2.63}}$ & $36.05\tiny{\substack{+1.65 \\-2.53}}$ & $45.08\tiny{\substack{+1.50 \\ -3.23}}$ & $42.93 \tiny{\substack{+1.59 \\ -3.65}}$ & $50.52\tiny{\substack{+0.94 \\ -3.15}}$ & $47.58\tiny{\substack{+1.93\\ -3.39}}$ \\ 
\textbf{MCD Entropy} & $40.05\tiny{\substack{+1.15 \\-2.99}}$ & $38.83\tiny{\substack{+0.68 \\-1.03}}$ & $47.96 \tiny{\substack{+2.91 \\ -5.28}}$ & $44.73 \tiny{\substack{+0.61 \\ -1.02}}$ & $53.72\tiny{\substack{+2.35 \\ -4.76}}$ & $50.06\tiny{\substack{+0.37\\ -0.75}}$ \\ 
\textbf{MCD Confidence} & $40.67\tiny{\substack{+0.87 \\ -1.89}}$ & $37.93\tiny{\substack{+0.35\\-0.81}}$ & $49.40\tiny{\substack{+2.86 \\ -4.44}}$ & $47.16 \tiny{\substack{+1.29 \\ -3.22}}$ & $54.51\tiny{\substack{+1.15 \\ -3.13}}$ & $51.91\tiny{\substack{+1.78\\ -2.67}}$ \\
\textbf{Latent EVT} & $44.67\tiny{\substack{+0.32 \\ -0.63}}$ & $43.79\tiny{\substack{+0.74 \\-1.72}}$ & $51.66\tiny{\substack{+1.05 \\ -1.69}}$ & $51.12 \tiny{\substack{+0.38 \\-0.91}}$ & $57.43\tiny{\substack{+0.51 \\ -1.09}}$ &  $56.83\tiny{\substack{+0.41 \\-0.78}}$\\   
\bottomrule
\end{tabular}}
\caption{\label{tab:active_learning_corruption} Active learning with and without partial dataset corruption. Uncorrupted values correspond to those visualized in Figure \ref{fig:active_learning}.}
\end{table}
\begin{table}[t]
\resizebox{0.975 \columnwidth}{!}{%
\begin{tabular}{ccc|cc|cc}
 & \multicolumn{6}{c}{\textbf{Accuracy [\%]}: mean $\tiny{\substack{+\texttt{difference to maximum} \\ -\texttt{difference to minimum}}}$ }\\
\cmidrule{2-7}
\textbf{CIFAR10 core set size} & \multicolumn{2}{c}{300} & \multicolumn{2}{c}{600} & \multicolumn{2}{c}{1500} \\ 
\textbf{Dataset} & regular & corrupted & regular & corrupted & regular & corrupted \\ 
\toprule
\textbf{Random} & $31.23\tiny{\substack{+3.94 \\ -9.14}}$ & $30.35\tiny{\substack{+1.88 \\ -5.92}}$ & $39.52\tiny{\substack{+3.61 \\ -7.95}}$ & $39.05\tiny{\substack{+1.99 \\ -5.89}}$ & $51.43\tiny{\substack{+3.33 \\ -6.12}}$ & $51.01\tiny{\substack{+2.30 \\ -4.49}}$ \\ 
\textbf{Greedy k-center} & $22.82\tiny{\substack{+3.05 \\ -1.65}}$ & $22.19\tiny{\substack{+1.76 \\-3.37}}$ & $29.33\tiny{\substack{+1.50 \\ -3.23}}$ & $29.48\tiny{\substack{+1.91 \\ -5.11}}$ & $42.41\tiny{\substack{+1.97 \\ -4.13}}$ & $42.37\tiny{\substack{+1.49 \\ -2.44}}$ \\ 
\textbf{Latent k-means} &  $32.76\tiny{\substack{+2.29 \\ -3.35}}$ & $29.00\tiny{\substack{+2.12 \\ -4.05}}$ & $39.49\tiny{\substack{+1.71 \\ -4.17}}$ & $35.71\tiny{\substack{+1.69 \\ -4.08}}$ & $50.01\tiny{\substack{+1.80 \\ -3.28}}$ & $48.52\tiny{\substack{+2.59 \\ -3.86}}$ \\ 
\textbf{Image k-means} & $32.85\tiny{\substack{+2.57 \\ -3.76}}$ & $30.74\tiny{\substack{+1.43 \\ -3.16}}$ & $37.86\tiny{\substack{+1.66 \\ -3.98}}$ & $36.38\tiny{\substack{+0.90 \\ -2.75}}$& $49.62\tiny{\substack{+2.83 \\ -8.09}}$ & $48.23\tiny{\substack{+1.78 \\ -2.50}}$ \\ 
\textbf{Latent herding} & $33.92\tiny{\substack{+0.61 \\ -1.45}}$ & $33.81\tiny{\substack{+0.82 \\ -1.39}}$ & $41.13\tiny{\substack{+1.18 \\ -2.29}}$ & $40.77\tiny{\substack{+1.34 \\ -1.57}}$ & $51.87\tiny{\substack{+1.12 \\ -1.85}}$ & $51.06\tiny{\substack{+2.43 \\ -2.30}}$ \\ 
\textbf{Latent EVT} & $34.16\tiny{\substack{+1.10 \\ -2.27}}$ & $34.18\tiny{\substack{+1.07 \\ -2.55}}$ & $41.78\tiny{\substack{+1.34 \\ -2.57}}$ & $41.67\tiny{\substack{+1.37 \\ -2.53}}$ & $53.35\tiny{\substack{+1.48 \\ -2.53}}$ & $53.28\tiny{\substack{+1.06 \\ -2.17}}$ \\ 
\bottomrule
\end{tabular}}
\caption{\label{tab:coreset_corruption} Core set selection and subsequent training with and without dataset corruption. Uncorrupted values correspond to those visualized in Figure \ref{fig:coreset_training}.}
\end{table}

\textbf{Comparison of core set selection and active query strategies in presence of corrupted data:} We show the originally obtained results in direct comparison with the results obtained under inclusion of the corrupted data in Tables \ref{tab:active_learning_corruption} and \ref{tab:coreset_corruption}. For better visualization and quantification we do not show plots, but have instead picked three evenly spaced points of Figure \ref{fig:coreset_training} and \ref{fig:active_learning}. From these quantitative results it is evident that only two techniques are robust in active learning: random sampling and our proposed EVT based approach. The logical explanation is that random sampling on average will pick roughly 7.5\% corrupted data, of which another 40\% feature only minor low severities. The small amount thus only has minor effect on the optimization. The EVT based algorithm is similarly unaffected as it does not query statistical outliers in the first place. If it includes corrupted examples then only those with minor severity that are statistically still largely similar to the uncorrupted data. All other methods are prone to the corrupted outliers in one way or another. Classifier uncertainty and reconstruction loss tend to pick very corrupted examples by definition. The k-means approach will have shifted centers or falsely query from new clusters that are centered around corruptions of the unknown pool. Looking at the quantitative accuracy values, we can in fact even conclude that all these methods perform worse than a simple random query. The continual learning core set construction picture is quite similar. Here, we can observe corruption robustness for random sampling, latent herding and our proposed approach. Latent herding is robust to outliers because it picks samples greedily one by one to best approximate the mean, which intuitively involves picking the next best example that is close to the class mean and does not involve outliers. However, the issue of including redundant samples into the core set remains unaddressed, and our EVT based method nevertheless outperforms all other approaches. \\

\begin{figure}[t]
	\centering
	\includegraphics[width=0.45 \columnwidth]{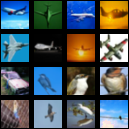} \quad
	\includegraphics[width=0.45 \columnwidth]{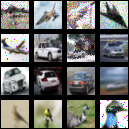} 
	\caption{\label{fig:coreset_corruption} Typically selected dataset examples in the core set construction using a greedy k-center algorithm. Qualitative illustration is intended to provide intuition for a method's failure. The left panel shows how picked exemplars from an uncorrupted dataset are unrepresentative of the average image, with unusual backgrounds, occlusion and scaling issues. The right panel shows how the core set is comprised of many corrupted examples if a small portion of the dataset is corrupted, a lack of robustness that many methods in Table \ref{tab:active_learning_corruption} and \ref{tab:coreset_corruption} suffer from.}
\end{figure}

\textbf{The qualitative intuition:} Interestingly, the greedy k-center approach also seems to be robust to the corruptions, although it performs equally miserably to the uncorrupted scenario. Recall that this algorithm greedily chooses the next data point for inclusion in a farthest-first traversal, by maximizing the distance to all presently existing core set elements. In other words, outliers are always queried as they are farthest away by definition. Only after a sufficiently large cover is obtained will representative data be queried. Because such unrepresentative outliers are already present in the uncorrupted data, the performance is consequently always low for small core set sizes. To visually illustrate this statement we show a uniform sub-sample of the acquired core set for the first four classes with and without corruption in Figure \ref{fig:coreset_corruption}. In the left panel we can observe the core set being comprised of atypical aeroplanes with deep green or black background, a captured overexposed sunset. There are partially occluded cars and birds close to bushes and fences or images where the animal is almost not discernible and comprises only a fraction of the image. Arguably these do not represent good exemplars. In the right panel, we can see that in the presence of corruption, the core set is comprised of noisy, blurry and otherwise distorted images. Ultimately neither of these core sets are particularly good dataset approximations, intuitively explaining the poor performance of this technique. 

\subsubsection{The effect of task order and curricula}
In a final set of experiments, we investigate the importance of orderings and whether the construction of a curriculum beyond alphabetical class order provides substantial learning benefits. We briefly re-emphasize that we primarily conduct this study to investigate the effects of order on continual learning (or pre-training), and not as a suggested primary mechanism to alleviate catastrophic forgetting. As previously noted, we leave the practical discussion of which continual learning scenarios allow for the control of a curriculum for different works. \\

\textbf{Comparison of order selection strategies:} We consider four conceivable scenarios:

\begin{enumerate}
	\item \textbf{Class sequential ordering:} learn the classes in order of their integer class label. For many datasets this is in alphabetical order. 
	\item \textbf{Random order:} randomized class order. 
	\item \textbf{Most outlying, dissimilar tasks first:} determine the next class to add by evaluating equation \ref{eq:task_order}, i.e. pick the next class that is most outlying and dissimilar with respect to the already seen classes.
	\item \textbf{Most inlying, similar tasks first:} determine the next class to add by evaluating equation \ref{eq:task_order}, but with a minimum over task outlier probabilities to include the most similar task in each increment.
\end{enumerate}

Note that for all strategies we always start with the same first task for comparability. To make sure that obtained results and found curricula are not just a result of sheer luck, we repeat each experiment five times. We then report the average and the minimum and maximum obtained accuracies at each step to gauge deviations. We conduct experiments on two datasets: the CIFAR100 and the AudioMNIST \citep{Becker2018} dataset. We follow the typical continual incremental learning procedure of adding classes in pairs of two. We chose the first dataset because it allows for the construction of a long task sequence. We chose the latter because it represents a non-image dataset. Previous work has observed that some classes can provide strong retrospective improvement \citep{Mundt2022}, an early indicator that the class ordering should be investigated further. In order to highlight the effect ordering can have on our system, we provide a two-fold analysis. We analyze ordering both, when independently evaluated from, or coupled to specific techniques that alleviate continual learning catastrophic forgetting. As such, we evaluate CIFAR100 in what is typically referred to as a continual learning upper-bound. The latter describes the maximum obtainable accuracy given a specific model choice and training procedure in which the data of each task is simply accumulated with each subsequent task. For the AudioMNIST we use generative replay to prevent catastrophic forgetting, where old tasks' data is rehearsed based on the trained generative model. We do not make use of any data augmentation. \\

\begin{figure}
	\centering
		\includegraphics[width=0.975 \columnwidth]{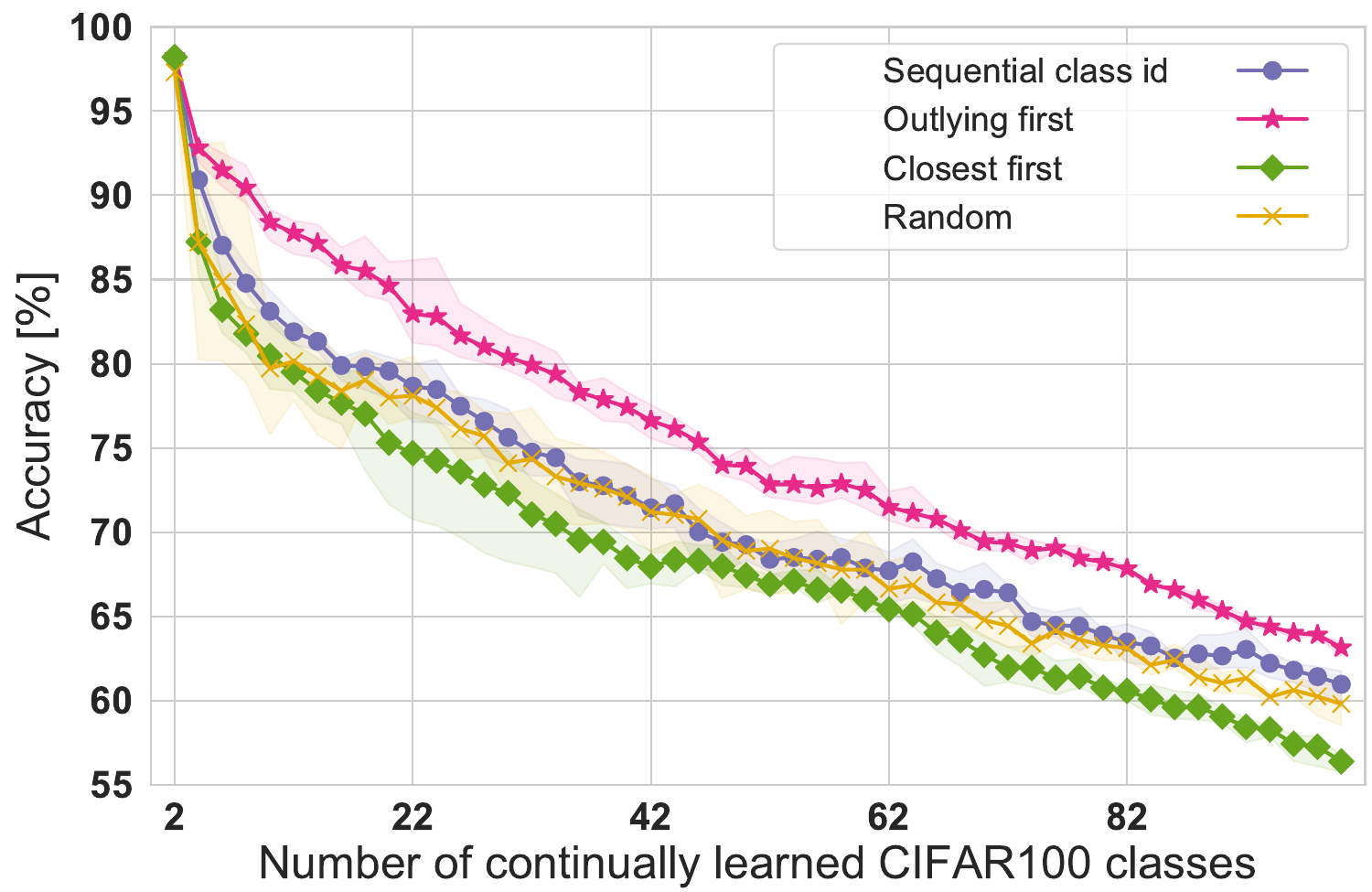} \\
		\includegraphics[width=0.975  \columnwidth]{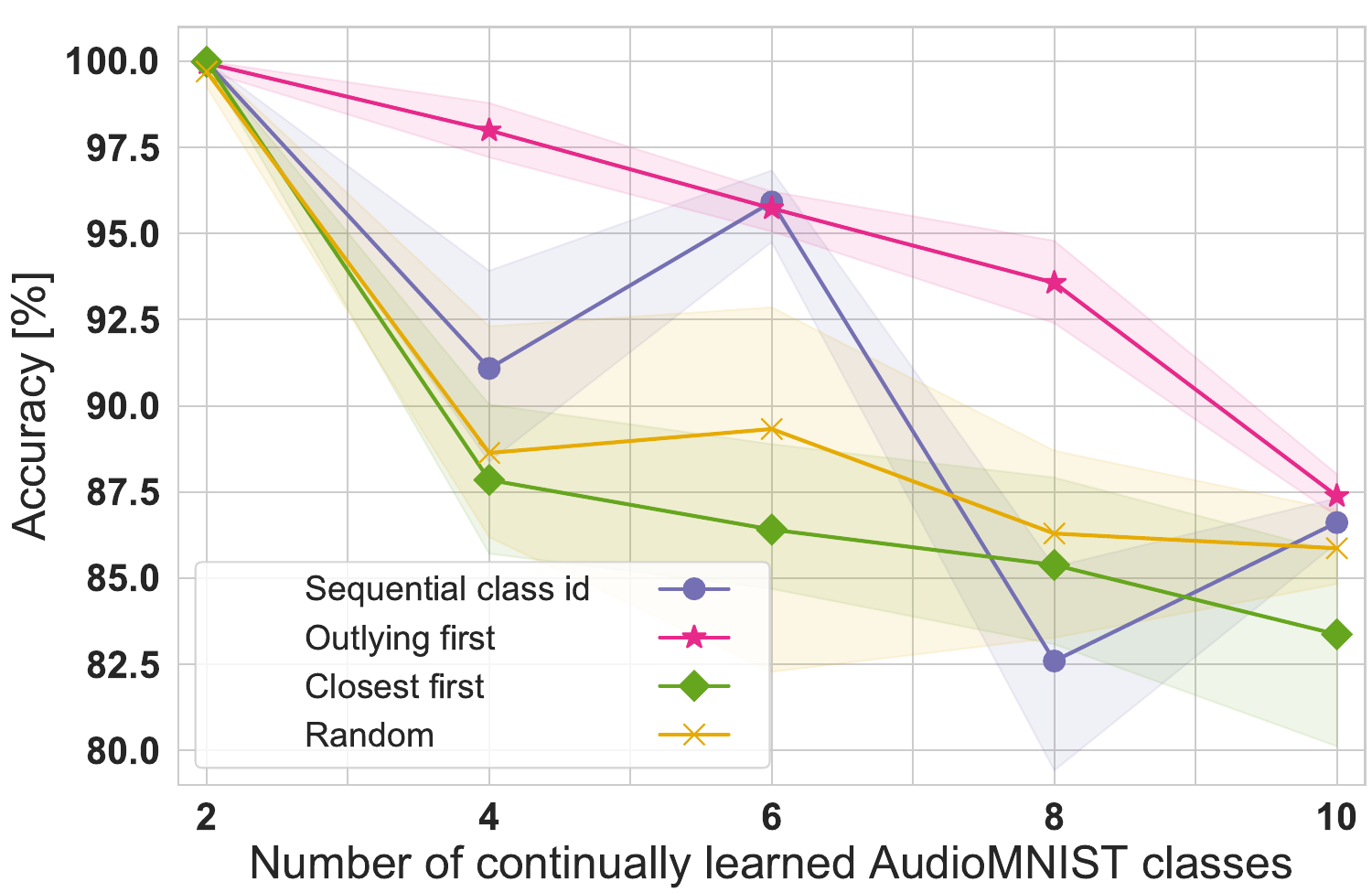}
	\caption{\label{fig:task_order}Continual learning accuracy of learning classes in increments of two in dependence on the choice of task order. Top panel shows the incremental upper-bound, i.e. a simple accumulation of the real data, for the CIFAR100 dataset. The bottom panel shows obtained performance on the AudioMNIST dataset with alleviated catastrophic forgetting through generative replay. For each of the order selection mechanisms the experiment has been repeated five times. The corresponding average together with the maximum and minimum deviation are reported respectively.}
\end{figure}

\textbf{Empirically observed effects of order:} The achieved accuracies at each task increment are shown in Figure \ref{fig:task_order}. We can observe that for the CIFAR100 dataset, random sampling seems to yield a very similar accuracy trajectory in comparison to sequentially learning the classes in order of their alphabetical class id, resembling earlier observations \citep{DeLange2019, Javed2018}. However, our observations based on our proposed framework's selection scheme seem to be in contrast to their conclusion that the order in which tasks are introduced is negligible. Here, selecting the most dissimilar task for inclusion consistently improves the accuracy by several percent, even at the end of training. Conversely, including tasks that are very proximate to existing concepts results in an all-time performance decrease. We hypothesize that this is due to the classifier experiencing immediate confusion. Our initial classes consist of ``apples'' and ``aquarium fish'' and the query consensus across repeated experiments is to continue with selecting the classes ``pears'' and ``whale'' or ``shark''. The opposite strategy that prioritizes dissimilarity in the curriculum instead includes unrelated classes such as ``lawnmower'', ``mountain'' or ``oak''. We believe that this allows the model to more rapidly acquire a diverse set of representations.

We can draw almost analogous conclusions for continually learning the AudioMNIST dataset with generative replay. Here, we additionally see that the conventional order of learning the sounds from ``zero'' to ``nine'' is accompanied by a pattern of repeated retrospective improvement. The first task increment results in a larger accuracy drop, that is rectified through backwards improvement of the next task increment. This pattern repeats for the next two classes and its consistent strong emergence is only visible when learning sequentially in order of class id. The accuracy at any time is again best for our proposed measure of dissimilarity and worst when selecting according to task proximity. For the latter, in analogy to the earlier hypothesized confusion of the classifier, the generative model is faced with difficulty to disambiguate the resembling classes and produce unambiguous output. \\
 
Our results indicate that using active learning techniques and taking into account the learning order can play a critical impact on the achieved performance of our continual learning set-ups. More so, the results provide an important signal for reproducibility and significance of various conjured continual learning benchmarks. In a world of benchmarking methods and regularly claiming advances when a method surpasses another by 1-2 \%, the observed absolute discrepancy between the different task orders for CIFAR100 is as large as 10\%. This is a substantial gap. Whereas we obviously believe that there lies value in analyzing and contrasting different techniques to alleviate catastrophic forgetting on a common dataset, it is clear that there is still much we need to learn about neural network training, the role of curricula, the importance of what has already been trained in continual learning, and evaluation in general. A more in-depth understanding is likely to develop by adopting more wholistic approaches to our systems, but will also need to be supported by moving away from our current rigid benchmarks. 

\section{Discussion: towards wholistic deep continual learning}
We have empirically corroborated our wholistic view in the previous experimental section, highlighted on a set of teaching examples and one conceivable practical realization of a unified framework. To conclude our paper, we now circle back to the preamble of the paper. Building on our wholistic view's insights, we revisit the ``definitions'' for continual machine learning and its role in presenting a guideline for evaluation protocols. We then discuss the entailed prospects of adopting our wholistic view and present some remaining limitations.

\subsection{A revisited continual machine learning definition as a guideline}
Taking into account the insights we have obtained through our wholistic view, we can now ask ourselves the natural question whether the initial ``definitions'' for continual machine learning \ref{def:ThrunLLML} and \ref{def:ChenLiu} require to be adapted. On the one hand, one could now argue that there is nothing technically wrong with the latter existing definition. After all, despite perhaps having a larger amount of ambiguity, observed flaws and limited evaluation are an issue with practically employed protocols. In a sense, the definition is likely intentionally abstract and not mathematically rigorous. A too intricate definition could come with the danger of being overly specific, resulting in conceivable exclusion of relevant factors and unnecessarily constrained scenarios. On the other hand, it may be precisely this lingering ambiguity that can result in misinterpretation or oversight of potentially important elements. Ultimately, every definition, be it mathematically rigorous or not, also plays a role in guiding research. If major factors are left to be assumed, they may easily be missed.
 
As a balance between these two points, we propose to extend the existing definition of \citet{Chen2017}, yet keep a similar level of abstraction. In this way, we can add several important factors for continual learning, as a guideline to researchers that these aspects are relevant, and at the same time avoid an excessive amount of introduced constraints. Our proposition for such a revised description ensues:

\begin{definition}{Continual Machine Learning - this work:} \small
The learner performs a sequence of \textit{N} continual learning tasks, $\mathcal{T}_1, \mathcal{T}_2, \ldots, \mathcal{T}_{\mathit{N}}$, that are distinct from each other in terms of shifts in the underlying data distribution. The latter can imply a change in objective, transitions between different domains or inclusion of new modalities. At any point in time, the learner must be able to robustly identify unseen unknown data instances. Depending on what is permissible in application contexts, the learner can either reject such instances in a non-controllable data stream or set them aside for later learning. In the latter scenario, the learner should be able to rank order unknowns according to similarity with existing tasks, in order to actively build a meaningful learning curriculum itself. If the system is desired to be supervised, a human in the loop may group and label the set of identified unseen unknowns to explicitly guide future learning.
When faced with a selected (\textit{N}+1)th task $\mathcal{T}_{\mathit{N}+1}$ (which is called the new or current task) with its data $\mathcal{D}_{\mathit{N}+1}$, the learner should leverage its dictionary of representations to accelerate learning of $\mathcal{T}_{\mathit{N}+1}$ (forward transfer), extend the dictionary with unique representations obtained from the new task's data (this can be completely new types of dictionary elements), while simultaneously maintaining and improving the existing representational dictionary with respect to former tasks (backward transfer). 
\end{definition}

In direct comparison with former continual learning definitions, introduced in the preamble of this paper, the description is now extended to include active data queries, captures the importance of data choice and curricula (if controllable by the learner), in coherence with awareness of the open world. In particular, note that the definition now explicitly includes the idea of an open world and required robustness. Following our previous exposition, we deem this to be a general requirement for any machine learner, with particularly manifesting importance in continual learning. In contrast, the notion of setting aside data instances for prospective learning to build up a self-selected learning curriculum is included to emphasize its importance and impact, yet left optional depending on considered level of data stream control. In this way, the potential elements of continual learning are captured without strictly dictating specific set-ups, following the same spirit as the original definition.

\subsection{Prospects}
We anticipate that our work leads to increased awareness of the dangers of our current closed world practices and the necessity of expanding our views towards more realistic real-world relevant evaluation. In doing so, we believe that further synergies between presently separately treated machine learning paradigms will be exposed and can be exploited. This should ultimately lead to improved, more robust and simpler machine learning systems. 

We imagine some immediate follow-ups to either make us of our VAE based framework directly or develop different practical alternatives to apply our wholistic continual learning vision to various applications. To name some examples that directly come to mind, one could consider medical imaging, where new disease variants arrive and need to be included, but different devices also feature distinct perturbations and data acquisition fluctuates heavily. Similarly, a robot needing to navigate and learn about the world may directly benefit from our view, where old knowledge needs to be retained, but new knowledge needs to be carefully acquired (e.g. in order to prevent getting stuck in noisy environments  or prioritize meaningless novelty that is unrelated to an overarching task). To name a third of countless examples, autonomous driving could provide an interesting platform. Here unexpected events can occur, sensor failure and deviations can arise, but generally need to be distinguished from learning important changes in the environment or previously unseen objects/obstacles.

The above paragraph already indicates that development towards such applications will benefit from respective datasets that enable required detailed investigations with more exhaustive evaluation protocols. Intuitively, we could now list the fact that we have only considered pseudo-continual datasets ourselves, even though they have been subject of our own critique earlier. For reasons of clarity, we repeat that we have deliberately chosen these datasets, such as CIFAR10/100, to have teaching examples that are very familiar to the majority of readers and easy to relate to. We thus list the investigation of concurrently developed very recent datasets, that are more befitting of continual learning, as a prospect to build up even more insights. Specifically, the newly established NeurIPS Dataset and Benchmarks Track has spawned several datasets with continual learning in mind. The individual datasets have different, yet complementary priorities. For instance, CLEAR \citep{Lin2021} and Wild-Time \citep{Yao2022} focus on different tasks, but share the idea of real-world data that naturally develops over time (e.g. think of a laptop or vehicle changing massively over the years and decades). CLiMB \citep{Srinivasan2022} establishes a benchmark that focuses on cross-modality and will thus be interesting to investigate in going beyond our computer vision examples in this paper. \citet{Zhuang2022} propose two benchmark variants to empirically analyze how continual machine learners compare to human learners on short and long time scales. Finally, \citet{Hess2021} has shared a parametrized graphics simulator in an autonomous driving context. A detailed generative model allows for the creation of sequences with various distribution shifts in order to systematically investigate the impact on continual learning. As detailed, we believe that each of these five datasets has their respective prospects and their investigation will contribute to further broaden our obtained insights to both real-world scenarios and controlled environments.

\subsection{Limitations}
At this point, we re-emphasize that we do not wish to claim that our particular neural network method provides the generally best solution to our previously conducted experiments. Although our specific neural network framework realization clearly shows quantitative promise, our main goal was to highlight the importance of the introduced consolidated viewpoint. Even though our approach has its limitations, it is therefore hard to already grasp them in tangible ways, given that we will first need to explore sets of alternative practical frameworks with the same scope.

To nevertheless give the reader an impression for where our particular VAE approach can be improved, we briefly summarize the main caveats. Primarily, many of these revolve around the choice of VAE. The respective limitations have been discussed in greater detail in our prior work \citep{Mundt2022}, upon which we have built our practical framework in the experimental section of this paper. To keep it concise, the limitations can be attributed to two groups: \\

\emph{Assumptions of the EVT approach:} The EVT approach makes an assumption of the existence of uni-modal clusters per class in the VAE's latent space. Whereas this is encouraged by the linear classification objective to achieve separation between classes under the constraint of following the prior, there is no strict theoretical guarantee. It is conceivable, particularly in higher dimensions, that there exist scenarios in which multiple sub-clusters manifest on one side of a decision boundary. Note that this has not yet been observed empirically in our empirical teaching examples and negative impacts on performance remain to be seen. \\

\emph{Limitations of the generative architecture:} It is well known that a conventional VAE may be outperformed by other methods when it comes to ``scale''. In other words, generation quality is often surpassed on very large datasets. Even though they are not the direct subject of this paper, these generated instances could be used for e.g. a generative rehearsal strategy, as an alternative to the presented real data core sets. Despite not being necessary yet for the teaching examples in this work, our previous work \citep{Mundt2022} has shown that several newer generative modeling advances for VAEs (e.g. autoregression, adversarial training, or introspection) can be used on top to largely overcome this challenge. For clarity, we emphasize that such additional advances are required in order to scale our framework to larger real world data. We note that such approaches often come with additional computational expense. However, we are unaware of any alternatives that presently meet the required criteria to represent our wholistic view, as alternatives are subject to future development.

\section{Conclusion}
We have presented a common viewpoint to naturally unite robust continual and active learning in the presence of the unknown. For each aspect, we have conducted an experimental investigation to provide empirical evidence in support of our viewpoint's benefits. Needless to say, each of our individually presented experiments can be extended with multiple facets and several nuanced applications can be derived and thoroughly investigated. Consequently, we encourage future works to adopt our framework or take a similarly wholistic approach with different practical instantiations. At the very minimum, we would expect future works to rethink current practices and question whether current benchmarks are a realistic reflection of our desiderata for continual machine learning systems. As illustrated throughout the paper, this necessitates stepping out of present closed world benchmark routines.

\section{Acknowledgements}
This work was supported by the German Federal Ministry of Education and Research (BMBF) funded project 01IS19062 ``AISEL'' and the European Union's Horizon 2020 project No. 769066 ``RESIST''.


\bibliographystyle{cas-model2-names}

\bibliography{references}

\end{document}